\documentclass{article}
\usepackage[utf8]{inputenc}
\usepackage{default}

\usepackage[T1]{fontenc}    
\usepackage{hyperref}       
\usepackage{url}            
\usepackage{booktabs}       
\usepackage{amsfonts}       
\usepackage{nicefrac}       
\usepackage{microtype}      
\usepackage[table]{xcolor}         
\usepackage[shortlabels]{enumitem}
\usepackage{graphicx}
\usepackage{subcaption}
\usepackage{xspace}
\usepackage{xr}
\usepackage{color}
\usepackage{lineno}
\usepackage{setspace}
\usepackage{mathtools,amsmath,amsthm,amssymb,amsbsy,pgfplots,graphicx,bbm, xspace, gensymb}
\usepackage{algorithm}
\usepackage{algpseudocode}
\usepackage{bookmark}
\usepackage{authblk}

\pgfplotsset{compat=1.17}

\DeclareMathOperator*{\argmin}{argmin}

\definecolor{blue}{rgb}{0,0.2,0.8}
\definecolor{red}{rgb}{1.0,0,0.0}
\definecolor{green}{rgb}{0, 0.5, 0}

\newcommand{\supplement}{SI\xspace}

\newcounter{oldtocdepth}
\newcommand{\hidefromtoc}{%
  \setcounter{oldtocdepth}{\value{tocdepth}}%
  \addtocontents{toc}{\protect\setcounter{tocdepth}{-10}}%
}
\newcommand{\unhidefromtoc}{%
  \addtocontents{toc}{\protect\setcounter{tocdepth}{\value{oldtocdepth}}}%
}

\title{Systematic human learning and generalization from a brief tutorial with explanatory feedback}

\author[1]{Andrew J. Nam}
\author[1]{James L. McClelland}
\affil[1]{
    Department of Psychology, Stanford University, Stanford, USA
    \protect\\
    \normalsize
    \texttt{\{andrewnam, jlmcc\}@stanford.edu}
}

\begin{document}

\maketitle

\begin{abstract}
Neural networks have long been used to model human intelligence, capturing elements of behavior and cognition, and their neural basis.
Recent advancements in deep learning have enabled neural network models to reach and even surpass human levels of intelligence in many respects, yet unlike humans, their ability to learn new tasks quickly remains a challenge.  People can reason not only in familiar domains, but can also rapidly learn to reason through novel problems and situations, raising the question of how well modern neural network models capture human intelligence and in which ways they diverge.
In this work, we explore this gap by investigating human adults' ability to learn an abstract reasoning task based on Sudoku from a brief instructional tutorial with explanatory feedback for incorrect responses using a narrow range of training examples.  We find that participants who master the task do so within a small number of trials and generalize well to puzzles outside of the training range.  We also find that most of those who master the task can describe a valid solution strategy, and such participants perform better on transfer puzzles than those whose strategy descriptions are vague or incomplete.
Interestingly, fewer than half of our human participants were successful in acquiring a valid solution strategy, and this ability is associated with high school mathematics education.  
We consider the challenges these findings pose for building computational models that capture all aspects of our findings and point toward a possible role for learning to engage in explanation-based reasoning to support rapid learning and generalization.
\end{abstract}


\clearpage

\hidefromtoc

\section{Introduction}
Psychologists and computer scientists alike have long attempted to capture human-level intelligence.
While numerous methods have been used to model intelligent behavior, such as rule-based systems \cite{newell1961gps, chomsky1957syntactic} and probabilistic models \cite{pearl2009causality, tenenbaum2011grow}, neural networks have become the dominant approach due to their overwhelming success in a wide range of domains, such as image recognition \cite{ciregan2012multi}, natural language \cite{vaswani2017attention}, and game playing \cite{mnih2015human, silver2016mastering, vinyals2019grandmaster}.
Foundation models \cite{bommasani2021opportunities} such as GPT-3 \cite{brown2020language} and ChatGPT can produce text that is indistinguishable from human-generated text, even writing full scientific articles capable of convincing peer reviewers \cite{cotton2023chatting}.
Moreover, neural networks can learn representations with structures similar to ones found in brains \cite{yamins2014performance, banino2018vector} and produce emergent behaviors consistent with human cognitive development \cite{munakata2003connectionist, smith2019modeling}, making them useful tools for understanding human intelligence.

However, neural networks still possess critical limitations compared to humans.
They require massive amounts of training data that is orders of magnitude greater than what humans need, \cite{tsividis2017human} and often fail to generalize outside the range of their training examples \cite{lake2018generalization, saxton2018analysing}.
In contrast, humans demonstrate remarkable capacities to learn novel concepts and systematically generalize broadly from one or a few examples \cite{ahn1992schema,stuhlmuller2010learning, lake2015human}.

How are we to reconcile these strengths and weaknesses?
Some researchers have shown that models relying on the principles of symbolic computation can provide a way to allow rapid learning and immediate generalization to a wide range of novel instances consistent with a specified domain \cite{lake2015human}, urging researchers to find ways to build compositionality in \cite{lake2017building} through domain-specific constraints on the structure and taxonomy of constituent types.
For example, learned weights can be shared such as in graph neural networks \cite{battaglia2018relational} by replicating the same set of connection weights across items of the same class to allow systematic generalization of what is learned for one item to all like items.
This weight-sharing scheme is consistent with the views of many theorists who argue that efficient learning requires strong domain-specific inductive biases \cite{spelke2007core}.

In contrast, foundation models \cite{bommasani2021opportunities} such as GPT-3 \cite{brown2020language} that make no domain-specific assumptions have demonstrated some capacity to generalize from one or a few examples of a task or procedure that remain in their context memory when new examples are presented, and recent extensions begin to apply these models to mathematical \cite{hendrycksmath2021, cobbe2021training, wei2022chain, lewkowycz2022solving}, analogical \cite{lampinen2022can, webb2022emergent}, and scientific reasoning \cite{lu2022learn} problems. 
Compared to human reasoning, however, these models appear to yet be limited in several key ways.
First, recent controlled studies on in-context learning suggest that few-shot gains may not depend on the validity of the training examples and instead reflect superficial adaptations \cite{webson2021prompt, min2022rethinking}.
Second, given the massive scale of datasets used to train foundation models, (e.g. Common Crawl \cite{raffel2020exploring} for GPT-3, MassiveText \cite{rae2021scaling} for Gopher and Chinchilla \cite{hoffmann2022training}) it is unclear what sorts of problems are actually novel.
In probing GPT-3 using problems adapted from well-known psychology experiments in the cognitive science literature, \cite{binz2022using} notes that the results may be contaminated by the possibility that GPT-3 has been trained on a corpus containing information about the studies the experiments were drawn from.
Finally, in-context learning typically uses solved examples, not with instructions and explanations about the task as humans often would when learning novel tasks.
In sum, it is unclear how well contemporary foundation models can learn something entirely new from only a few examples.

Building a model that can learn to reason requires benchmarks that evaluates how well an agent can adapt to a novel class of problems with entirely new axioms and properties.
However, despite the growing interest in few-shot learning and reasoning, the base of empirical data focusing specifically on rapid human learning and generalization from one or a few examples is relatively thin.
Existing studies rely on domains where the participants can bring extensive prior experiences to bear, e.g. hand-written alphabets \cite{lake2015human} or cultural practices \cite{ahn1992schema}.
Furthermore, in logical reasoning, an exemplary domain for the role of systematicity in human thought \cite{fodor1988connectionism}, humans can fail to exhibit consistency with basic laws of valid logical inference \cite{wason1968reasoning}, instead exhibiting dependence on the specific context of the propositions they are given to reason from (see \cite{johnson2001mental} for a review).
Thus, many questions remain about the basis on which humans can learn rapidly and generalize broadly beyond the range of examples they have experienced in a novel domain.

A central issue that we believe requires deeper investigation is the role of instructions and explanations in humans' ability to recognize and utilize structure within a novel domain.
When understanding a new conceptual structure such as a cultural practice or scientific procedure \cite{ahn1992schema}, it has been observed that experimenter-provided explanations facilitate participants' learning from a single example.
Humans can also learn to play Atari games by watching 2 minutes of expert play \cite{lake2017building} or by reading instructions about the games \cite{tsividis2017human}.
It has been argued that formal education promotes abstraction, such as identifying relevant and irrelevant features when classifying novel items into categories and that this enhances generalizable reasoning abilities \cite{cole1971cultural, scribner1973cognitive}.
Indeed, it was proposed nearly a century ago that systematic reasoning is first acquired through a form of formal education which then transfers to enable spontaneous generation \cite{vygotsky1934thought}.
These observations raise the question whether contemporary neural networks’ limited capacity to generalize could be overcome by drawing upon explicit instructions and explanations.

The present work seeks to contribute to a consideration of these issues by exploring several questions on human learning and generalization of a problem-solving skill from a brief instructional and explanation-based learning experience using a restricted range of examples.
First, how rapidly do successful learners acquire the solution strategy?
Second, after successfully learning the skill, how well do humans generalize to out-of-distribution samples?
Third, how well can those who acquire their solution strategy describe the strategy they use, and does performance covary with the ability to give a valid description?
Lastly, how universal is the ability to acquire this new problem solving skill and if the ability to acquire the skill is not universal, what factors co-vary with people’s ability to learn the skill quickly and generalize it broadly?  Although we focus on a specific example skill, we believe the answers we obtain to these question will be relevant to understanding
human learning and generalization in systematic problem solving domains, and provide benchmarks and clues that can inform efforts to achieve human-like performance in such domains in artificial systems. 

We address these questions through a novel task which we call the hidden single puzzle, based on a solving technique of the same name in the puzzle Sudoku.
The technique requires the solver to use the digits already present in a grid and the principle of mutual exclusivity to deduce the content of a single designated empty cell (see Section~\ref{sec:main:experiment}). 
The task is appealing as a domain in which to explore the general features of human systematic reasoning ability \cite{lee2008psychological} since a solution technique, such as the hidden single technique as presented in our experiment, can be described in simple explanatory language without the need to appeal to technical concepts, making it potentially accessible to a wide range of human participants. 
Moreover, the task is characterized by the same symmetries, group properties, and combinatorics that characterize Sudoku in general \cite{felgenhauer2006mathematics, russell2006mathematics}, allowing procedural transformations, such as re-assigning the roles of digits, shuffling rows and/or columns, and rotating or transposing the grid. 
This allows us to explore the process of learning within a controlled task subspace and to assess how well learning generalizes outside of the narrow range of examples used in the tutorial and in an initial practice phase of the experiment.

In our study, participants with no prior exposure to Sudoku completed a guided tutorial walking through the solution of a single puzzle, then went through 25 practice puzzles with explanatory feedback but with limited variation in certain puzzle features.  Following this, they received 64 test puzzles with systematic variation in puzzle features designed to assess out-of-distribution generalization. Finally, they completed a brief questionnaire about the strategies they used, their level of general education and their coursework in mathematics.  

Our results provide evidence for the following.  
First, while 1/3 of our participants (those we call \textit{solvers}) showed clear evidence of learning the full solution strategy from the tutorial and practice phase, most of the remaining participants acquired a less successful strategy that yielded the correct solution only 50\% of the time.
Self-reported education, and particularly education in basic high school mathematics, was associated with solver status.  
The successful strategies were acquired rapidly, with 50\% of solvers demonstrating consistently high accuracy starting from the 3rd trial and 90\% by the 10th, and solvers successfully generalized to puzzles outside of the training distribution, albeit with selective performance costs we will detail.
A further finding was that among the solvers, most articulated a description of a valid solution strategy, but some did not, and those that did tended to perform more accurately and generalize more effectively on transfer puzzles.

We bring these findings together in the \textit{Discussion} by proposing that what binds these observations together is learning and reasoning using explanations, and connect these insights to computational models by noting points of divergence between human cognition and machine learning.
We conclude by proposing that in addition to contemporary methods, models will need explanation-based reasoning to reach human levels of abstract reasoning abilities and offer our results as points of comparison for what one might expect from models that do so.

\section{Materials and methods}
\label{sec:main:experiment}

\subsection{Design and participants}
After extensive pilot testing and refinement of our experimental protocol, we preregistered the design and some of the key analyses of the results of our experiment (see the preregistration document at \url{https://osf.io/7rehu}).  Both the pilot and final study were conducted using Amazon Mechanical Turk. Workers were required to be US nationals. 

Several features of our design revolved around the goal of characterizing generalization performance of participants who successfully acquired a new skill through our explanatory tutorial and practice with a narrow range of problem variations.  We used a stringent criterion of learning success (described below), and sought to obtain a large enough sample of participants who met this criterion to allow reliable detection of small but potentially meaningful accuracy differences between control and generalization test problems within the first few test trials after the practice phase, using findings from our pilot studies to guide our choice of sample sizes to allow replication of suspected effects.  To this end, we recruited new batches of participants until we had at least 75 that met our criterion for solving the task by the end of the 25 practice trials. Participants received a base payment for completing the study plus a bonus for each puzzle solved on the first attempt (median total payment \$9.93 in 47.3 minutes).
Participants were not notified of the purpose of the experiment.

\subsection{Task description}

\begin{figure}[!ht]
\begin{subfigure}{.33\linewidth}
  \centering
  \caption{Tutorial}
  \includegraphics[width=.95\linewidth]{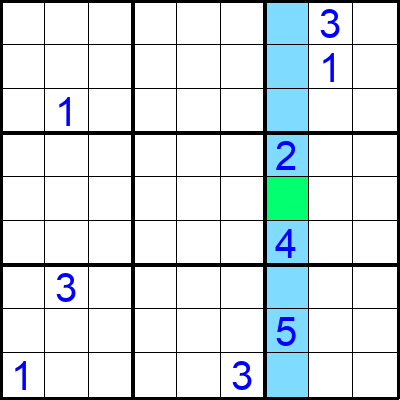}
  \label{fig:puzzles_d}
  \vspace{14pt}
\end{subfigure}
\begin{subfigure}{.33\linewidth}
  \centering
  \caption{House Type}
  \includegraphics[width=.95\linewidth]{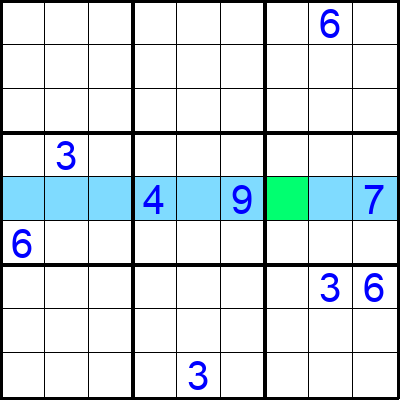}
  \label{fig:puzzles_e}
  \vspace{14pt}
\end{subfigure}
\begin{subfigure}{.33\linewidth}
  \centering
  \caption{House Index}
  \includegraphics[width=.95\linewidth]{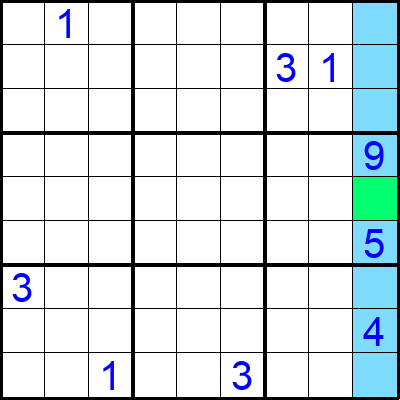}
  \label{fig:puzzles_f}
  \vspace{14pt}
\end{subfigure}
\begin{subfigure}{.33\linewidth}
  \centering
  \caption{Cell Index}
  \includegraphics[width=.95\linewidth]{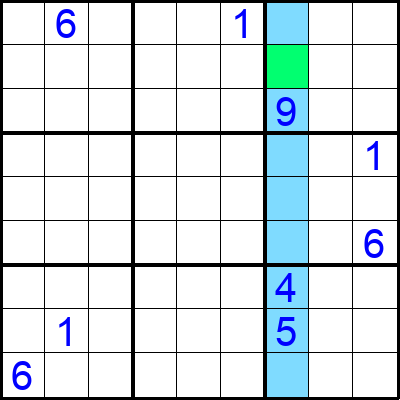}
  \label{fig:puzzles_g}
\end{subfigure}
\begin{subfigure}{.33\linewidth}
  \centering
  \caption{Digit Set}
  \includegraphics[width=.95\linewidth]{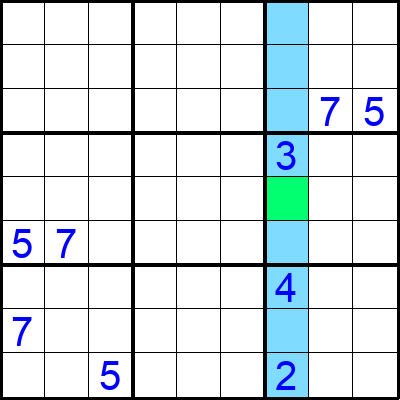}
  \label{fig:puzzles_h}
\end{subfigure}
\begin{subfigure}{.33\linewidth}
  \centering
  \caption{All}
  \includegraphics[width=.95\linewidth]{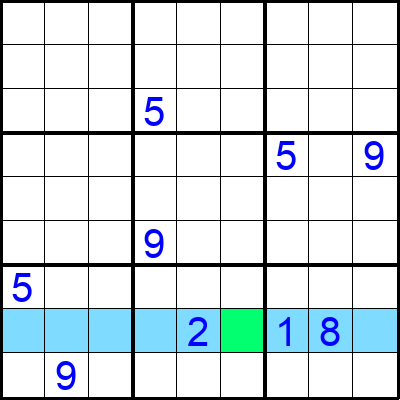}
  \label{fig:puzzles_i}
\end{subfigure}

\caption{Examples of hidden single puzzles.
\\
Puzzles were generated using the participant's individualized, random assignment of the training house type, house index, cell index, and digit set. (a) The full hidden single puzzle configuration as the participant might see it part way through the tutorial, conforming to the house type, house index, cell index, and containing the target (3) and the distractor (1) digits, both drawn from the training digit set.  
Examples used in the practice phase and control examples from the test phase would all use the same house type, house index, cell index, and training digit set, in this case the digits (1, 3, 6, 8).
(b-f) Examples of non-control puzzles used in the test phase containing changes in features from the puzzles used in the tutorial and practice phase. (b-e) Puzzles with exactly one changed feature. (f) Puzzle with all four features changed.}
\label{fig:puzzles}
\end{figure}

The hidden single puzzles (Figure~\ref{fig:puzzles}) follows the same constraints as Sudoku: in a valid solution, each row, column, and 3x3 box must contain exactly one instance of each number from 1 to 9.  However, the hidden single puzzles are simplified so that there is a single \textit{target} digit that must go into a green \textit{goal} cell and not any other cell in the blue-highlighted \textit{target house}.  Puzzles are procedurally generated to have controlled variations while also maintaining standardized difficulty (see \supplement Section~\ref{sec:supp:puzzle_generation}).
Puzzles always contain 5 different digits called \textit{hints}, one of which is the target and another of which is the \textit{distractor}.
The target and distractor digits each have 3 instances arranged in the grid in a way that prevents participants from performing reliably based on perceptually obvious heuristics such as counting the number of instances of a digit.  The other three hints, called \textit{in-house} digits, each occur once in the highlighted house. The remaining four of nine digits, called \textit{absent} digits, do not appear in the grid.

There are four experimentally controlled variable features of the puzzles. The \textit{house type} is the type of house (row or column) to apply the hidden single technique to, indicated by whether it is a row or column that is highlighted in blue. 
The \textit{house index} is the house to apply the hidden single technique to, also indicated by the blue highlighting. 
The \textit{cell index} indicates which cell to solve for within a house, indicated by the cell highlighted in green. Lastly, the \textit{digit set} is the set of 4 digits from which the target and distractor digits are drawn. 
The digits used in each puzzle are determined by first selecting a digit set, then assigning a number from the set as the target digit and a different number from the same set as the distractor digit. 
The three in-house digits are then randomly selected from the remaining 7 digits.

\subsection{Procedure}

Here, we provide a brief characterization of the experimental protocol for each of the phases of the experiment.
See \supplement Section~\ref{sec:supp:exp_design} for further details.

At the start of the experiment, each participant was randomly assigned a specific house type, house index, goal cell index, and a training set of four digits to be used in the tutorial and practice puzzles. 
For each participant, a \textit{transfer} set of four digits was then selected from the 5 digits remaining.

The experiment began with a tutorial centered on one hidden single puzzle randomly generated for each participant, such that all exercises in the tutorial built off of one specific exemplar. 
The restriction to a single puzzle was deliberate, as we wanted to later evaluate the participant's ability to generalize to out-of-distribution transfer puzzles.
First, participants were provided a one-sentence description of Sudoku as ``a puzzle with a 9x9 grid of numbers where each row, column, and 3x3 box must contain exactly one of each number from 1 to 9''. 
After two brief exercises illustrating this description, the tutorial walked the participant through a sequence of exercises demonstrating the solution steps to the participant's specific puzzle.
Throughout the tutorial, we used different colors to highlight specific elements in the grid and refer to them, e.g. `the purple cell' or the `blue row' (see \supplement Figure \ref{fig:tutorial}), taking care to avoid any abstract statements that described the strategy using general principles or rules and to avoid any indication that this pattern of reasoning could be applied beyond the exemplar puzzle.
Throughout the sequence, participants were required to enter responses that, if incorrect, resulted in an explanation and a requirement to correct the error before proceeding.

Following the tutorial, participants were given 25 puzzles to solve as part of a \textit{practice phase}. 
All puzzles in this phase shared the same house type, house index, cell index, and digit set as the tutorial, only varying in the hint locations and the choices of specific digits to serve in the various hint roles, subject to the constraints imposed by the digit set.
Participants were allowed unlimited attempts and time to solve each puzzle with the goal of giving them the best chance of mastering the hidden single technique. 
During this phase, all incorrect attempts produced detailed explanations specific to the puzzle and given response (see \supplement Section~\ref{sec:supp:practice_phase}, referring to the particular digits and hints for why the response was incorrect.
The participants were not informed of the relationship between the puzzle used in the tutorial and the puzzles in this phase.

Next, in the \textit{test phase}, participants were given 64 puzzles with systematic variations from the puzzles encountered thus far to evaluate transfer ability.  We defined four conditions based on whether or not a puzzle differed from the tutorial and practice puzzles along the four features (digit set, house type, house index, cell index), yielding 16 combinations of possible puzzle conditions (including control puzzles which share all features with the tutorial and practice phase puzzles). The trials were internally (not revealed to participants) grouped into 8 sets of 8, where every set contained all 8 combinations of house type, house index, and cell index conditions, and an equal number of puzzles with and without a change in the digit set, and every two sets of 8 puzzles contained all 16 combinations including the digit set. During this phase, participants were allowed a single attempt and a 2-minute time limit for each puzzle. After submitting their responses, participants only received feedback on whether the response was correct or incorrect.

\section{Results} 
\label{sec:main:results}

\begin{figure}[!ht]
\centering
\includegraphics[width=\linewidth]{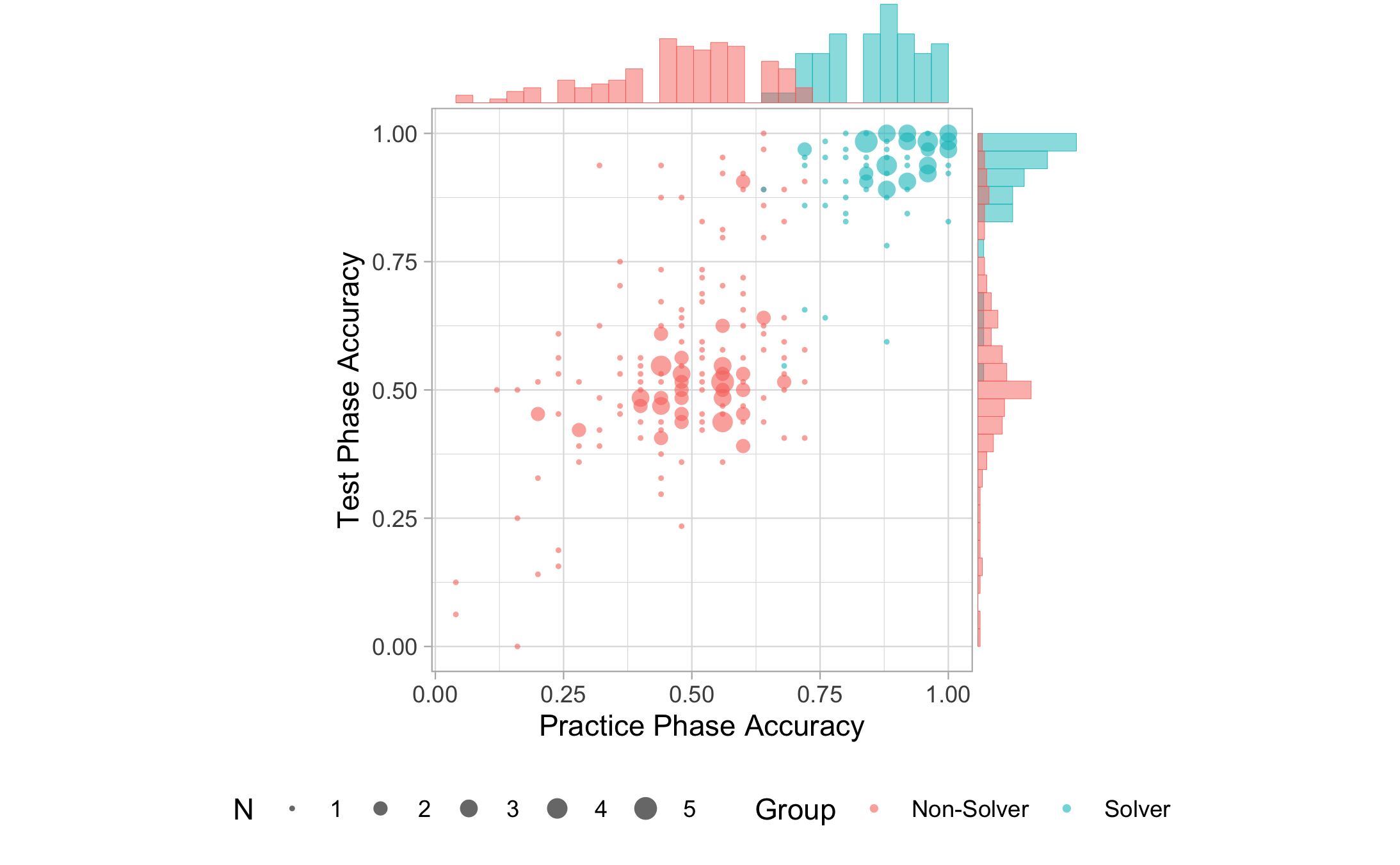}
\caption{Overall accuracy during the practice and test phases. 
\\
Each point represents one participant, colored to indicate whether the individual was classified as a solver or non-solver, as described in the main text. Marginal histograms show the proportion of individuals in each group achieving the each level of accuracy.}
\label{fig:group_accuracy}
\end{figure}

As previously stated, a key goal in our study was to understand the degree of systematic generalization in humans who had acquired a valid solution strategy from a narrow range of experience like that provided during the practice phase which we could then probe for in the test phase.  1,985 people entered the study and after screening out 1,714 who demonstrated or attested to prior Sudoku ability or experience (See \supplement Section \ref{sec:supp:diagnostics}), we collected data from 271 participants (\textit{age}: M = 39, SD = 12; \textit{gender}: 119 male, 149 female, 1 other, 2 no response), of whom 88 met the criterion.  Of these, a subset who we refer to as \textit{solvers}, acquired a successful solution strategy within the 25 puzzles of the practice phase according to our preregistered classification method, developed during pilot studies. 
Specifically, we fitted a logistic regression (see \supplement Section~\ref{sec:supp:reg_overall_accuracy}) to predict the accuracy of the practice phase trials and classified participants as solvers if their predicted accuracy on the 25th trial exceeded a decision threshold of 0.8.
This criterion may have been conservative in that it excluded some participants that were classified as non-solvers but reached a high level of accuracy during the test phase (see Figure~\ref{fig:group_accuracy}), as well as leaving out participants that may have learned partially successful strategies (see \supplement Section~\ref{sec:supp:qrating_options}).

We identified 88 participants who met this criterion, leaving 183 participants whom we refer to as \textit{non-solvers}.   Figure~\ref{fig:group_accuracy} shows the overall accuracy of each participant in the experiment, color coded by solver status, plotting the participant's accuracy in the tear phase against their accuracy in the practice phase.

The bulk of those classified as solvers solved puzzles with very high levels of accuracy in the test phase, while most non-solvers appeared to have adopted a strategy they continued to employ throughout the test phase, in which they chose between the target and the distractor with only a 50\% chance of being correct.  However, a subset of those classified as non-solvers achieved high levels of accuracy during the test phase, and some others may have adopted partially successful strategies.
In what follows, we focus primarily on the performance of the solvers, contrasting their performance with that of non-solvers in certain cases.

\subsection{Dynamics of strategy acquisition} 
\label{sec:main:results_p1}

Since all puzzles used in our experiment share the same structure, we classify responses to a puzzle into four categories based on the role of the digit given as the response:
1.~\textit{target}: the correct digit for the goal cell, which appears three times in each puzzle; 
2.~\textit{distractor}: the incorrect digit, which also appears three times in each puzzle; 
3.~\textit{absent}: any digit that does not appear in the puzzle at all; and 
4.~\textit{in-house}: any digit that already appears in the blue-highlighted house.
Each puzzle contains one, one, four, and three digits in each category, respectively.
In the example in Figure~\ref{fig:puzzles_d}, these digits are $\{3\}$, $\{1\}$, $\{6, 7, 8, 9\}$, $\{2, 4, 5\}$.

The response categories also help define four different strategy classes based on which responses are expected to occur under each:
1.~\textit{uninformed guess} (UG): responses are completely unconstrained by the hints in the puzzle, selecting from any of the 9 digits.
2.~\textit{avoid direct contradictions} (ADC): responses only avoid in-house digits, selecting from any of the remaining 6 digits.
3.~\textit{prevalent digits} (PD): distractor and target digits are selected with equal probability.
4.~\textit{successful} (S): the target digit is consistently selected. 

Ignoring errors, the four strategy classes are expected to produce correct responses with probabilities 11.1\%, 16.7\%, 50\%, and 100\% respectively. 
Note that different specific strategies can produce responses consistent with each of the listed strategy classes.  
For example, a participant choosing randomly between the two prevalent digits, and a participant choosing between them on an arbitrary basis unrelated to the correctness of the response (e.g., always choosing the numerically larger of the two) would both be assigned to the PD strategy class. 
While other strategies and heuristics may be possible (see \supplement Section~\ref{sec:supp:qrating_options} basis A), these strategy classes account for most of the participants' response profiles and are consistent with post-experiment survey responses (Section~\ref{sec:questionnaire}).  Below we use the word 'strategy' to refer to a strategy class unless otherwise indicated.

Using these strategy definitions, we model the participants' aggregate pattern of responses by assigning a weight to each strategy at each trial.
First, combining the normative response probability for each strategy and how many digits belong to each category, we define a response emission matrix $\mathbf{R}$ where $\mathbf{R}_{i, j} = P(r_j | s_i)$, or the probability that response in category $r_j$ would be produced under strategy $s_i$.
However, participants may make errors, such as selecting a distractor digit even when using a successful strategy.
We account for possible errors by allowing each strategy to occasionally fall back onto weaker strategies, but never stronger ones, represented by an error matrix $\mathbf{W}$ where $\mathbf{W}_{i, j}$ is the probability that strategy $s_i$ will use the response probabilities of strategy $s_j$.
Next, we use a transition probability matrix $\mathbf{X}$ where $\mathbf{X}_{i, j} = P(s_j^{(t+1)} | s_i^{(t)})$, representing the average rate that participants transition from state $s_i$ to state $s_j$.
Presumably, participants would use the best strategy available to them and not regress to a worse strategy with a lower expected accuracy, such as using the ADC strategy after having discovered the PD strategy.
We explicitly adopt this assumption in our model by setting the transition probability from superior strategies to inferior strategies to 0.
Finally, we represent the probability that a participant begins the practice phase with each strategy using the vector $\mathbf{a}$.

We use these parameters to define an aggregate model that describes the behavior of a group of participants collectively: 
\begin{equation}
\label{eq:hmm_agg}
    P(r^{(t)}) = \mathbf{a} \mathbf{X}^{t-1} \mathbf{WR}
\end{equation}
with the following parameters:
\begin{align*}
  &&
    \mathbf{a} = 
        \begin{bmatrix}
          a_1 & a_2 & a_3 & a_4 \\
        \end{bmatrix}
  \\\\
  \mathbf{X} = 
    \begin{bmatrix}
      x_{11} & x_{12} & x_{13} & x_{14} \\
      0      & x_{22} & x_{23} & x_{24} \\
      0      & 0      & x_{33} & x_{34} \\
      0      & 0      & 0      & 1 
    \end{bmatrix}
  &&
  \mathbf{W} = 
    \begin{bmatrix}
      w_{11} & 0 & 0 & 0 \\
      w_{21} & w_{22} & 0 & 0 \\
      w_{31} & w_{32} & w_{33} & 0 \\
      w_{41} & w_{42} & w_{43} & w_{44}
    \end{bmatrix}
  &&
  \mathbf{R} = 
    \begin{bmatrix}
      3/9 & 4/9 & 1/9 & 1/9 \\
      0 & 4/6 & 1/6 & 1/6 \\
      0 & 0 & 1/2 & 1/2 \\
      0 & 0 & 0 & 1
    \end{bmatrix}
\end{align*}
Since we expect solvers and non-solvers to have very different behaviors, we fit the model parameters separately for each group.
We estimate the parameters by minimizing a cross-entropy loss function
\begin{equation}
    \mathcal{L} = - \sum\limits_{p} \sum\limits_{t=1}^{25} r_p^{(t)} \log (\mathbf{a} \mathbf{X}^{t-1} \mathbf{WR})
\end{equation}
where $r_p^{(t)}$ is the one-hot vector representation of participant $p$'s response on trial $t$.
We perform this optimization using gradient descent with Adam \cite{kingma2014adam}.
For fitted parameter values, see \supplement Section~\ref{sec:supp:agg_model_params}.

Assuming that participants make their decisions using exactly one of these strategies on each puzzle, we can model individual participants using a hidden Markov model (HMM) that assigns a single strategy at each trial.
Although there are $4^{25}$ combinations of strategies, due to the monotonic improvement constraint, our model only spans 3,276 possible sequences of strategies.
We use the aggregate parameters as priors for inferring the probability that a participant used a particular strategy on a given trial.
For a given strategy path $\pi$, its prior probability is 
\begin{align}
    P(\pi) = P(s^{(1)}, ..., s^{(25)} ) &= P(s^{(1)}) P(s^{(2)} | s^{(1)}) ... P(s^{(25)} | s^{(24)})
    \\
    &= a_{s^{(1)}} x_{s^{(1)} s^{(2)}} ... x_{s^{(24)} s^{(25)}}
\end{align}
The likelihood of a participant's sequence of responses under $\pi$ is 
\begin{align}
    P(r^{(1)}, ..., r^{(25)} | \pi) &= P(r^{(1)} | s^{(1)}), ..., P(r^{(25)} | s^{(25)}) )
    \\
    &= r_{s^{(1)} r^{(2)}} ... r_{s^{(25)} r^{(25)}}
\end{align}
where $r_{ij}$ is the entry in $\mathbf{R}$ on row $i$ and column $j$.
We then use Bayes' rule to find the posterior probability of a strategy path given a participant's responses
\begin{align}
    P(\pi | r^{(1)}, ..., r^{(25)}) = \dfrac{P(r^{(1)}, ..., r^{(25)} | \pi) P(\pi)}{\sum\limits_{i = 1}^{3276} P(r^{(1)}, ..., r^{(25)} | \pi_i) P(\pi_i)}
\end{align}

\subsubsection{Solvers' strategy dynamics}

\begin{figure}[!h]
\begin{subfigure}{.5\linewidth}
  \centering
  \caption{Actual incorrect responses}
  \vspace{-6pt}
  \includegraphics[width=.95\linewidth]{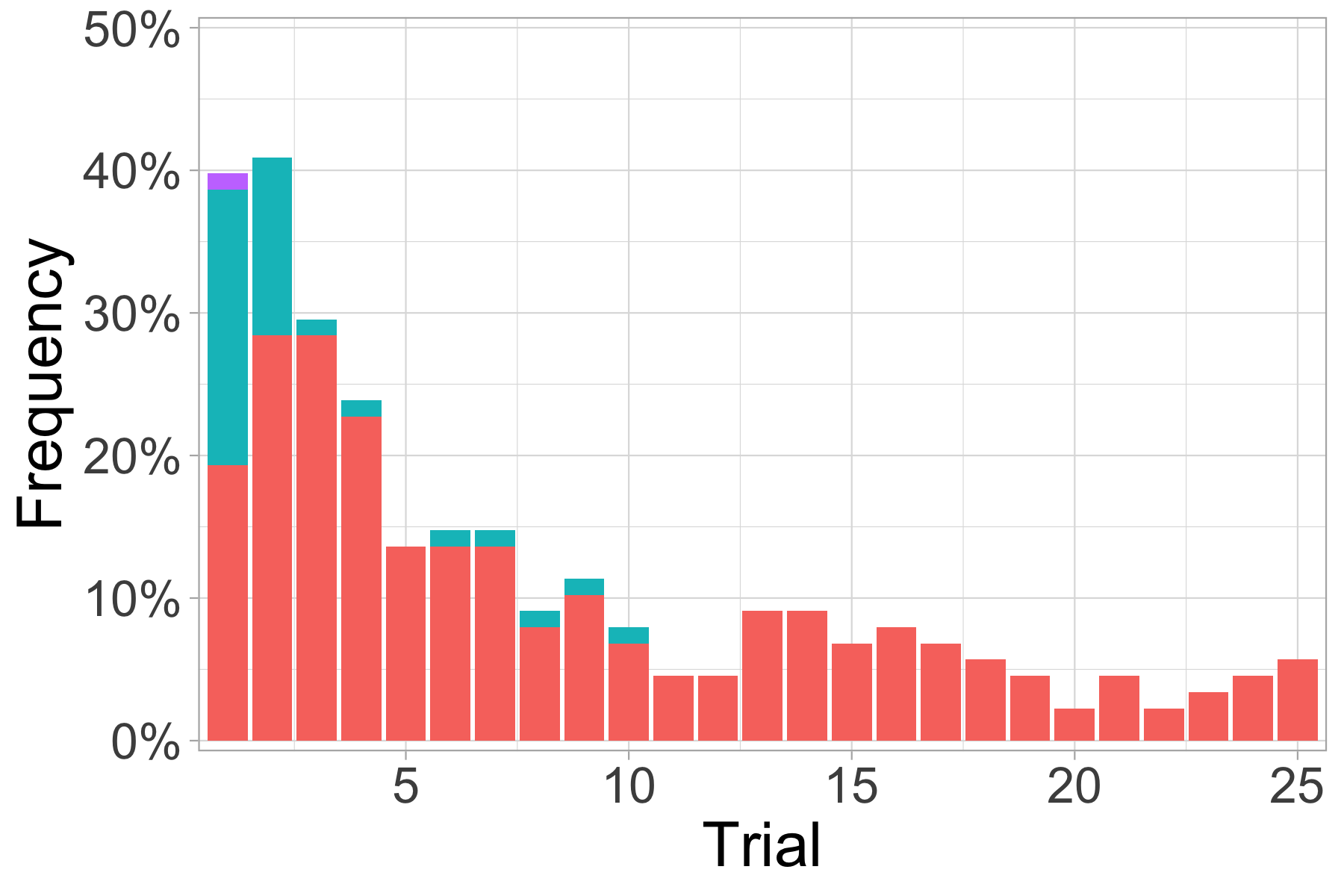}
  \label{fig:phase1:solvers:a}
\end{subfigure}
\begin{subfigure}{.5\linewidth}
  \centering
  \caption{Inferred aggregate distribution of strategies}
  \vspace{-6pt}
  \includegraphics[width=.95\linewidth]{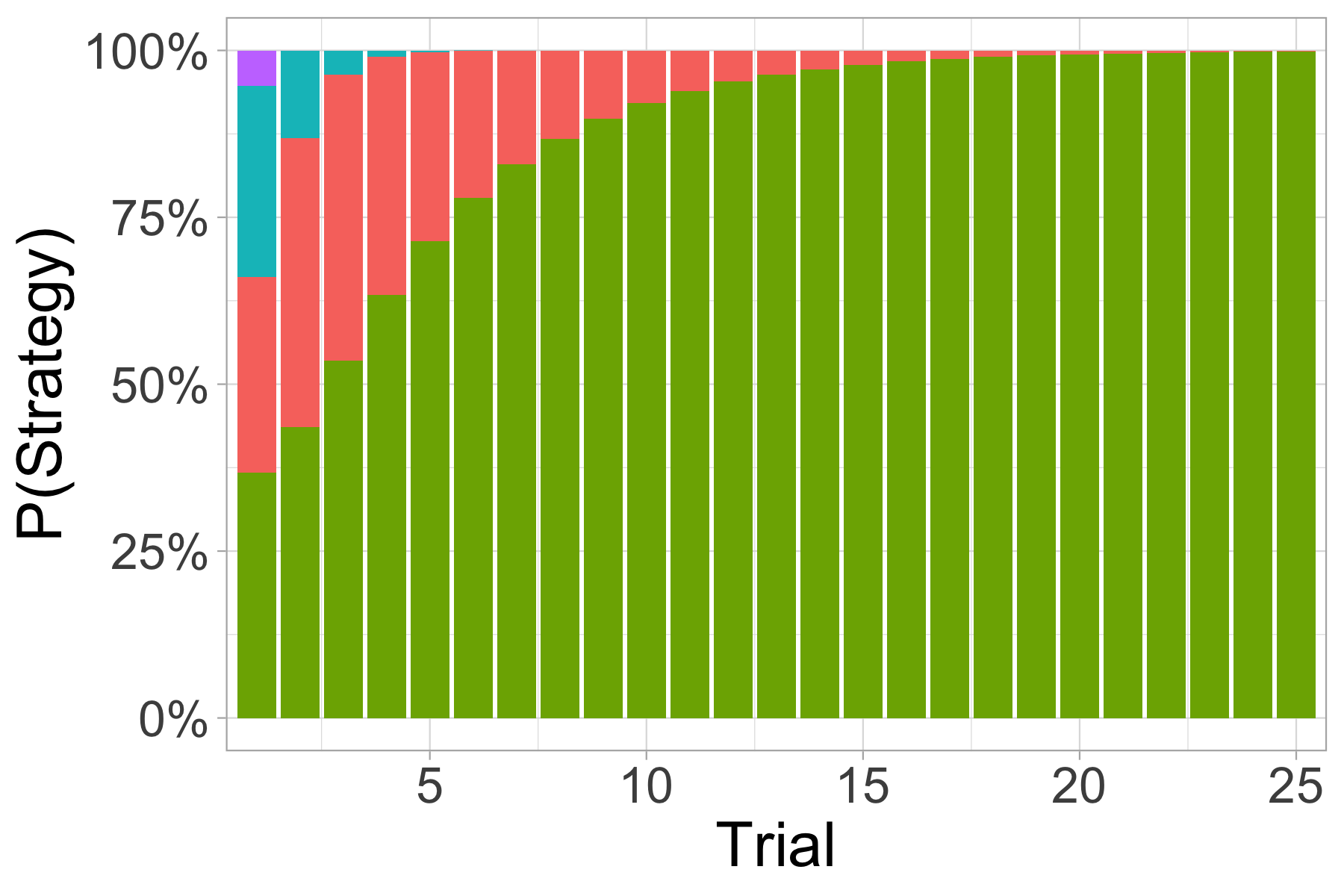}
  \label{fig:phase1:solvers:b}
\end{subfigure}
\begin{subfigure}{.5\linewidth}
  \centering
  \caption{Predicted distribution incorrect responses}
  \vspace{-6pt}
  \includegraphics[width=.95\linewidth]{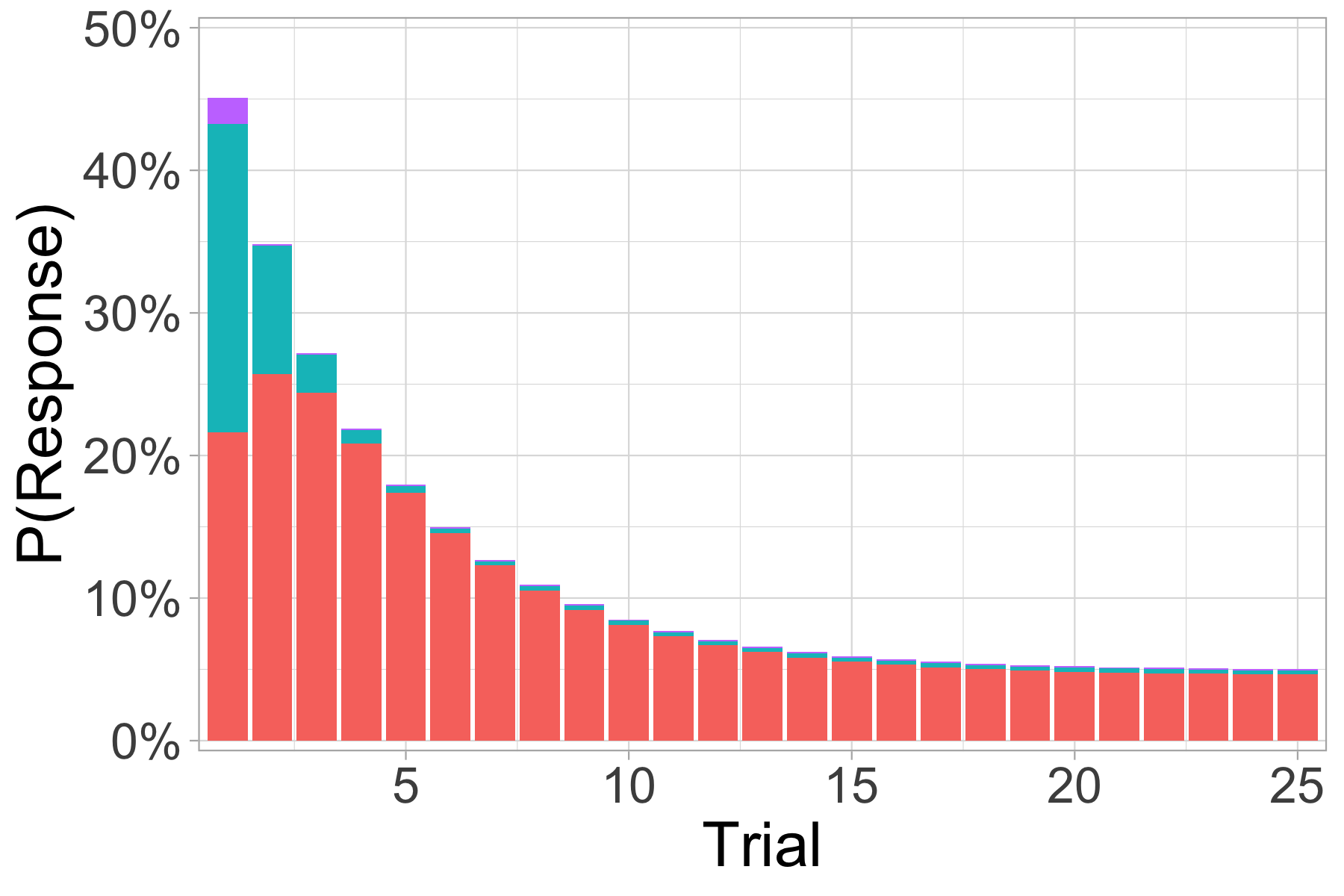}
  \label{fig:phase1:solvers:c}
\end{subfigure}
\begin{subfigure}{.5\linewidth}
  \centering
  \caption{Inferred distribution of individuals' strategies}
  \vspace{-6pt}
  \includegraphics[width=.95\linewidth]{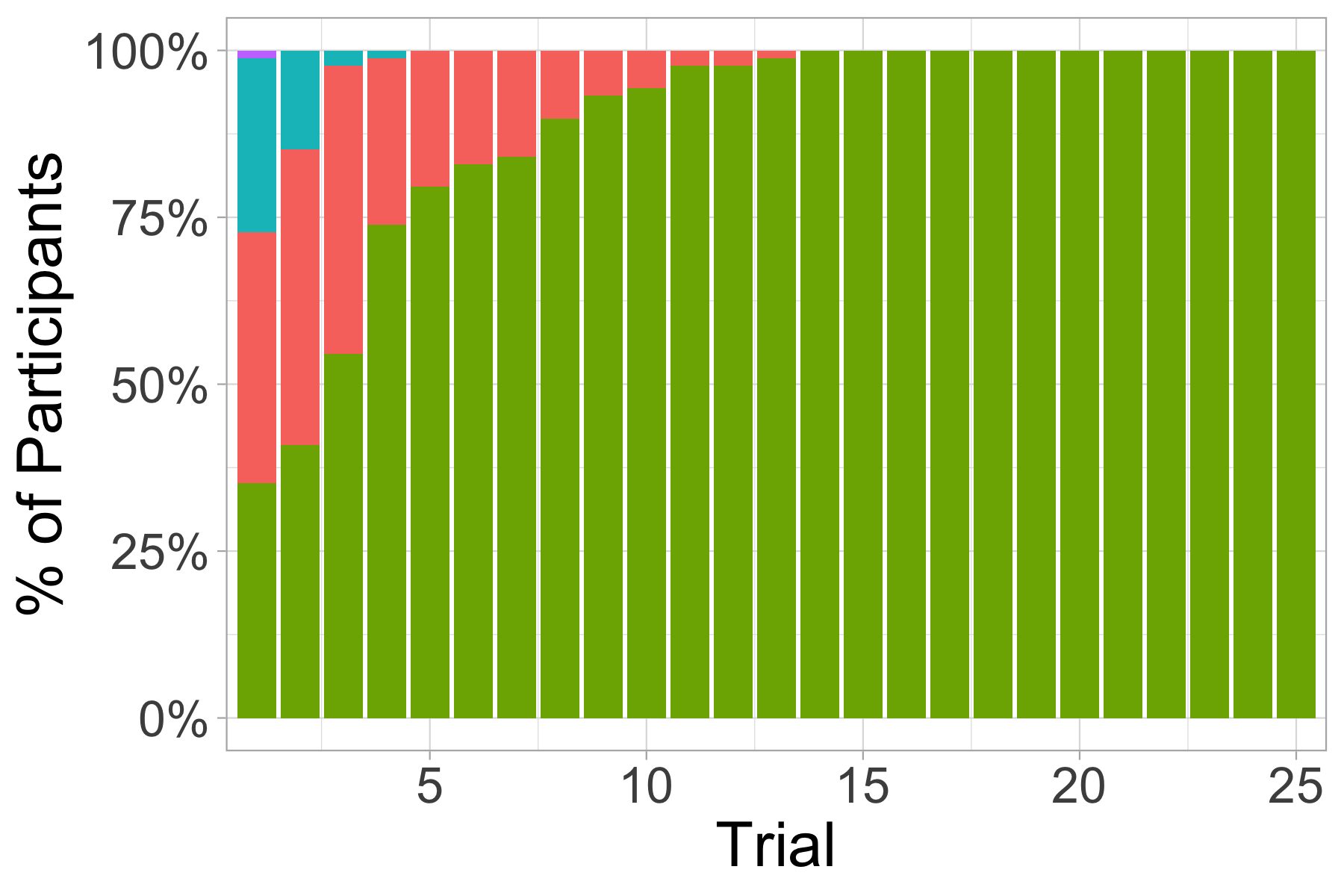}
  \label{fig:phase1:solvers:d}
\end{subfigure}
\begin{subfigure}{.5\linewidth}
  \centering
  \vspace{6pt}
  \includegraphics[width=\linewidth]{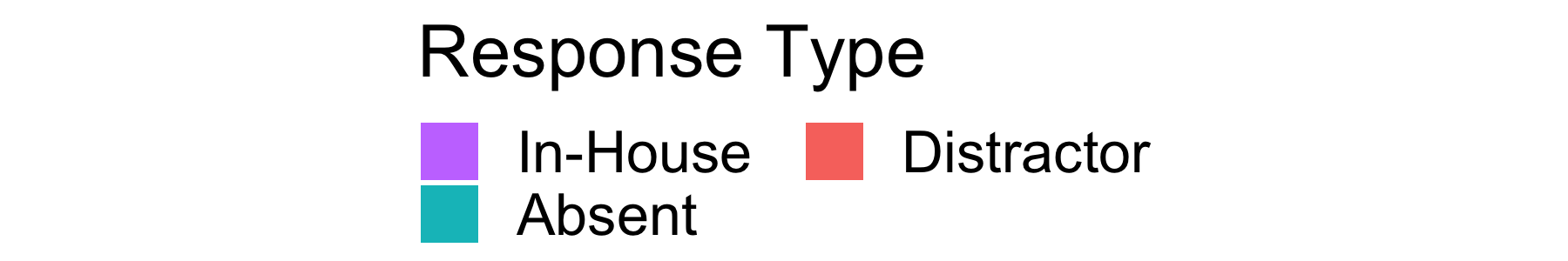}
\end{subfigure}
\begin{subfigure}{.5\linewidth}
  \centering
  \vspace{6pt}
  \includegraphics[width=\linewidth]{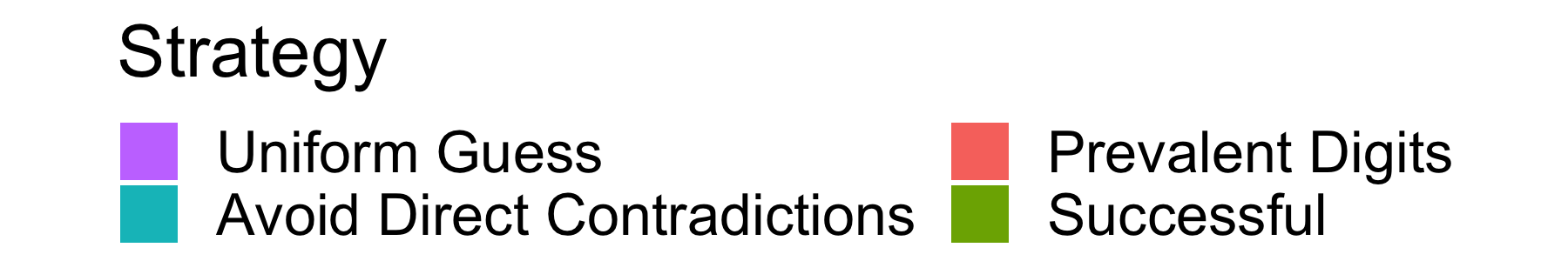}
\end{subfigure}
\caption{Response types and inferred strategies during the practice phase for participants identified as solvers 
\\ 
(a) Actual frequency of incorrect response types (b) Inferred distribution of strategies by the aggregate model (c) Predicted distribution of incorrect response types by the aggregate model (d) Proportion of solvers inferred by individually fitted models to be using a strategy at least as effective as the strategy indicated}
\label{fig:phase1:solvers}
\end{figure}

The aggregate group patterns suggest that participants in the solver group learned successful strategies well before the end of the practice phase (Figure~\ref{fig:phase1:solvers:b}), demonstrating an impressive sample efficiency relative to what is usually seen when training a neural network.

As the figure shows, the inferred initial strategy distribution specifies that 36.8\% of the responses were based on a successful strategy on the first practice trial, with 29.3\%, 28.6\% and 5.3\% based on PD, ADC, and UG strategies respectively.
Use of the UG and ADC strategies rapidly disappears, and the transition to a successful strategy is 92.1\% complete by trial 10.
The model's corresponding expected response distribution, shown in Figure~\ref{fig:phase1:solvers:c}, captures the main features of the pattern of the participants' actual error responses shown in Figure~\ref{fig:phase1:solvers:a}.
The residual distractor responses shown are attributed to errors that occur with a probability of 4.5\% under the successful strategy.

\begin{figure}[!h]
\begin{subfigure}{.5\linewidth}
  \centering
  \caption{Participant 1}
  \vspace{-6pt}
  \includegraphics[width=.95\linewidth]{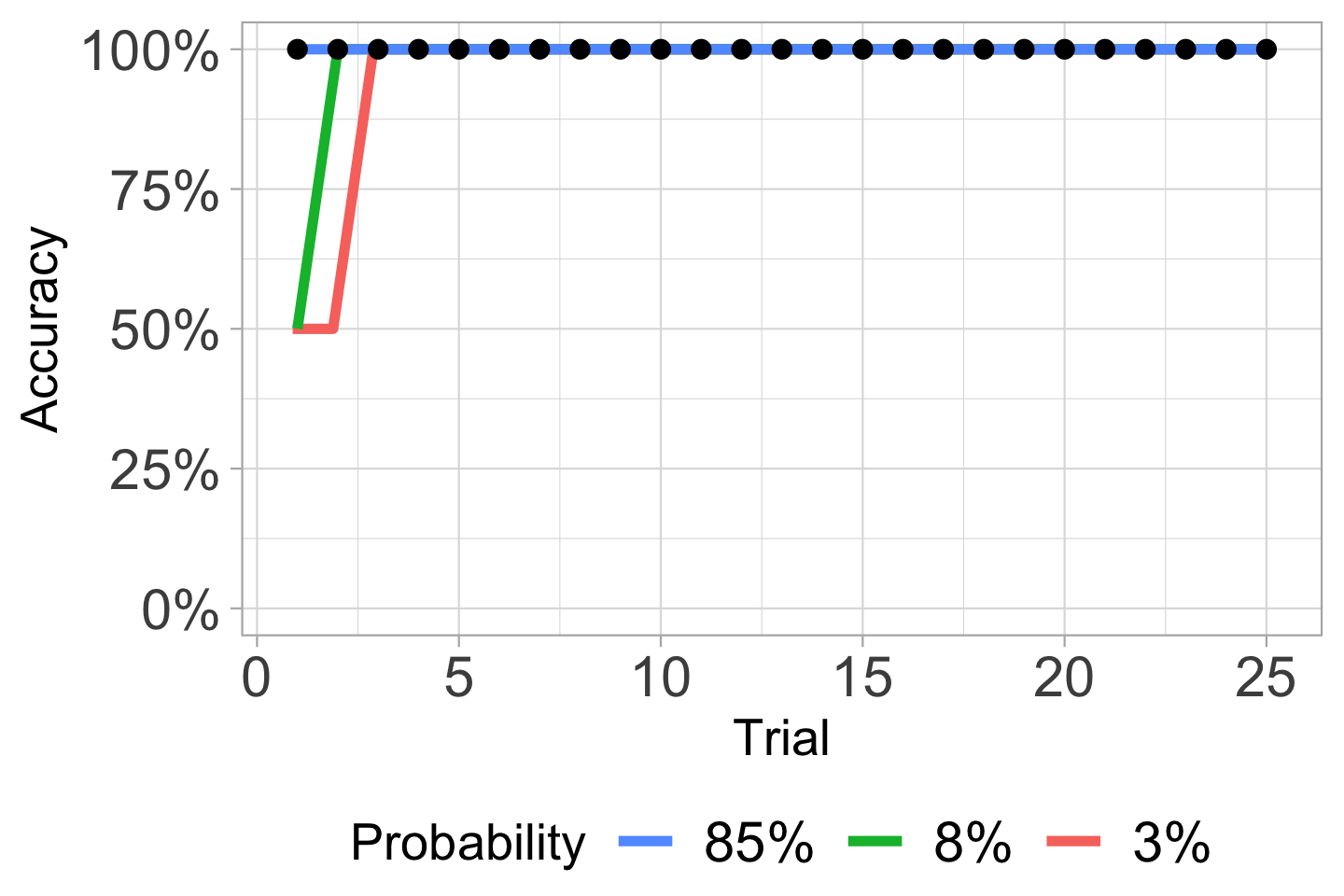}
  \label{fig:p1_paths:a}
\end{subfigure}
\begin{subfigure}{.5\linewidth}
  \centering
  \caption{Participant 80}
  \vspace{-6pt}
  \includegraphics[width=.95\linewidth]{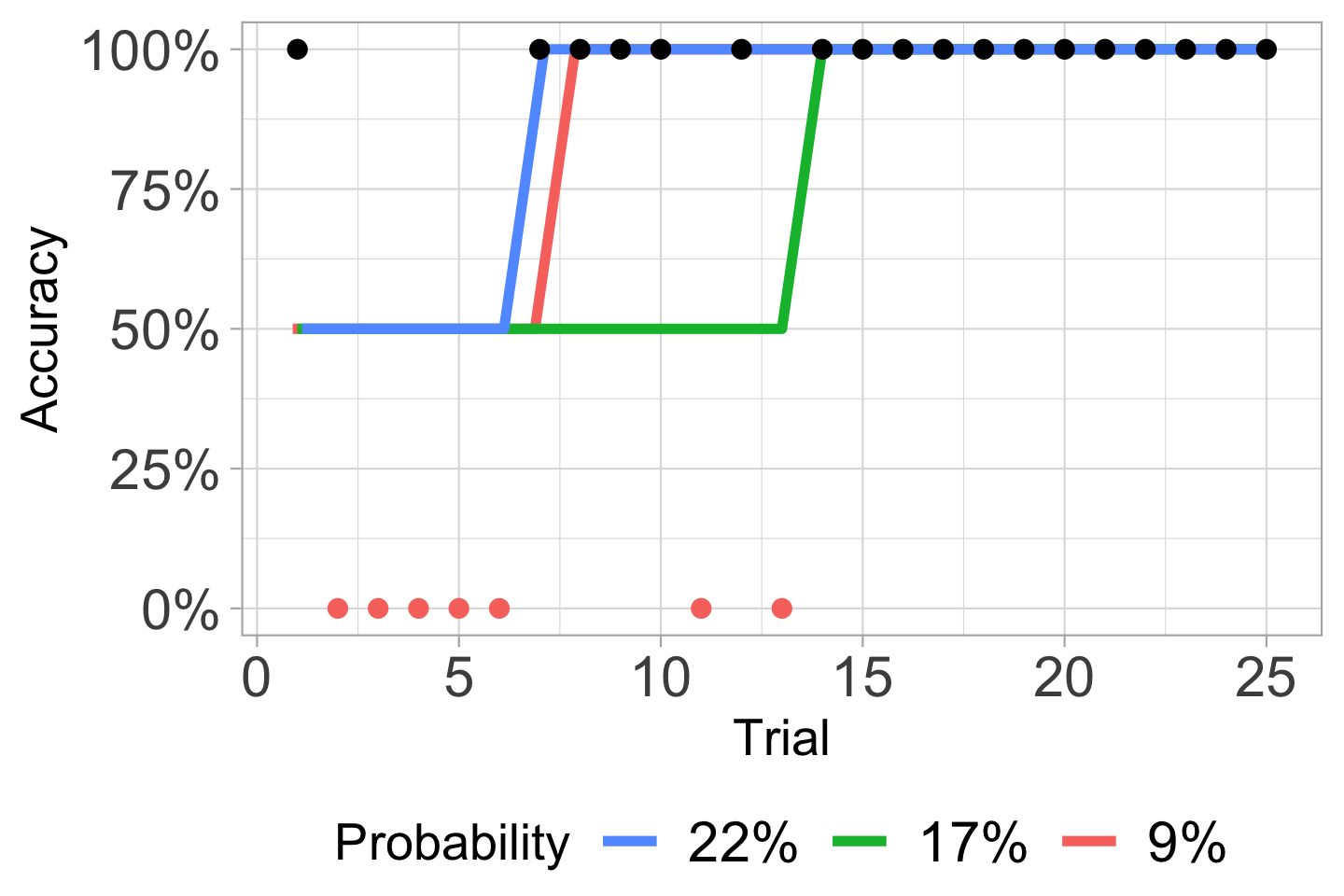}
  \label{fig:p1_paths:b}
\end{subfigure}
\par\bigskip
\begin{subfigure}{.5\linewidth}
  \centering
  \caption{Participant 81}
  \vspace{-6pt}
  \includegraphics[width=.95\linewidth]{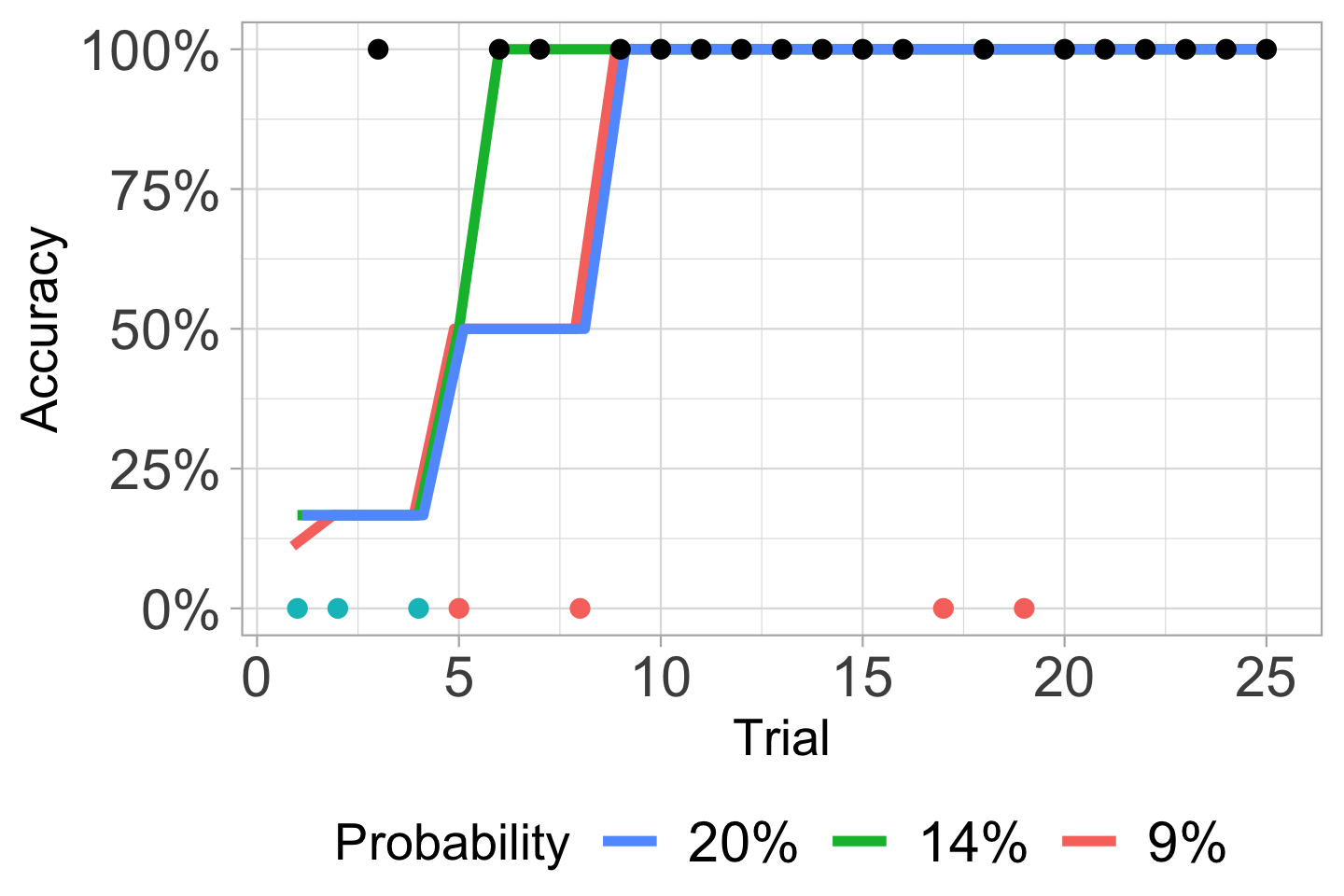}
  \label{fig:p1_paths:c}
\end{subfigure}
\begin{subfigure}{.5\linewidth}
  \centering
  \caption{Participant 154}
  \vspace{-6pt}
  \includegraphics[width=.95\linewidth]{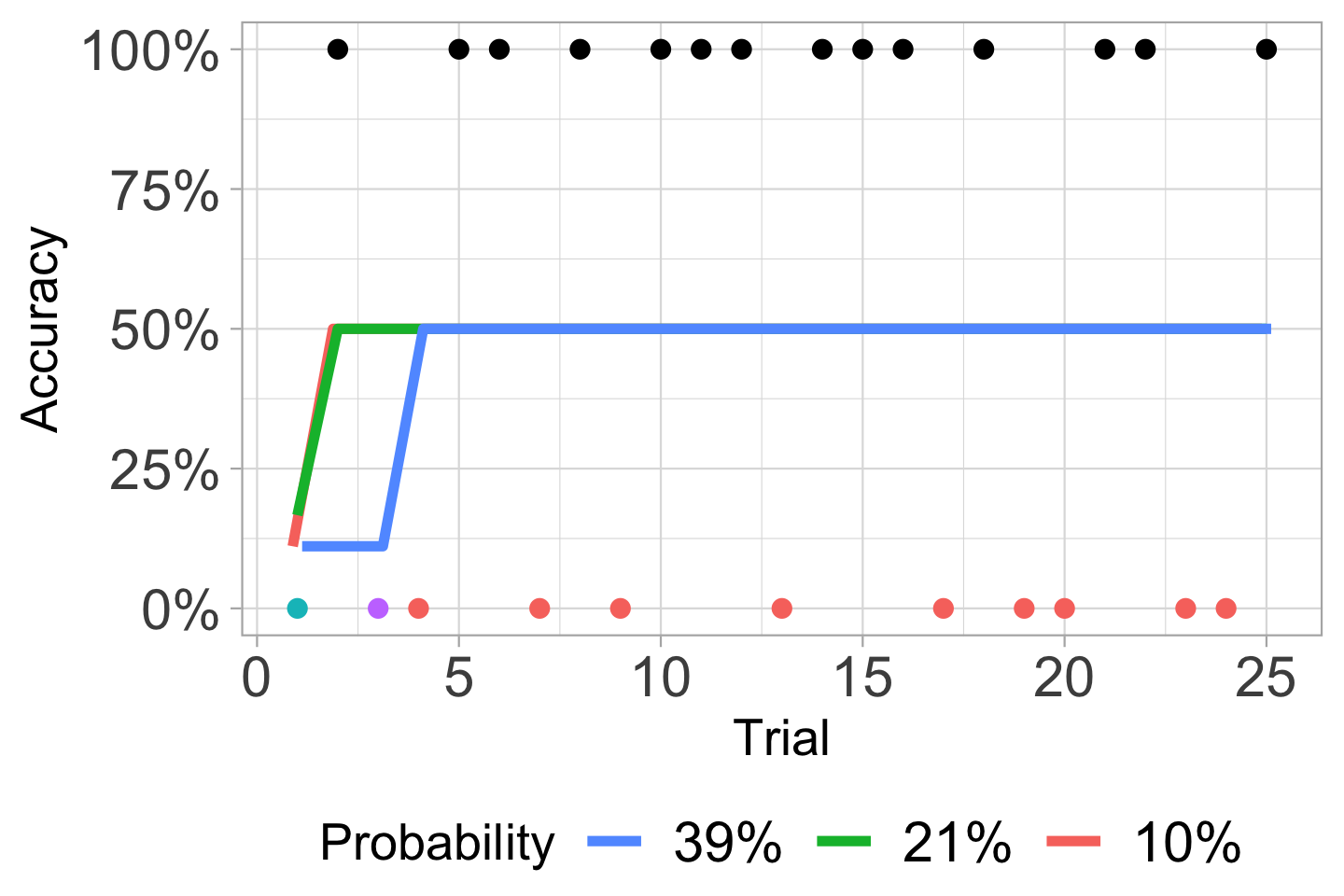}
  \label{fig:p1_paths:d}
\end{subfigure}
\begin{subfigure}{\linewidth}
  \centering
  \includegraphics[width=.7\linewidth]{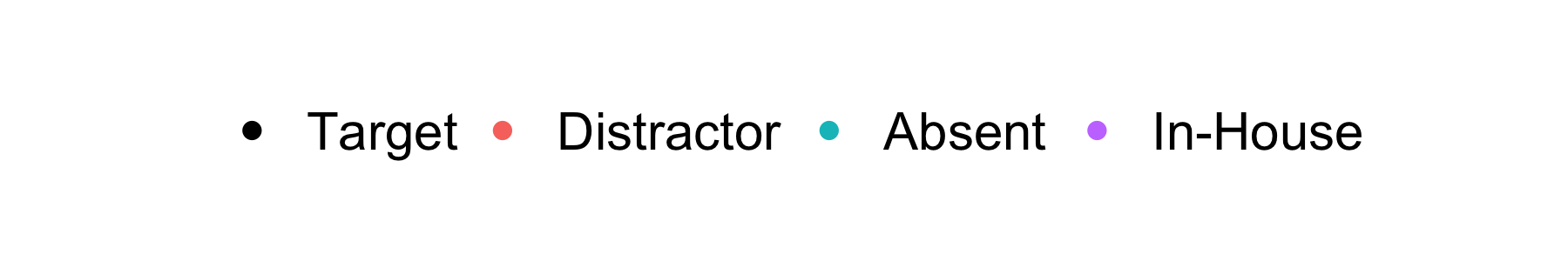}
\end{subfigure}
\caption{Responses of selected participants during the practice phase and their top 3 Viterbi paths.
\\
Panels (a), (b), and (c) show the profiles of solvers and (d) shows the profile of a non-solver. 
Strategy path probabilities were computed using individually fitted HMMs.
Target (correct) responses are placed at accuracy of 100\% whereas incorrect responses are placed at 0\%. 
Viterbi path lines are placed horizontally at positions corresponding to the probability that responses based on the strategy will be correct, e.g. successful at 100\% and prevalent digits at 50\%, ignoring strategy execution errors.}
\label{fig:p1_paths}
\end{figure}

Posterior probabilities of individual participants' strategy paths reveal the different ways that they may have learned to solve the puzzles.
Figure~\ref{fig:p1_paths} shows the 3 most likely candidate strategy transition paths for four example participants, along with their response profiles.
The uncertainty in these trajectories, particularly in the precise timing of the transition to a successful strategy as exhibited by participants in Figures~\ref{fig:p1_paths}b and \ref{fig:p1_paths}c, arises because both target and distractor responses occur with equal probability under PD strategies, and distractor responses occasionally occur as errors under successful strategies.
Even with an error-free participant (Figure~\ref{fig:p1_paths}a), the model assigns some probability to the possibility that some initial trials were lucky guesses under the PD strategy.

Given these uncertainties, we cannot know exactly when a particular solver transitioned to the successful strategy.
By marginalizing over all possible paths, however, we can attain the probability that a participant used a particular strategy at each trial $P(s_t)$, which we can use to predict when the participant began to use different strategies.
Taking the strategy with the highest posterior probability is problematic when the distribution does not strongly favor a particular strategy, for example if the posterior probabilities for all four strategies are close to 25\%.
However, what we can say with higher confidence is the probability that a participant used a strategy $s_t$ that is at least as strong as strategy $s$: $P(s_t \ge s)$.
Using this method, we can determine if a participant transitioned into strategy $s$ at trial $t$ by finding the first trial $t_x$ when $P(s_t \ge s) > 50\%$:
\begin{equation}
    t_x(s) = \argmin\limits_t P(s_t \ge s)
\end{equation} 

As shown in Figure~\ref{fig:phase1:solvers:d}, we infer that 35.2\% of the participants used a successful strategy from the first trial of the practice phase.
By the same criterion, 54.5\% of participants used a successful strategy by trial 3, 79.5\% by trial 5, and 94.3\% by trial 10.

\subsubsection{Non-solvers' strategy dynamics}

\begin{figure}[!h]
\begin{subfigure}{.5\linewidth}
  \centering
  \caption{Actual incorrect responses}
  \vspace{-6pt}
  \includegraphics[width=.95\linewidth]{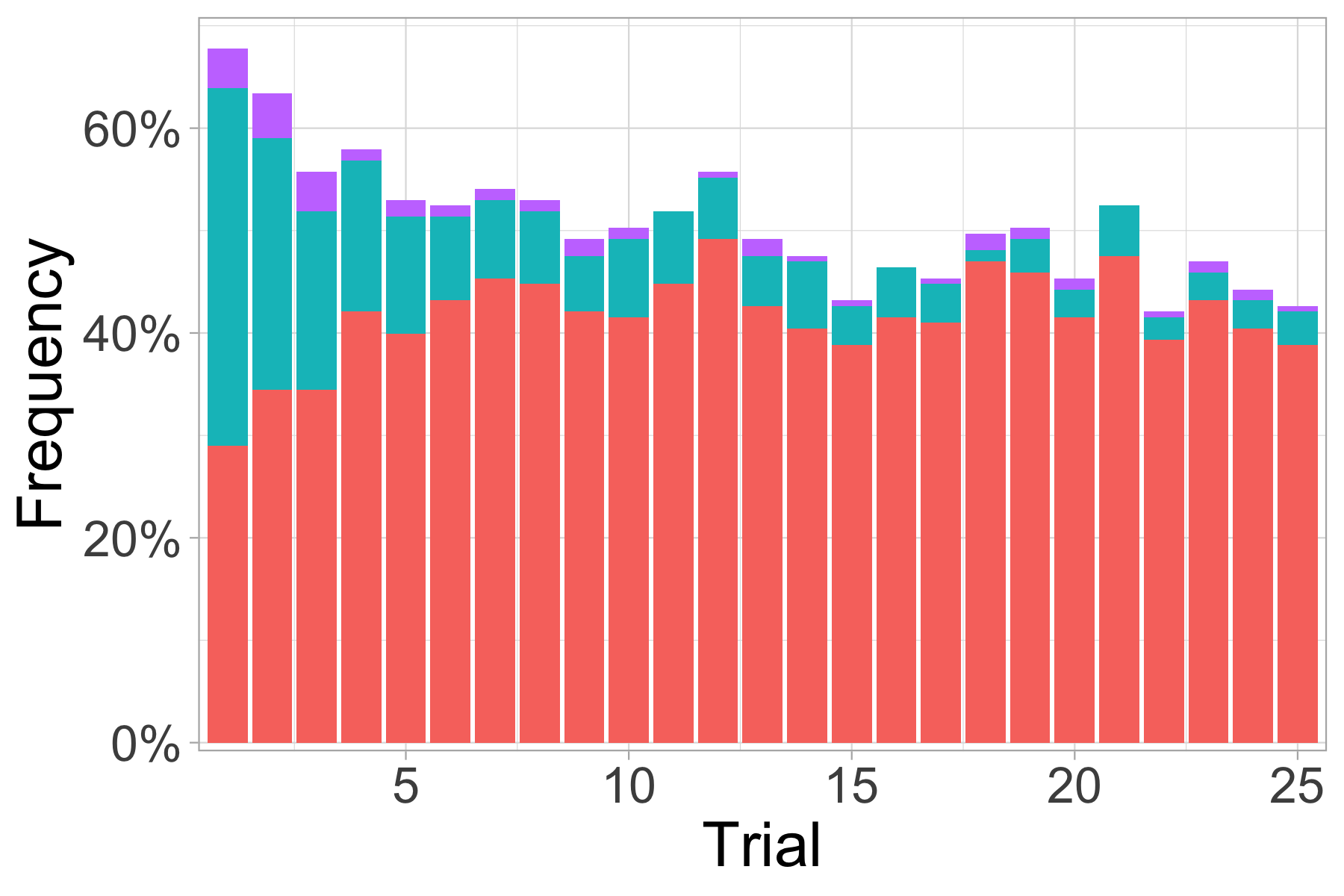}
  \label{fig:phase1:nonsolvers:a}
\end{subfigure}
\begin{subfigure}{.5\linewidth}
  \centering
  \caption{Inferred aggregate distribution of strategies}
  \vspace{-6pt}
  \includegraphics[width=.95\linewidth]{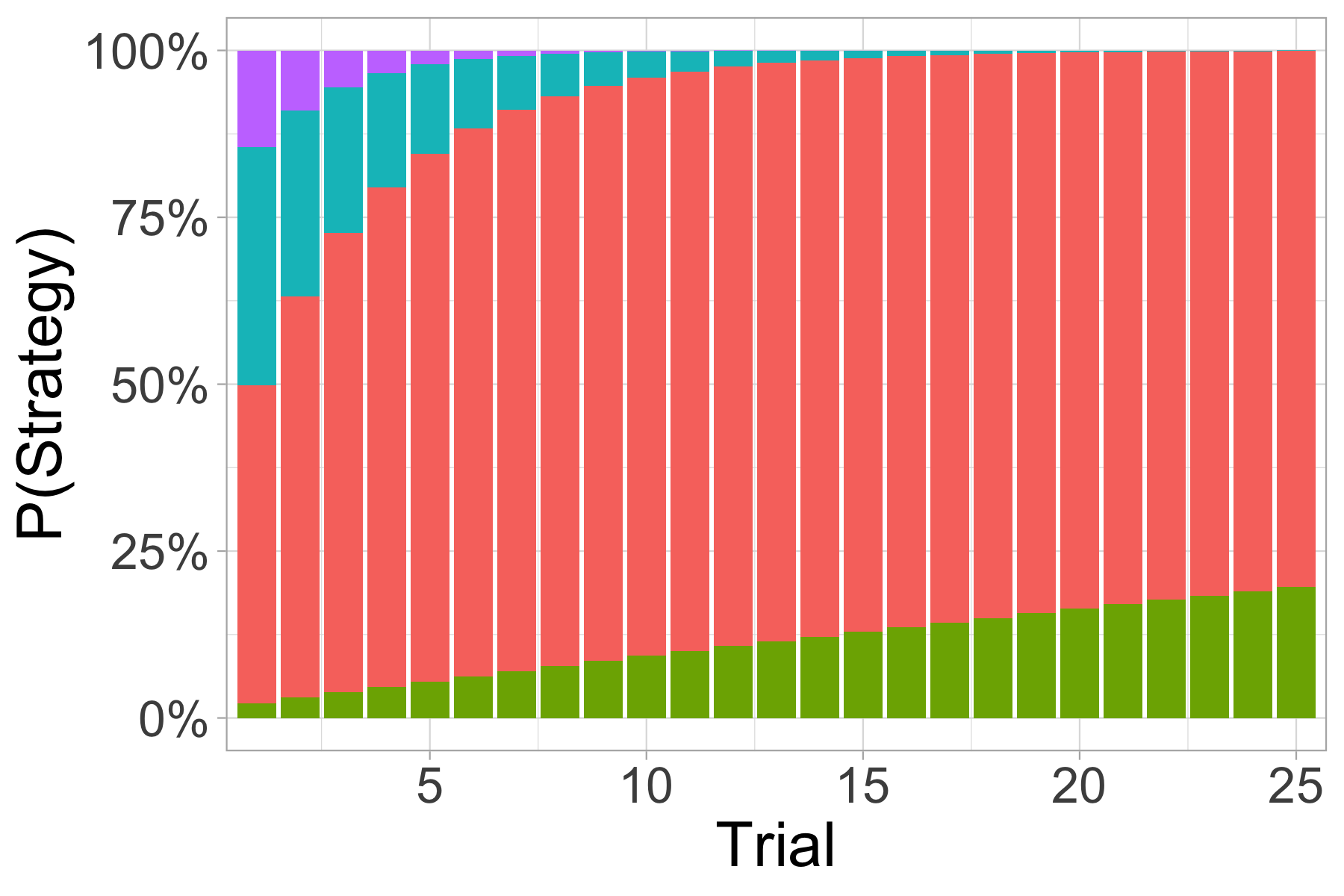}
  \label{fig:phase1:nonsolvers:b}
\end{subfigure}
\begin{subfigure}{.5\linewidth}
  \centering
  \caption{Predicted distribution of incorrect responses}
  \vspace{-6pt}
  \includegraphics[width=.95\linewidth]{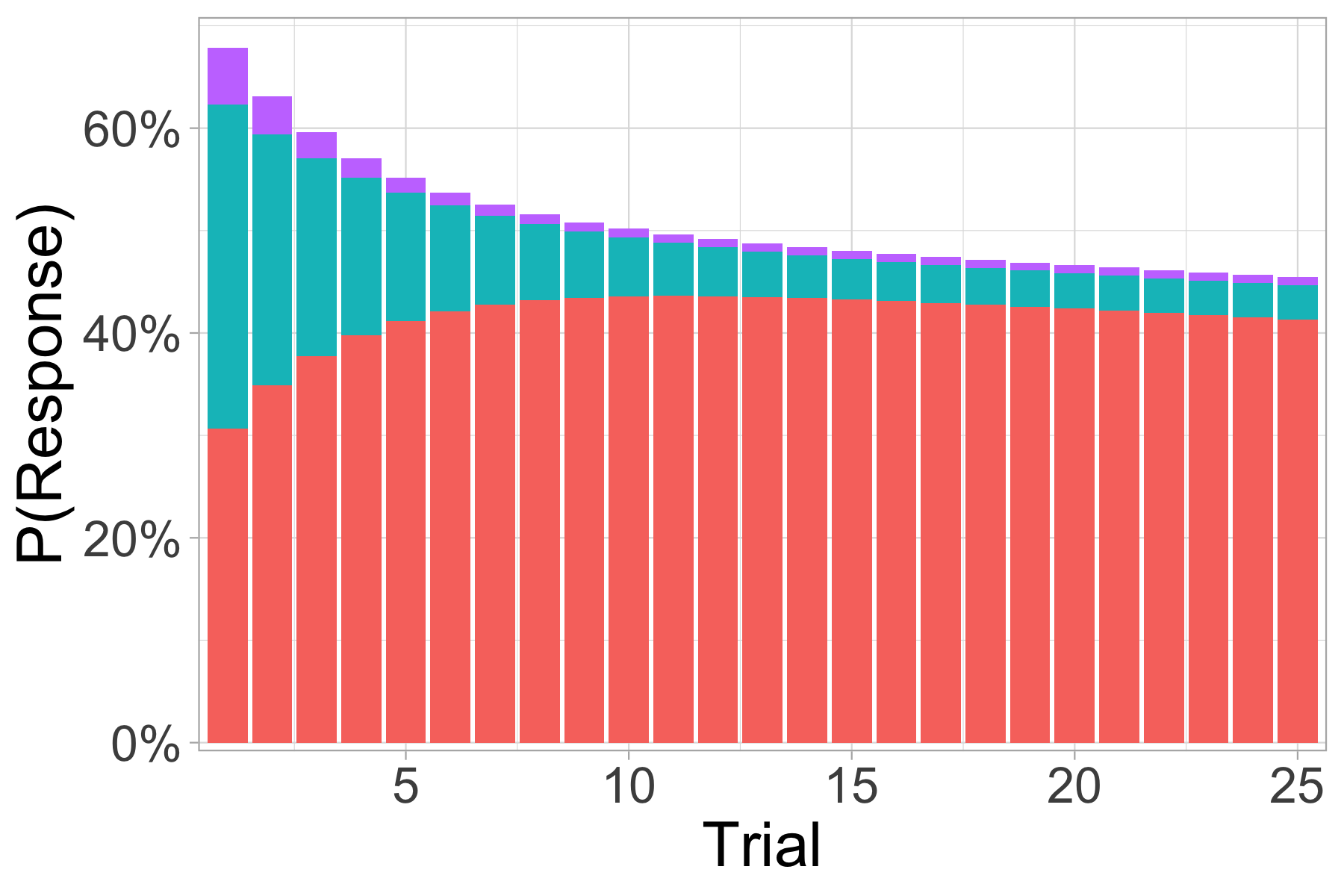}
  \label{fig:phase1:nonsolvers:c}
\end{subfigure}
\begin{subfigure}{.5\linewidth}
  \centering
  \caption{Inferred distribution of individuals' strategies}
  \vspace{-6pt}
  \includegraphics[width=.95\linewidth]{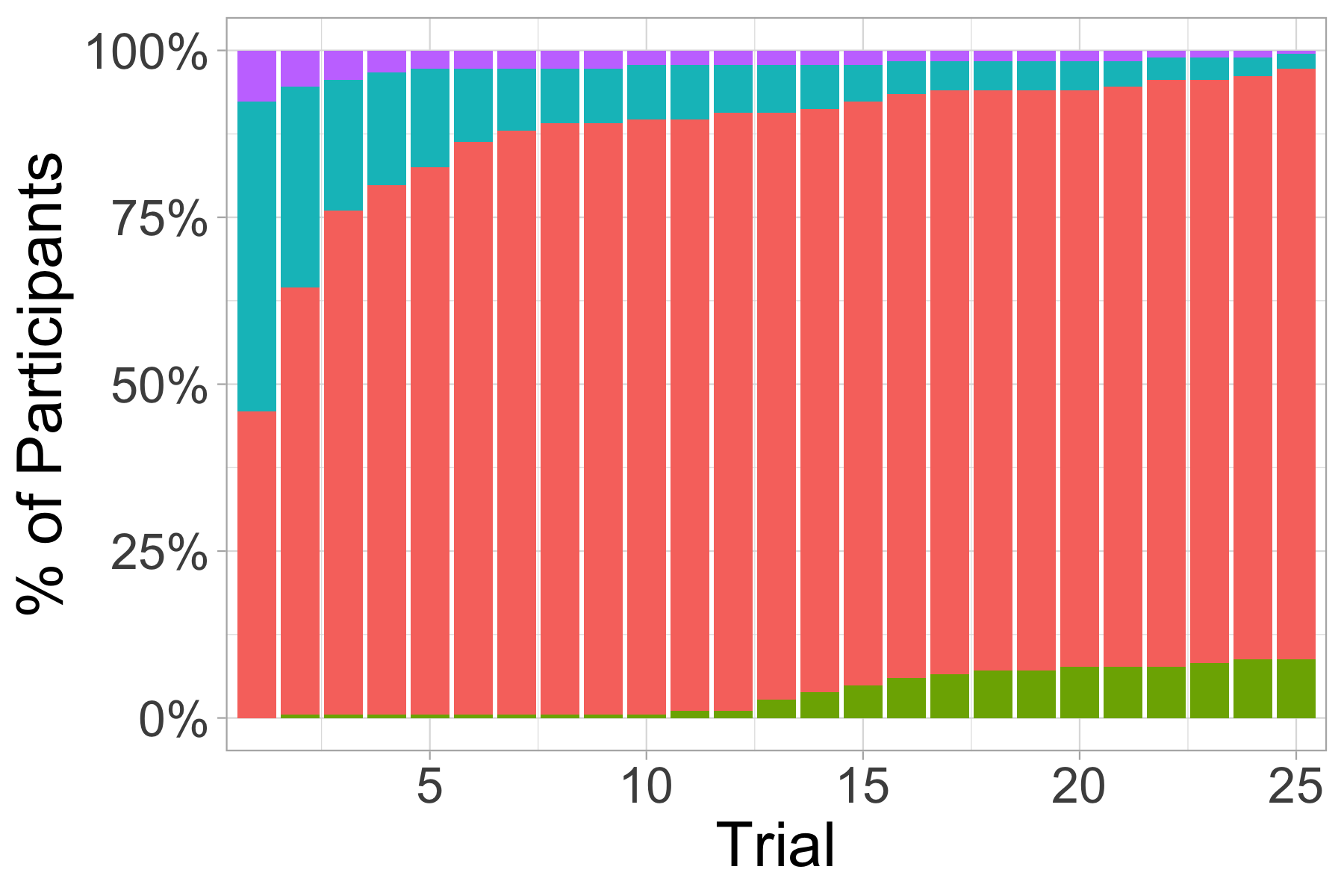}
  \label{fig:phase1:nonsolvers:d}
\end{subfigure}
\begin{subfigure}{.5\linewidth}
  \centering
  \vspace{6pt}
  \includegraphics[width=\linewidth]{figures/practice_phase/legend_responses.png}
\end{subfigure}
\begin{subfigure}{.5\linewidth}
  \centering
  \vspace{6pt}
  \includegraphics[width=\linewidth]{figures/practice_phase/legend_strategies.png}
\end{subfigure}
\caption{Response types and inferred strategies during the practice phase for participants identified as non-solver
\\
(a) Actual frequency of incorrect response types (b) Inferred distribution of strategies by the aggregate model (c) Predicted distribution of incorrect response types by the aggregate model (d) Proportion of non-solvers inferred by individually fitted models to be using a strategy at least as effective as the strategy indicated}
\label{fig:phase1:nonsolvers}
\end{figure}

Although our main focus is on the solvers, we briefly present evidence of the strategies used by the non-solvers, based on fitting a same model used for the solvers to the data of the non-solvers (see Figure~\ref{fig:phase1:nonsolvers:a}).  
Considering first the actual incorrect responses of these participants, we see that they initially exceed 50\%, and gradually approach about 50\% correct by the end of the practice phase.  
The inferred distribution of strategies from the aggregate model indicates that on the very first trial of the practice phase, 14.5\% of responses are attributed to uniform guessing, 35.7\% to avoiding direct contradictions, 47.6\% to choosing randomly between the prevalent digits, and only 2.2\% to a successful solution strategy. 
Non-solvers tend to transition away from the UG and ADC strategies, with 80.3\% of responses attributed to the PD strategy by the end of the practice phase, and smaller fractions attributable to successful (19.0\%), ADC (0.1\%) and UG (0.0\%) strategies.  

Even among responses attributable to a successful strategy, the HMM estimates an 88.0\% actual correct response rate, significantly lower than the 95.5\% of the solvers, with most of the remaining responses (9.7\%) corresponding to the distractor.  This higher rate of choices of the distractor digit make is much less certain when individuals who do transition to a successful strategy actually do so.  That said, the cumulative estimates shown in Figure~\ref{fig:phase1:nonsolvers:d} can be used as an approximate guide to determining when these transitions may have occurred.  By these estimates, most of these transitions occurred between trials 13 and 20, considerably later than nearly all of the estimated times for those in the solver group.

In summary, the bulk of the non-solvers appear to have adopted some variant of the PD strategy, a minority of participants classified as non-solvers may have acquired a successful strategy but failed to meet our preregistered solver criterion, in part because of a less reliable dependence on this strategy, and in part because they generally acquired the strategy later during the practice phase, or in some cases during the test phase.

\subsection{Solvers' generalization to out-of-distribution puzzles}
\label{sec:main:results_test}

Returning to our focus on the 88 participants we identified as solvers at the end of the practice phase, we examine whether changing the digit set, house type, and goal cell position affected their accuracy and response times across the 64 test phase trials. 
Having observed potentially significant but short-lived effects of problem variations on accuracy and reaction time at the outset of the test phase on a pilot sample, we conducted preregistered analyses on performance on the first block of 16 trials separately from those in the remaining 3 blocks of 64 trials.
The first 16 trials were chosen because that was when each participant had seen all 16 combinations of changed and unchanged puzzle features.
Overall, there was substantial transfer across all variations, as detailed below.

We used Bayesian mixed-effects models using BRMS \cite{brms} for our analyses with logit transformations for accuracy models and logarithmic transformations for response time models.
As such, we report parameter estimates with 95\% highest density credible intervals around the estimates with the parameter coefficients in logits for accuracy models and $\log_2(\text{seconds})$ for response time models. 
We note that, although house index and cell index were coded as two separate conditions for the experiment, and included as regressors in the preregistration, we chose to recode them as a single variable indicating change vs no change in the goal cell position for better interpretability (see \supplement Section~\ref{sec:supp:unreg_hi_ci}). 
Exact regression formulas, the full table of statistical measures, details of deviations from the preregistration, and regressions not reported in the main manuscript can be referenced in \supplement Sections~\ref{sec:supp:lm_reported} and \ref{sec:supp:lm_unreported}.

\subsubsection{Digit sets} 

\begin{figure}[!ht]
\begin{subfigure}{.49\linewidth}
    \centering
    \caption{Digit set accuracy}
    \label{fig:test_phase:ds_acc}
    \includegraphics[width=.95\linewidth]{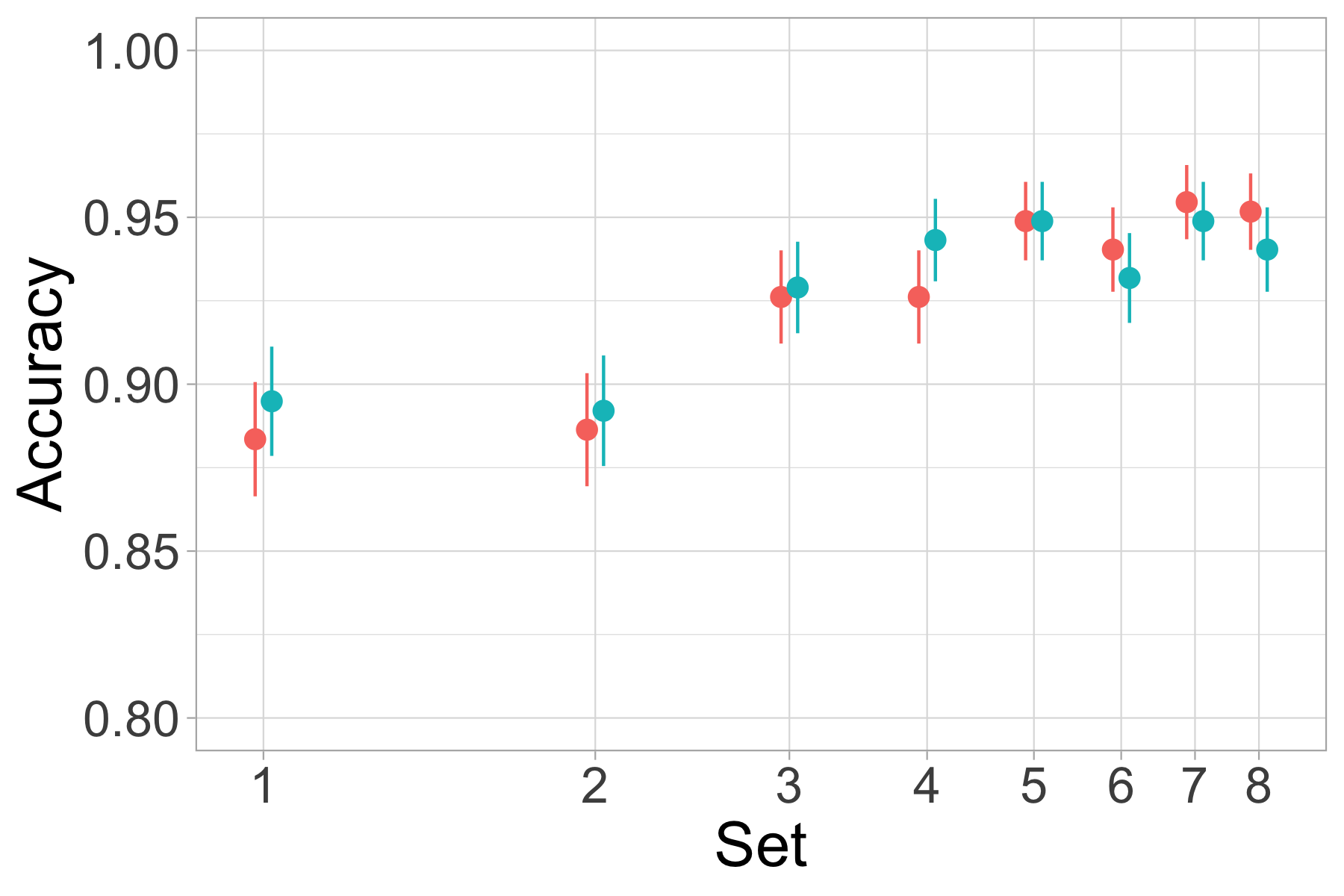}
\end{subfigure}
\begin{subfigure}{.49\linewidth}
    \centering
    \caption{Digit set response time}
    \label{fig:test_phase:ds_rt}
    \includegraphics[width=.95\linewidth]{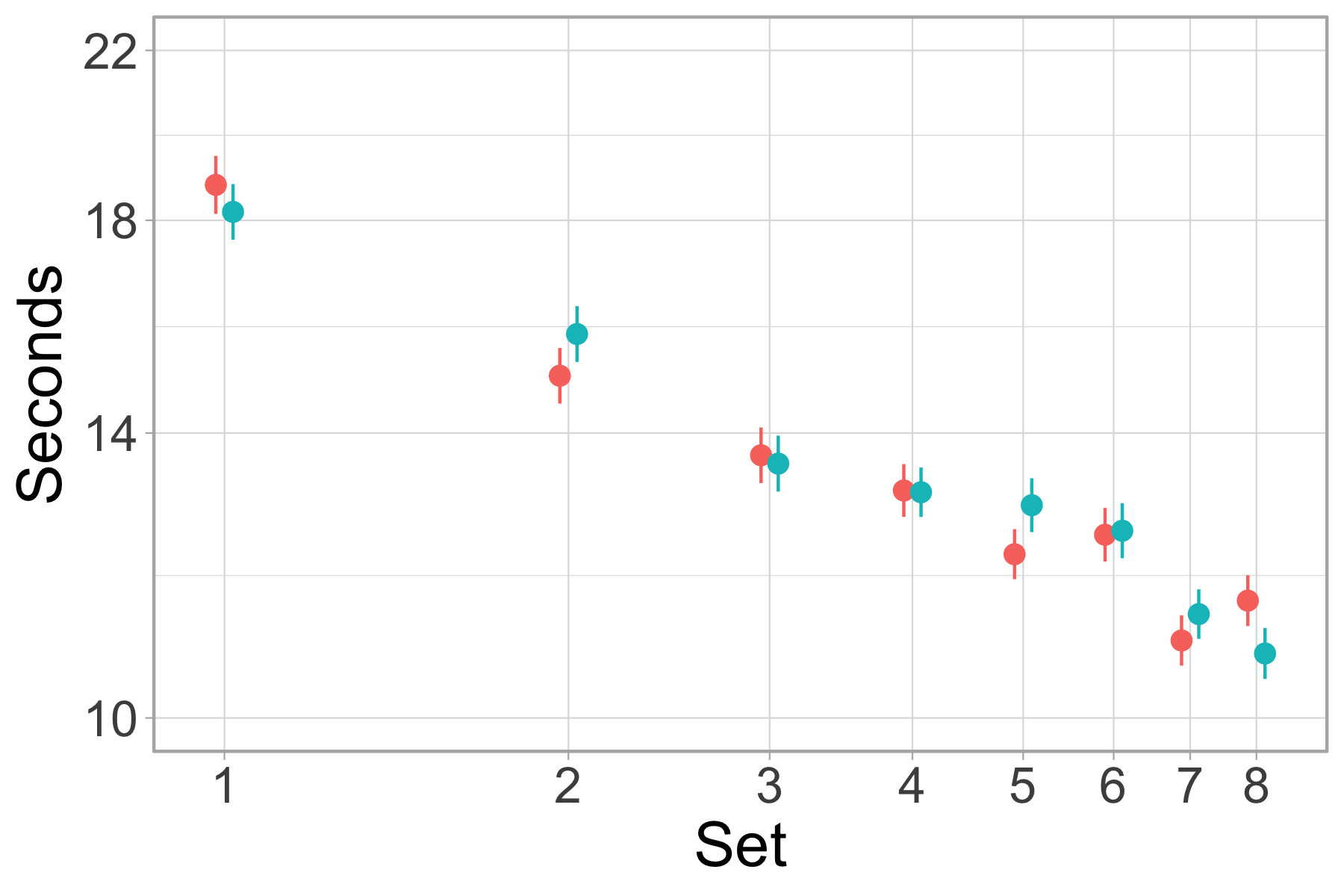}
\end{subfigure}

\begin{subfigure}{.49\linewidth}
    \centering
    \caption{House type accuracy}
    \label{fig:test_phase:ht_acc}
    \includegraphics[width=.95\linewidth]{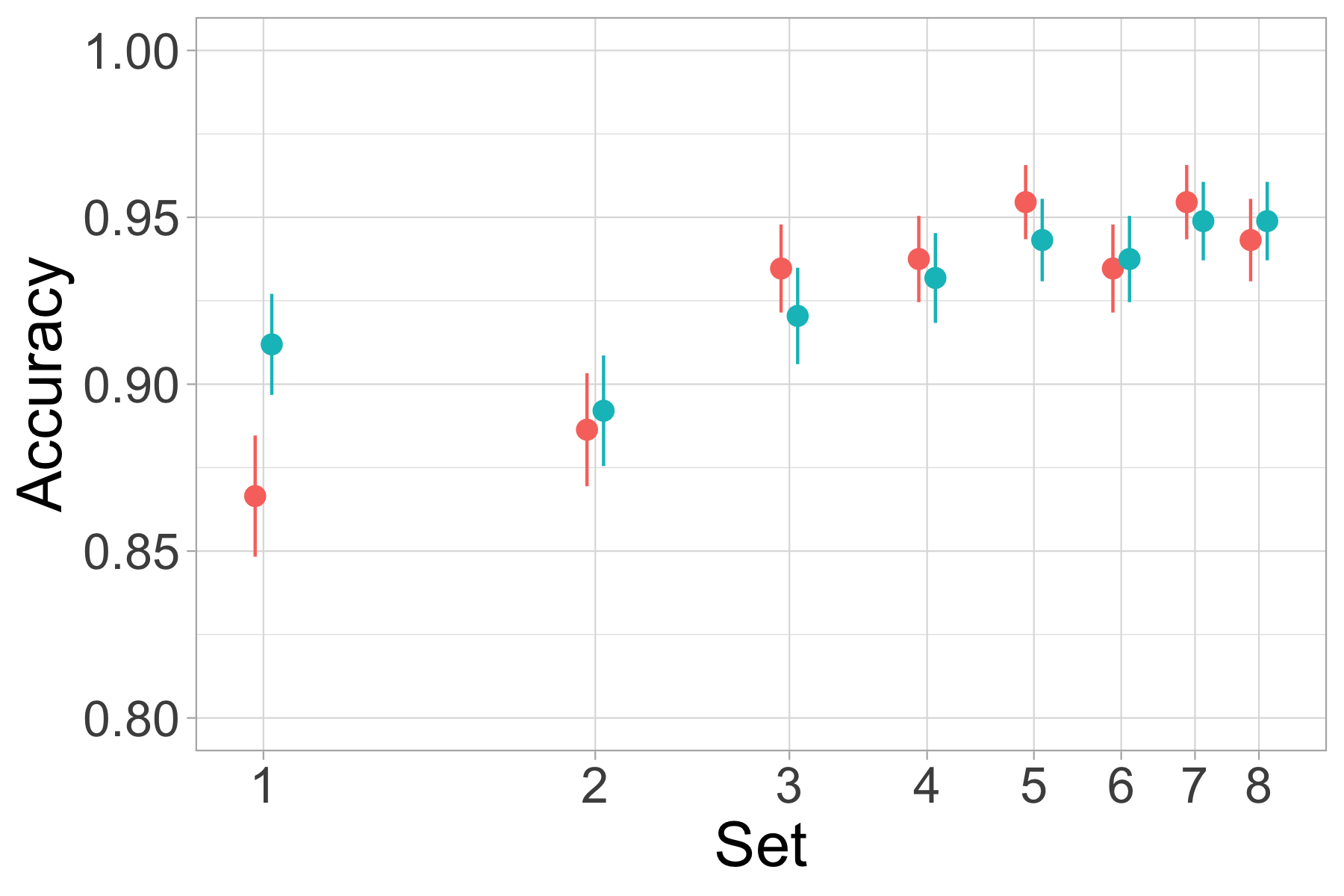}
\end{subfigure}
\begin{subfigure}{.49\linewidth}
    \centering
    \caption{House type response time}
    \label{fig:test_phase:ht_rt}
    \includegraphics[width=.95\linewidth]{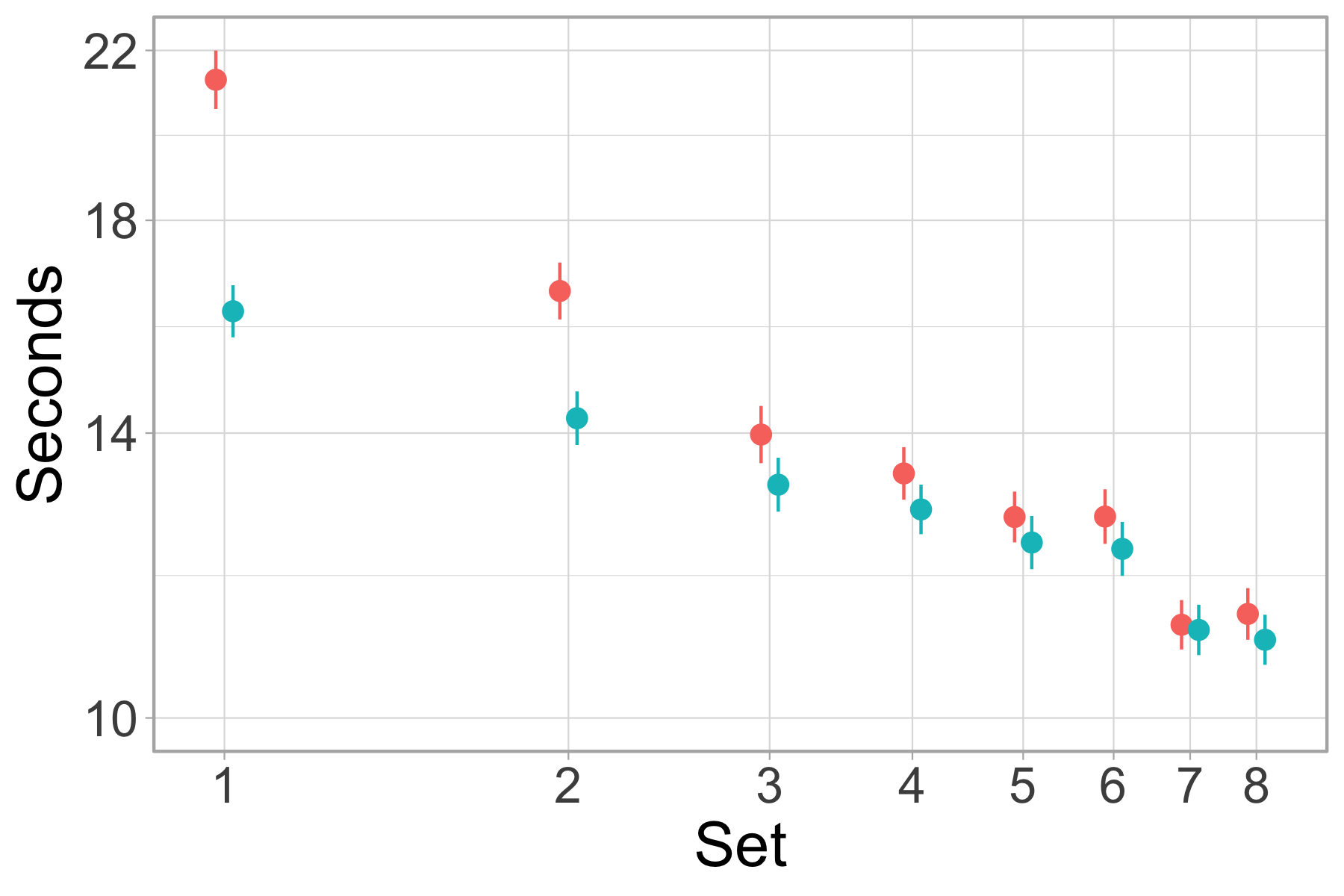}
\end{subfigure}

\begin{subfigure}{.49\linewidth}
    \centering
    \caption{Goal position accuracy}
    \label{fig:test_phase:gp_acc}
    \includegraphics[width=.95\linewidth]{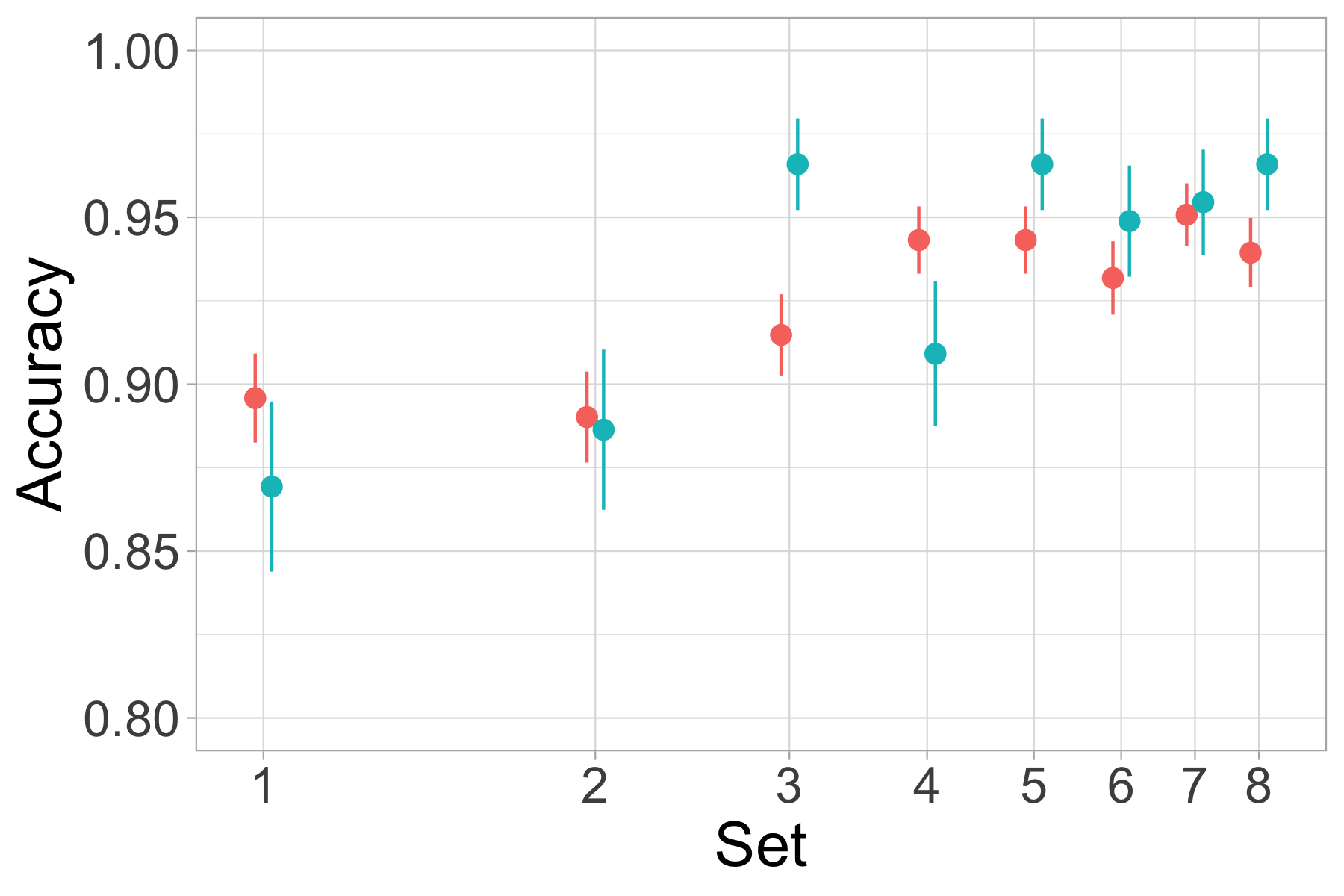}
\end{subfigure}
\begin{subfigure}{.49\linewidth}
    \centering
    \caption{Goal position response time}
    \label{fig:test_phase:gp_rt}
    \includegraphics[width=.95\linewidth]{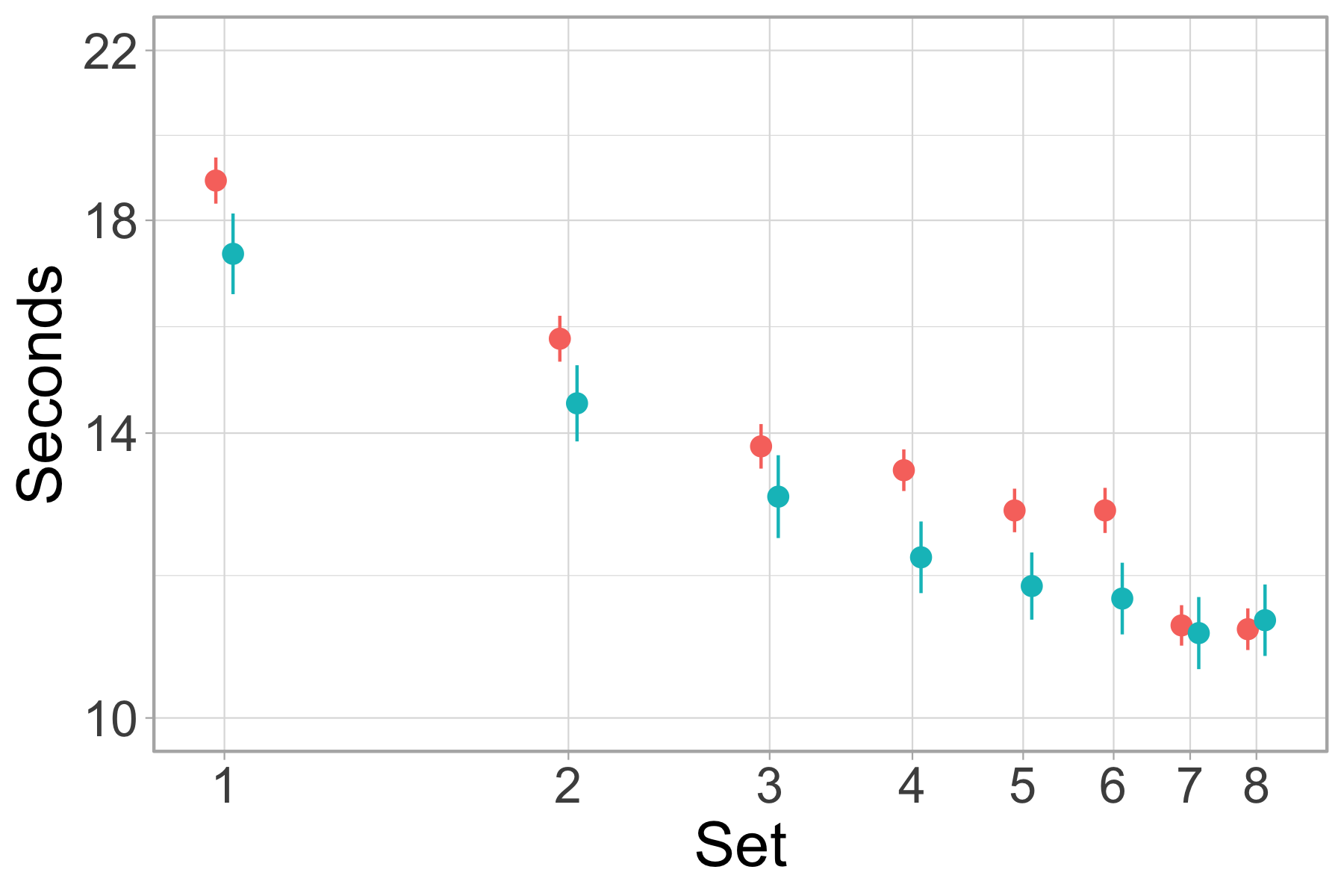}
\end{subfigure}

\begin{subfigure}{\linewidth}
    \centering
    \includegraphics[width=.5\linewidth]{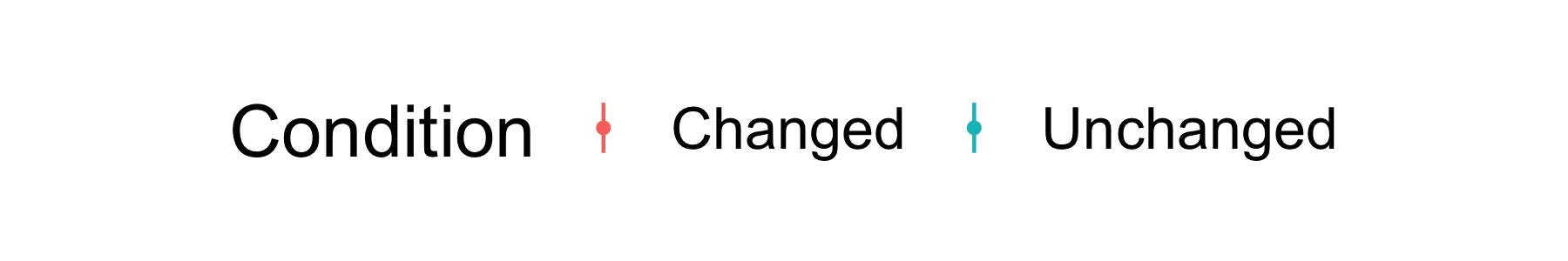}
\end{subfigure}
\caption{Average and 95\% CI of accuracy and response times.
\\
Each set represents 8 trials, e.g. set 1 contains trials 1-8 and set 2 contains 9 to 16. Note that the y-axis for the response time plots and the x-axis for all plots are in logarithmic scale.  Only trials with correct responses included for response time plots.}
\label{fig:test_phase}
\end{figure}

Solvers were able to transfer immediately when tested with target and distractor digits selected from a set never used in either of these roles during the tutorial or practice phases. 
Indeed, the effect of a change in the digit set, either for accuracy or response time, is negligible: all estimates are strikingly close to 0, falling well within the 95\% CI, as shown in Table~\ref{table:test_coeffs} and Figures~\ref{fig:test_phase:ds_acc} and \ref{fig:test_phase:ds_rt}.

\begin{table}[!h]
\begin{center}
\caption{Test phase regression coefficient estimates.}
\label{table:test_coeffs}
\begin{tabular}{ll|lll|lll}
\toprule
& & \multicolumn{3}{c}{Trials 1-16} & \multicolumn{3}{c}{Trials 17-64} \\
Term & DV & Estimate & CI-L & CI-U & Estimate & CI-L & CI-U \\
\midrule
DS & Acc. & -0.098 & -0.472 & 0.268 & 0.020 & -0.248 & 0.268 \\
\rowcolor[HTML]{EFEFEF} 
DS & RT & -0.022 & -0.086 & 0.043 & 0.000 & -0.032 & 0.032 \\
HT & Acc. & -0.302 & -0.666 & 0.059 & 0.095 & -0.170 & 0.364  \\
\rowcolor[HTML]{EFEFEF} 
HT & RT & 0.319 & 0.258 & 0.379 & 0.055 & 0.022 & 0.088 \\
GP & Acc. & 0.165 & -0.240 & 0.570 & -0.314 & -0.668 & 0.019  \\
\rowcolor[HTML]{EFEFEF} 
GP & RT & 0.115 & 0.040 & 0.189 & 0.089 & 0.050 & 0.128 \\
\bottomrule
\end{tabular}
\vspace{.5cm}
\caption*{DS: digit set; HT: house type; GP: goal cell position; CI-L and CI-H: lower and upper bounds of the 95\% credible interval. Accuracy (Acc.) coefficients are presented in logits and response time (RT) measure in coefficients are presented in log$_2$(seconds).}
\end{center}
\end{table}

\subsubsection{House type} 

Solvers were able to apply what they had learned after a switch in house type, albeit with a small initial reduction in accuracy 
and a substantial initial increase in response time as shown in Figures~\ref{fig:test_phase:ht_acc} and \ref{fig:test_phase:ht_rt}. 
Although the 95\% CI for the effect of a change in house type on accuracy includes 0, we note that about 90\% of the probability mass is below 0 and an effect of similar size with 0 falling outside the 95\% CI was obtained in pilot work reported in the preregistration of the current study \cite{nam_mcclelland_2020}.
Thus, while this small accuracy decrement is likely to be a real effect, it is noteworthy that it is small and very short-lived: the effect was more prominent in the first half of the first block of 16 trials than in the second and is numerically reversed in trials 17-64.
Response time increased by 0.319 log$_{2}$(s) in the first 16 trials, or roughly a 25\% increase from an average of 19.59 seconds to 24.44 seconds.
The effect on RT decreases rapidly, down to 0.055 log$_{2}$(s), or a 3.9\% increase from 15.04 seconds to 15.63 seconds in the last 48 trials, and the effect appears to be gone by the end.
There were no significant differences between changing from row to column and from column to row (see \supplement Section \ref{sec:supp:reg_ht0}).

\subsubsection{Goal position} 

A change in goal position did not significantly affect accuracy in the first 16 trials (0 fell well inside the CI) but produced a small decrease from 95\% to 93\% in the later 48 trials.
For response times, we find main effects of 0.115 log$_{2}$(s) (8.3\% increase) in the first 16 trials and of 0.089 log$_{2}$(s) (6.4\% increase) in the last 48 trials. The 95\% CI does not include 0 in either case. 

In interpreting the effect of goal position, we note that, throughout the test phase, 25\% of the puzzles used the same goal cell as the puzzles during the tutorial and practice phases, whereas when the goal position was changed, the goal cell would be any one of 16 cells with $\frac{1}{32}$ probability or of the remaining 64 cells with $\frac{1}{256}$ probability.
Thus, the persistent effect of goal position might result from a justifiable bias in attention toward the most common goal cell location, producing a small cost when attention must be deployed to a less likely position.
No such difference in relative likelihood applies either to the house type or the digit set variables, since both house types and both digit sets are used in the test problems with equal frequency.

\subsection{Explicitness of acquired strategies} 
Following the test phase, participants solved one last puzzle and were then asked to answer a free-response prompt: 
``Explain as clearly as possible the steps you went through to choose your answer. Please be as detailed as possible so that someone else could replicate your strategy by following your response.''
For the following analyses, we sought to ensure that the verbal reports we considered were obtained from participants whose behavioral profiles were consistent with either successful or PD guessing strategies. 
Therefore, we screened out solvers who failed to maintain a high level of accuracy throughout the test phase and non-solvers whose pattern of responses was suggestive of strategies other than PD guessing (see \supplement Section~\ref{sec:supp:qrating_subselection}). 
This left 84 participants in the group we call \textit{persistent-solvers} and, coincidentally, exactly 84 participants in the group we call \textit{PD-guessers}.

Two raters independently classified the responses of 168 participants into one of 9 categories (see \supplement Section \ref{sec:supp:qrating_options}).
The first three categories represented three \textit{valid} variations of the successful strategy that would yield the correct answer to the participant's given puzzle, based on the rules of Sudoku and the specific constraints employed in constructing the puzzles used in the experiment. 
Three other categories represented \textit{invalid} bases that would not reliably give the correct answer, such as randomly guessing between the two prevalent digits.
Two categories were used for \textit{uncertain} responses due to vague or unclear descriptions.
A last category was used for \textit{missing} or otherwise completely uninformative responses.
Although all 168 responses were rated, we focus only on the responses of participants who correctly solved the final puzzle to avoid the possibility that differences between the ratings of responses by persistent solvers and PD guessers could be attributed to the rater's awareness of the correctness of the solution, yielding 80 and 42 responses respectively for a total of 244 ratings between the two raters.

\begin{figure}[ht!]
\centering
\includegraphics[width=.7\linewidth]{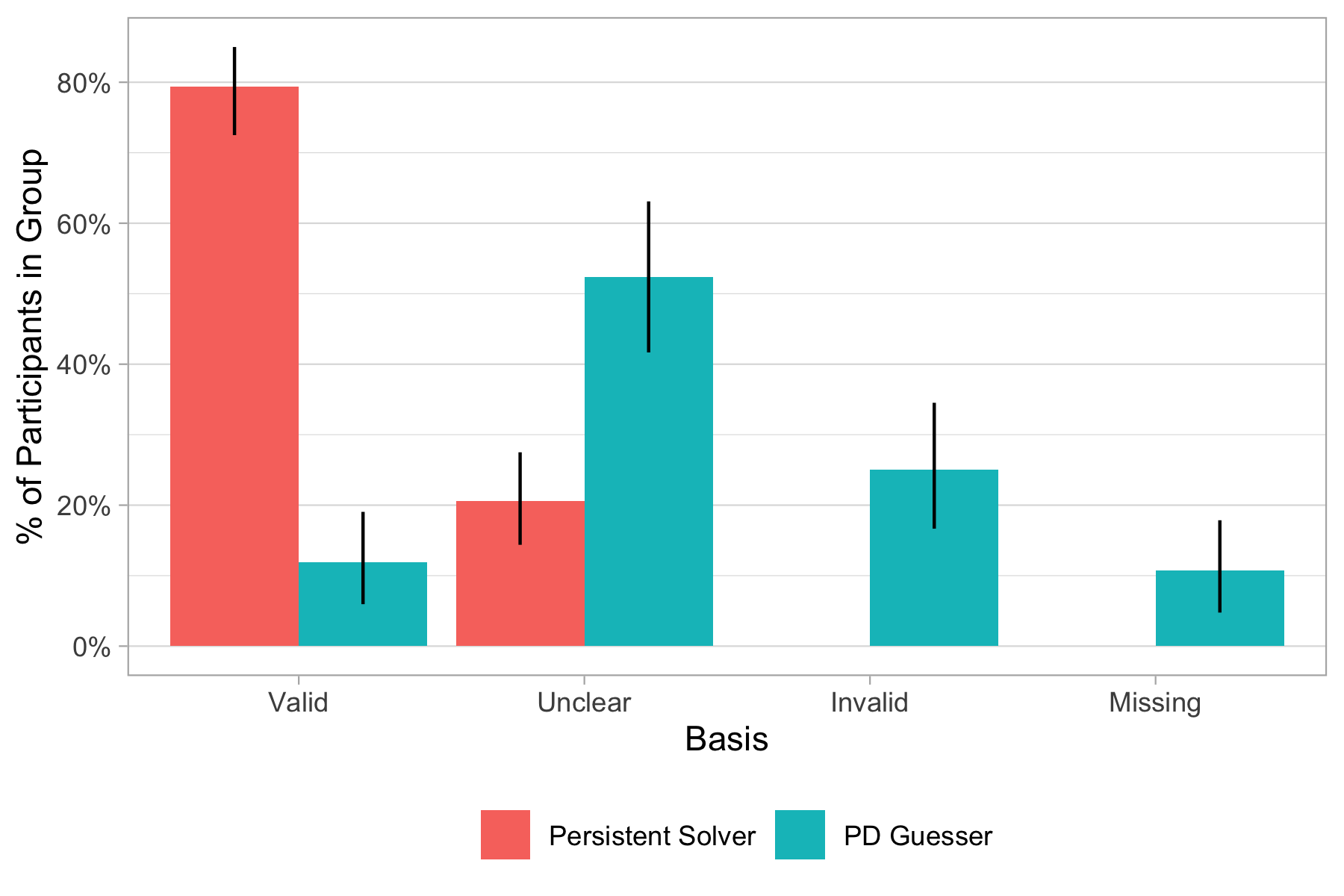}
\caption{Ratings of self-reported strategies between persistent solvers and PD guessers}
\label{fig:q_ratings}
\end{figure}

As shown in Figure~\ref{fig:q_ratings}, 79\% of persistent solvers' strategy descriptions were rated as valid compared to 12\% of PD guessers ($\chi^2$ = 99.12, df = 1, p < .001).  
Each bar represents the proportion of persistent solvers' or PD guessers' ratings (treating each rater's rating of each of the participants as a separate rating). 
The two raters' ratings had 86.9\% concordance (number of agreements divided by number of participants rated) at the superordinate level (valid, uncertain, invalid, or missing) and 69.6\% across all individual categories (see \supplement Section~\ref{sec:supp:questionnaire_rating_consistency}), with most disagreements involving an unclear rating from one of the raters.

\subsubsection{Implicit vs. explicit reasoning}
\label{sec:main:imex}

Although some persistent solvers' self-reported strategies were rated as vague or unclear, none were classified as unambiguously invalid.
While some vague or unclear responses may be due to low effort from the participants, it is also possible that some participants actually reached consistently high accuracy without being able to put their strategies into words, i.e., that the strategies they acquired were implicit \cite{reber1989implicit}.  The bulk of the solvers, however, did report valid strategies.  Based on previous findings indicating that self-explanations often produce more robust understanding and better transfer to related problems \cite{chi1989self, chi1994eliciting}, we hypothesized that explicitly learned strategies may result in higher accuracy during the test phase, in which all but 1/16th of the puzzles fell outside of the range of puzzles used during the practice phase.

We tested this hypothesis by splitting the 80 persistent solvers that correctly solved the questionnaire puzzle into two groups: 60 \textit{valid} solvers whose responses were classified as valid by both raters and 20 \textit{unclear} solvers whose responses were classified as unclear by at least one rater.
Comparing the overall performance during the test phase, valid solvers were significantly more accurate by 0.593 [95\% CI: 0.083, 1.125] logits than unclear solvers, even though the unclear solvers were significantly more accurate (0.582 [95\% CI: 0.023, 1.126] logits) during the practice phase.

\subsection{Solver status and education} 

\begin{figure}
 \centering
  \includegraphics[width=.7\linewidth]{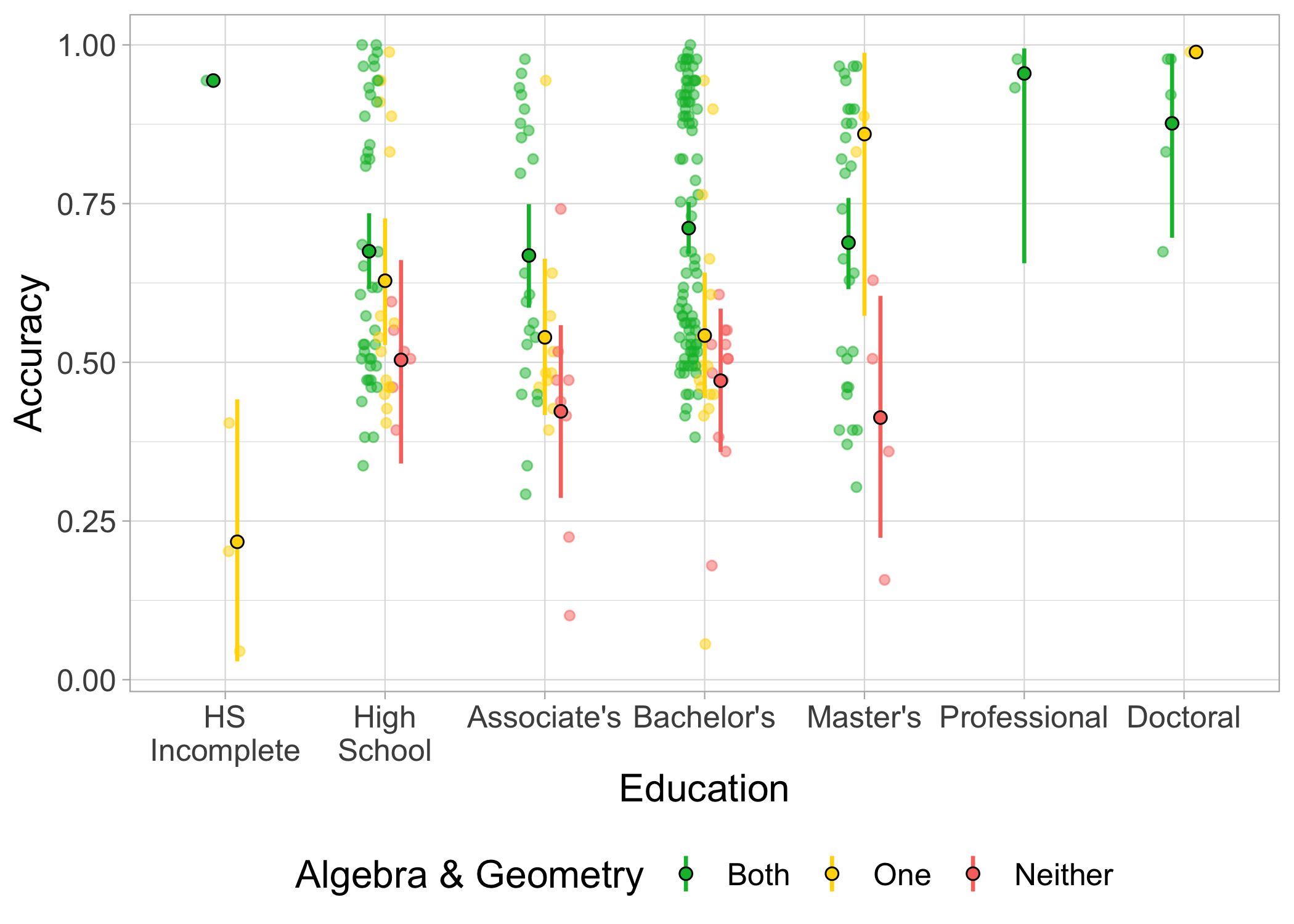}
  \label{fig:q_edu}
\caption{Overall accuracy across both the practice and test phases of the experiment by highest education.
\\
Color represents whether the participant has had formal education in both, one, or neither of algebra and geometry. Darker, bordered points represent group means and lighter points represent individual participants. 
All error bars show 95\% highest density intervals (groups with N < 2 excluded).}
\end{figure}

We next considered the relationship between education and participants' ability to solve the puzzles. 
First, using self-report data on highest level of education pursued, we fitted a regression model to predict the total number of puzzles solved across both practice and test phases, finding that the number of years of education was a small but significant predictor of puzzles solved ($\beta$ = 1.46 [0.45, 2.45], $R^2$ = 0.032).
Next, we fitted a second regression using self-report data of various math courses taken and found that of the 9 different topics, both high-school algebra ($\beta$ = 9.68 [2.87, 16.78]) and high-school geometry ($\beta$ = 9.81, [3.78, 16.03]) were significant independent predictors.  
Notably, \textit{none} of the 40 participants who reported having taken neither HS algebra nor HS geometry were solvers.
In further analyses combining years of education and math courses (see \supplement Section~\ref{sec:supp:reg_education}), we found that education predicted a small amount of independent variance when combined with HS algebra and geometry ($R^2$ = 0.168 with education, $R^2$ = 0.152 without, Bayes factor = 11.517), but its significance is lost when considered together with all of the math course variables ($R^2$ = 0.213 with education, $R^2$ = 0.206 without, Bayes factor = 2.624). 

\section{Discussion}
\label{sec:main:discussion}
Humans possess a remarkable ability to learn novel tasks and can do so rapidly and flexibly.
In this work, we explored how people learn to solve abstract reasoning puzzles, extend their knowledge to handle new puzzles with previously unseen features, and describe their strategies.
In light of the evidence presented in our work, we revisit the questions introduced at the beginning of this paper.

First, we found that the majority of the successful learners began to reliably solve the hidden single puzzles within 10 trials, some even from the first trial without any experience beyond the initial tutorial.  About 20 participants that were originally identified as non-solvers achieved high accuracies during the test phase to be what can be described as late-solvers, and some of these participants performed at or below 50\% during the practice phase of the experiment.
Thus, while people who learn the task generally do so from only a small number of examples, the distribution does appear to have a long tail that we have not captured in our study and requires further investigation.

Focusing on those who do learn the task from only a few examples, it is important to ask how they are able to do so. In this context, it is relevant to remember that neither the tutorial nor the explanatory feedback described the hidden single technique in abstract terms, nor suggest how the procedure would be applied to other puzzles.
One approach to this would be to appeal to the idea that the participants who learned rapidly relied on abstract representations capturing the entities (digits, cells, and houses) in the puzzle and the crucial relations among them (e.g. the presence or absence of an instance of a digit in a house intersecting with a specific target cell).
Such representations could have allowed them to construct a generalizable strategy from the examples described in the tutorial and presented with explanatory feedback after any errors they made within the first few practice trials.  To do so, they may have relied on prior knowledge and/or inductive biases that allowed them to construct such representations quickly.

Second, successful solvers generalized to all out-of distribution puzzles.  Strikingly, generalization from one set of digits to another produced no detectable change in accuracy or reaction time, while generalization from one house type to another produced a small initial drop in accuracy as well as a pronounced initial slowing of response time.  Considering first the generalization across digit sets, this is consistent with the idea that participants engaged with the digits in each puzzle as freely assignable to the various roles of digits within the puzzles, independently of what the identity of the digit happened to be.  
Indeed, our findings are consistent with the possibility that both solvers and non-solvers quickly came to appreciate that the two digits each occurring three times outside of the target house in each puzzle were the two they needed to consider as candidates, and could do so with equal facility whether or not they had encountered a particular digit in this role during practice.  
The initial accuracy and speed decrement we observe when switching from one house type to another, on the other hand, is consistent with the possibility that participants acquired a strategy during the practice phase that depended at least in part of whether their target house was a row or a column.  
Here, it is more difficult to determine whether participants strategies were initially abstract at an underlying level or not.  
One possibility is that their underlying representations were abstract but the switch in house type required an adjustment of the process required to map from the concrete display configuration into the relevant abstract representations.  
Another possibility is that participants were able to adapt a strategy specific to their initial house type onto a strategy that could be successfully applied to the orthogonal house type, perhaps relying on a structure-mapping or analogical reasoning process \cite{falkenhainer1989structure,gick1980analogical}.  (We also note that participants were slightly slowed by a switch of the target cell used in practice to a different cell in the grid.  As explained in results, this could reflect an acquired attentional bias favoring this cell, which remained the most likely target cell throughout the test phase of the experiment).  

Third, most persistent solvers (those who met our solver criterion based on performance in the practice phase, and maintained at least an 80\% level of performance during the test phase) provided descriptions of their solution strategies that could be identified as instances of specific valid strategies, possibly indicating that these individuals actually followed an explicit reasoning process consistent with their verbal descriptions.  Interestingly, however, a quarter of the persistent solvers provided vague or incomplete descriptions, suggesting that they may have possessed at least vaguely cohesive explanations, even if they did not articulate them fully or clearly  \cite{dejong1986explanation, ahn1992schema}.  Clarity of the strategy description was associated with better generalization and faster solution times during the test phase, where most of the puzzles required generalization in one way or another. 
This finding suggests that people can acquire a form of knowledge that supports generalization fairly well without being able to describe their procedure in a written description -- a form of knowledge that some describe as implicit \cite{reber1989implicit}.  At the same time, our findings are also consistent with other findings showing that explicit description leads to better performance on transfer tasks \cite{chi1989self}.

It is important to acknowledge that participants' own descriptions of their strategies may not accurately reflect their solution processes. For instance, some solvers may not have fully reported explicit reasoning steps that they followed, or may have reasoned explicitly but with such proficiency that their strategy descriptions suffered from the expert blind spot \cite{nathan2003expert} due to weaker memory traces \cite{ericsson1980verbal}, resulting in being misclassified as unclear solvers.  They also may have used multiple methods, which may have added to the difficulty of articulating a single strategy \cite{ericsson2017protocol}.  Conversely, solvers could have learned and generalized without formulating explicit strategies and might have formulated such strategies upon being prompted, possibly even believing they had followed them all along \cite{nisbett1977telling}.
Further research is needed to understand precisely how people engage in implicit and explicit abstract reasoning, and what differences arise as a consequence. 

Lastly, we found that out of 271 participants, only 88, or roughly a third, of them could learn to solve the hidden single puzzles rapidly and reliably enough to meet our criterion for classifying them as solvers, despite the extensive tutorial and feedback provided.
There was a strong association between high-school mathematics education and solver status, and none of the participants lacking both high-school algebra and geometry backgrounds met our criteria for classification as solvers.  How should be think about this relationship?

As discussed in the \textit{Introduction}, there is an important tradition that proposes that humans rely on built-in systems that support rapid learning and systematic reasoning \cite{fodor1988connectionism, lake2017building}, such as core systems for numbers \cite{feigenson2004core} and universal grammar \cite{chomsky1968language}.
Our finding that the ability to acquire one systematic reasoning strategy -- the hidden single strategy -- is limited to a subset of adult participants, all of whom had at least some prior exposure to high-school level mathematics calls this idea into question.  As an alternative to this perspective, our findings raise the possibility that the effect of education is due to the practice of learning through instructions including explanatory discourse and through justifying one's responses with explanations in exams, assignments, and classroom discussions.
Indeed, many have argued that systematic reasoning ability exhibited by solvers is a consequence of formal schooling \cite{scribner1973cognitive, vygotsky1934thought, burger1986characterizing}.  While variation in proclivity for explicit reasoning based on factors other than experience with explicit reasoning tasks might also play a role, we believe it will be important to continue to explore the possibility that systematic reasoning and generalization are acquired abilities that depends on relevant experience.

In sum, we have found that a subset of our participants robustly acquired the hidden single strategy from our brief explanatory tutorial and practice with feedback; that these participants were successful in generalizing to out-of-distribution puzzles; that most of these participants articulated valid strategies; and that success in acquiring the hidden single strategy was associated with mathematics education.  These findings are consistent with the idea
that most successful participants engaged in structured, at least partially abstract, and often explicit reasoning and self-explanation, and that acquisition of these abilities is at least in part experience dependent. With this idea in consideration, we now turn our attention to its implications for computationally modeling systematic and algorithmic reasoning in modern machine learning-based systems.

\subsection{Implications for computational models of intelligence}
Neural network models have come a long way since the early criticisms of their limited capacity for systematicity and compositionality \cite{fodor1988connectionism, marcus2003algebraic}.
They are showing increasing proficiency in solving mathematics problems \cite{cobbe2021training, hendrycksmath2021, lewkowycz2022solving, uesato2022solving}, reason analogically \cite{webb2022emergent}, and generate images compositionally \cite{ramesh2022hierarchical}.
Moreover, language-based models can even exhibit biases and errors similar to those of humans \cite{webb2022emergent, kumar2022using, dasgupta2022language, binz2022using}, suggesting that they may not only be human-like when systematic, but also when unsystematic.

Yet despite these accomplishments, our results highlight several key differences between modern deep learning models and human intelligence. First, a persistent challenge for deep learning models is the amount of data required to train them, which can be orders of magnitude more data than what humans require \cite{tsividis2017human, lake2017building}.
Second, neural networks often struggle to generalize to samples outside their training distribution \cite{lake2018generalization, bahdanau2018systematic, hupkes2020compositionality} whereas humans can often do so with relative ease.
Lastly, neural networks do not readily yield interpretable results, sometimes using features imperceptible or unintuitive to humans for decision making \cite{szegedy2013intriguing, geirhos2018imagenet, su2019one}, requiring sophisticated tools to extract decision criteria \cite{guidotti2018survey}.

Two alternative bases for rapid learning and robust generalization are worth noting here, both due to their empirical successes in addressing some of these limitations and their popularity in recent years.
The first approach is to design models with domain-specific inductive biases \cite{lake2017building, battaglia2018relational} with architectural constraints that match the systematicity inherent to the task \cite{bahdanau2018systematic, hupkes2020compositionality}, implementing the structured arrangements of thoughts \cite{fodor1975language, marcus2003algebraic,rumelhart1980schemata}
that enable systematicity and compositionality through roles and fillers.
Highly structured tasks such as Sudoku strongly benefit from this approach by representing individual cells as nodes of a graph with edges shared among all cells in the same house (row, column, or box), as in the recurrent relational network (RRN) \cite{palm2018recurrent}, a graph neural network that can learn to solve hard Sudoku puzzles far beyond the skill level of many experienced Sudoku players.
Since each cell is represented explicitly, it is also possible to track which cells contribute to the solution of a given cell, offering greater visibility and interpretability of the model's solution processes.
However, although building in the exact connectivity required to capture the specific constraints of a particular problem can improve sample efficiency and generalizability, it also faces the risk of failing to generalize for constraints not explicitly built in.
As a case-in-point, when adapted to our hidden single puzzles, the RRN model generalizes flawlessly to held-out house types and goal positions but fails catastrophically on held-out digits (see \supplement Section~\ref{sec:supp:rrn}), since only positional and not digit-identity constraints were built in by the designers of the model.

The second approach is to pre-train models with large quantities of related tasks for the purposes of adapting \cite{hospedales2020meta,finn2017model,wang2016learning}, fine-tuning \cite{cobbe2021training, hendrycksmath2021}, or using in-context learning \cite{brown2020language, wei2022chain} for downstream tasks, imitating the rich knowledge and priors people build upon when learning novel concepts \cite{bruner1961act, lake2017building}. 
However, even when using pre-trained foundation models such as GPT-3, successful fine-tuning to downstream tasks such as mathematics can require thousands \cite{cobbe2021training, uesato2022solving} to millions \cite{hendrycksmath2021} of additional samples.
While chain-of-thought prompting and in-context learning \cite{brown2020language,wei2022chain, wang2022self, zhou2022least, suzgun2022challenging} may seem more promising given the low sample requirements, it is unclear whether performance improvements using these approaches arise from acquisition of genuinely novel knowledge, better retrieval of existing knowledge, or mere superficial adaptations to the task format \cite{webson2021prompt, min2022rethinking}.  
Taken together, then, these approaches still leave us without an approach that can demonstrably capture rapid human learning from a narrow range of examples with broad out-of-sample generalization, except in cases where the basis for generalization is explicitly built in.

We believe that simply asking whether a machine can imitate human-like responses as originally proposed by Turing \cite{turing1950computing} is no longer an adequate measure for evaluating computational models of intelligence.
Echoing the succession of behaviorism \cite{watson1913psychology, skinner1974behaviorism} by the study of mental processes \cite{chomsky1957syntactic, neisser1967cognitive} in the cognitive revolution, 
We see it as imperative to begin considering not only whether computational models can produce correct answers to large batteries of problems \cite{srivastava2022beyond}, but also whether they can do so with the amount of experience and capacity to generalize that characterize human reasoning.  We theorize that, at least in abstract reasoning tasks, human-like behavior is often a consequence of human-like thought, which we have argued depends at least in part of hypothesis formation and explanation-based reasoning.  

To be clear, both of the approaches we have described above will likely be important contributions towards models with human-level intelligence.  Some architectural constraints will likely be necessary to equip neural network models with sufficient expressivity, and large, diverse data sets will  provide the experience and knowledge to handle tasks of varying complexity and diversity.  Even so, we suggest that addressing these limitations may depend on creating models with the capacity to learn and reason explicitly through hypotheses and explanations \cite{rumelhart1980schemata, ahn1992schema, chi1989self, lombrozo2006structure, pearl2019seven}. 
People, including children \cite{berk1994children, kelemen1999rocks}, naturally engage in explanatory behavior in any situation, and this tendency is so strong that people confabulate and believe explanations that are not true \cite{nisbett1977telling, rozenblit2002misunderstood, cushman2020rationalization}.  However, explanations provide conceptual coherence \cite{murphy1985role}, identify gaps in knowledge \cite{keil2006explanation, chi1989self}, elicit counterfactual reasoning \cite{lewis1974causation, halpern2005causes}, constrain under-determined problems \cite{lombrozo2006structure}, assign informational utility \cite{lombrozo2006functional}, assign blame \cite{keil2006explanation}, and consider other agents' intentions \cite{kelemen1999function}. Thus, we argue that incorporating this core feature of human intelligence will play an important role in capturing human intelligence in neural networks.

As deep learning practitioners have begun to take greater interest in reasoning problems, research on explanation-based models has also begun to emerge. For instance, models trained using explanations \cite{mishra2022towards}, whether from scratch \cite{camburu2018snli, lampinen2022tell}, fine-tuned \cite{lampinen2022can}, or conditioned with in-context prompts at evaluation time \cite{wei2022chain, lu2022learn}, have been shown to outperform models without explicit explanations. However, much of the existing literature remains largely empirical with limited theoretic accounts for the phenomenon \cite{xie2021explanation}, successes to date with these approaches have been modest \cite{lampinen2022tell,lampinen2022can}, and further research will be required to explore the multifaceted properties of explanation-based learning.   

As of this writing in late March 2023, recent instances of OpenAI's GPT models may be beginning to exhibit some ability to produce explanations, and even to use them to guide their behavior.  Their ability to do so may depend on the fact that these models contain instances of explanatory discourse in their training data and are fine tuned with structured, explanation-based dialog, sometimes provided by human experts \cite{ouyang2022training}.   Our experiments with the GPT-3 variant \textit{text-davinci-003} on our Hidden Singles task provide tantalizing hints that these models may have some ability to engage with explanations, but still fall short of fully integrating them into their actual problem solving behavior (see \supplement Section~\ref{sec:supp:gpt3}).  Interestingly, we find that the model can provide a correct 
explanation of the hidden single technique when asked, but achieves only 70\% accuracy in solving our hidden single problems even after extensive chain-of-thought prompting.  Taking this together with the fact that the huge corpus used to train these models likely contains quite a lot of Sudoku-related content, it is uncertain whether current iterations of large language models are capable of learning and reasoning with explanations in novel domains, and whether explanations would even result in improvements comparable to those of human solvers.

\section{Conclusion}
Our findings contribute both to understanding how humans learn and generalize abstract strategies for novel tasks and to identifying where humans and contemporary models diverge in these respects, offering insight on what behavior we ought to expect from a model that reasons as humans do beyond common measures of accuracy.
Training models to hypothesize and explain would be an important step towards what we believe will begin to truly rival human reasoning, but the multifaceted advantages of explanation-based learning remain yet to be explored in computational models.
Likewise, many of the attributes of human reasoning that we have observed in our study remain incompletely understood.
Continued effort to understand these human processes and to capture them in computational models offer exciting directions for future research.

\clearpage
\bibliography{main}
\bibliographystyle{unsrt}

\clearpage
\begin{center}
    \LARGE Supplementary Materials
\end{center}
\tableofcontents

\unhidefromtoc
\clearpage

\section{Experiment design}
\label{sec:supp:exp_design}

We describe important details about the experiment in this section, but the full experiment may be found in the GitHub repository. Although the full code for the experiment is available, we also provide a PDF of screenshots in \verb+figures/screenshots.pdf+. Note that the screenshots were generated for one possible participant and what might be shown at the start of each screen. The screenshots were taken within a development environment and contain artifacts that would not have appeared for actual participants, such as the accumulated compensation always being \$0.00, the screen number, and "You solved out of 89 puzzles" on the final screen which should contain an actual number of puzzles solved. Some screens had short pop-up messages during the tutorial in response to participants' actions that are not depicted in the screenshots. Screens with the blue \verb+Submit+ button had a required task and most would not allow the participant to proceed unless they provided valid responses. Although the PDF shows screenshots for every screen in the experiment, not all screens were shown to all participants depending on their responses. We describe the control flow below.

\begin{enumerate}
    \item Screen 2 (diagnostic survey): if the participant either correctly solved the diagnostic puzzle in Screen 1 and/or did not respond with "None" to the question "About how many Sudoku puzzles have you successfully completed?", skip to Screen 120.
    \item Screen 5 (agreement): participants were allowed to proceed if their responses matched the prompt by at least 85\%.
    \item Screens 6-14 (tutorial): participants could only proceed after submitting correct responses (where relevant). Hints were provided through short pop-up messages if they submitted incorrect responses.
    \item Screens 18-42 (practice phase): participants could only proceed after inputting the correct digit in the goal cells.
    \item Screens 44-107 (test phase): if participants correctly solved a puzzle, they were allowed up to 10 seconds before automatically proceeding to the next screen (this period may be skipped). If they submitted an incorrect response (including a blank) or did not submit anything for 2 minutes, they were required to wait 10 seconds before automatically proceeding to the next screen.
    \item Screen 108 (questionnaire attention check): participants could only proceed after answering all 3 questions, regardless of correctness.
    \item Screens 109-119 (questionnaire): see Section \ref{sec:questionnaire}
\end{enumerate}

The experiment program was implemented using Facebook's React, hosted on Amazon AWS, and deployed using Psiturk \cite{eargle_david_2021_4606854, gureckis2016psiturk}.

\subsection{Diagnostic test and survey}
\label{sec:supp:diagnostics}

At the beginning of the experiment, in order to filter out any participants who had prior experience in solving Sudoku puzzles, we presented a simple diagnostic test and a survey. Participants were not informed about the purpose of the diagnostic material.

\begin{figure}
    \centering
    \includegraphics[width=.5\linewidth]{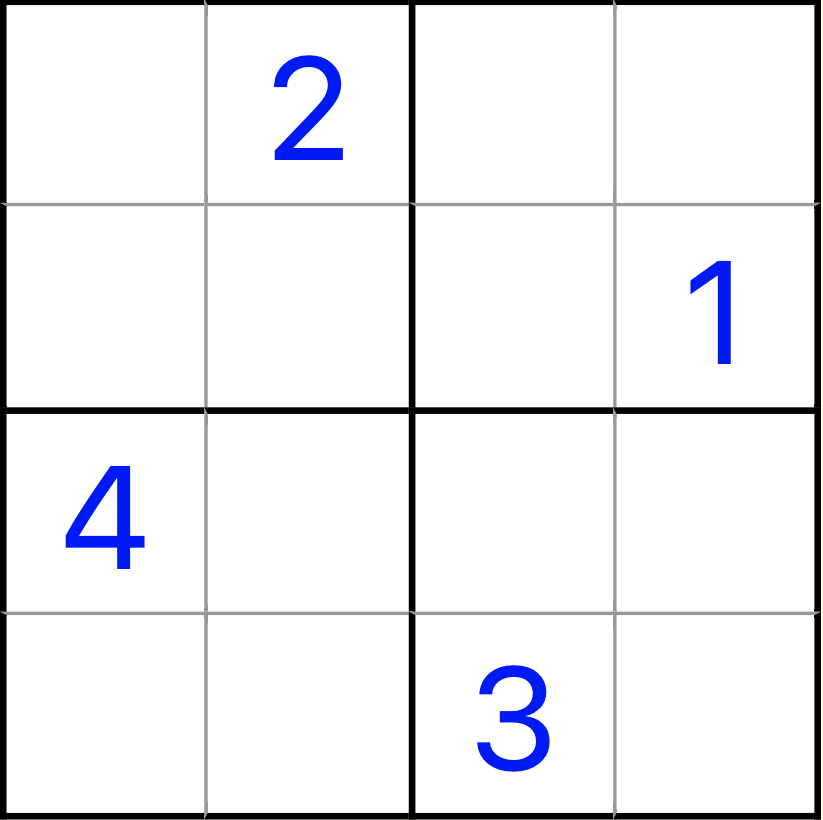}
    \caption{Diagnostic Puzzle}
    \label{fig:diagnostic_puzzle}
\end{figure}

The diagnostic puzzle was a 4x4 Sudoku grid as shown in Figure \ref{fig:diagnostic_puzzle}.
Participants were told that the puzzle was a 4x4 variant of Sudoku and were instructed in how to interact with the program interface.
They were offered \$0.25 to complete the puzzle with a \$0.01 penalty for each incorrect attempt. All cells needed to be correct for the puzzle to be considered solved.

After the diagnostic puzzle, participants were presented with the following survey:

\begin{enumerate}
    \item Have you heard of Sudoku before?
    \begin{enumerate}
        \item Yes
        \item No
        \item Not sure
    \end{enumerate}
    \item Have you ever attempted to solve a Sudoku puzzle?
    \begin{enumerate}
        \item Yes
        \item No
        \item Not sure
    \end{enumerate}
    \item About how many Sudoku puzzles have you successfully completed?
    \begin{enumerate}
        \item None
        \item 1 to 3
        \item 4 to 6
        \item 7 to 9
        \item 10 or more
    \end{enumerate}
\end{enumerate}

Only participants that did not solve the diagnostic puzzle and also responded that they had never successfully completed any Sudoku puzzles were proceeded to the rest of the experiment.
For all others, the program skipped to the demographics survey and terminated thereafter.
Of the 1,985 people that originally entered the study, 1,384 successfully solved the diagnostic puzzle and 1,668 responded that they had completed one Sudoku puzzle in the past, leaving 271 participants in neither group to complete the experiment.

\subsection{Tutorial}
\label{sec:supp:tutorial}

At the start of the experiment, participants were given a brief description of Sudoku with the following sentence: "Sudoku is a puzzle with a 9x9 grid of numbers where each row, column, and 3x3 box must contain exactly one of each number from 1 to 9." Beyond this initial statement, we took as much care as we could to avoid terms referring to variables and roles. For exact phrasing used in the experiment, please refer to the experiment screenshots in the online GitHub repository.

\begin{figure}[!ht]
\begin{subfigure}{.245\linewidth}
  \centering
  \includegraphics[width=\linewidth]{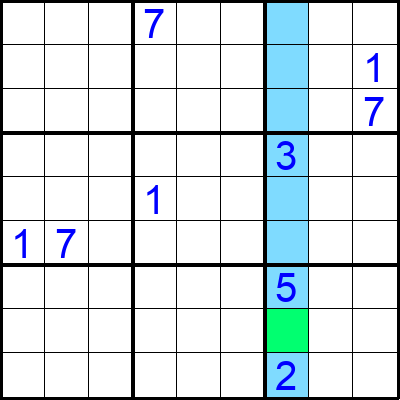}
  \caption{}
  \label{fig:tut_puzzle}
\end{subfigure}
\begin{subfigure}{.245\linewidth}
  \centering
  \includegraphics[width=\linewidth]{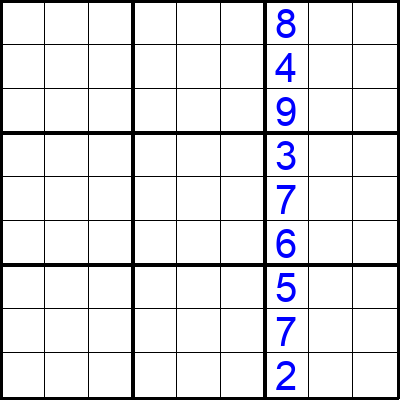}
  \caption{}
  \label{fig:tut_contradiction}
\end{subfigure}
\begin{subfigure}{.245\linewidth}
  \centering
  \includegraphics[width=\linewidth]{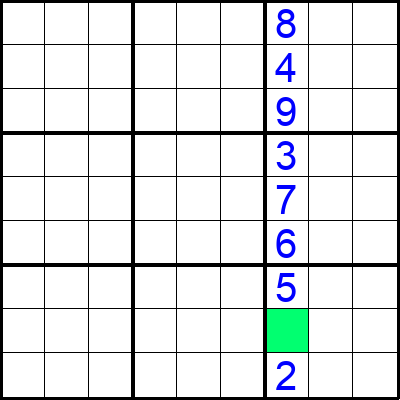}
  \caption{}
  \label{fig:tut_fullhouse}
\end{subfigure}
\begin{subfigure}{.245\linewidth}
  \centering
  \includegraphics[width=\linewidth]{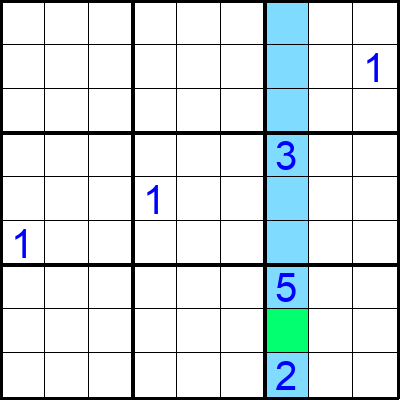}
  \caption{}
  \label{fig:tut_hs1}
\end{subfigure}
\begin{subfigure}{.245\linewidth}
  \centering
  \includegraphics[width=\linewidth]{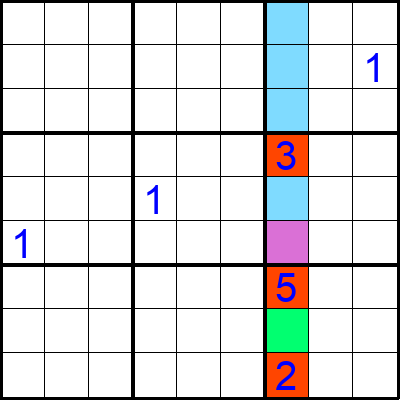}
  \caption{}
  \label{fig:tut_hs2}
\end{subfigure}
\begin{subfigure}{.245\linewidth}
  \centering
  \includegraphics[width=\linewidth]{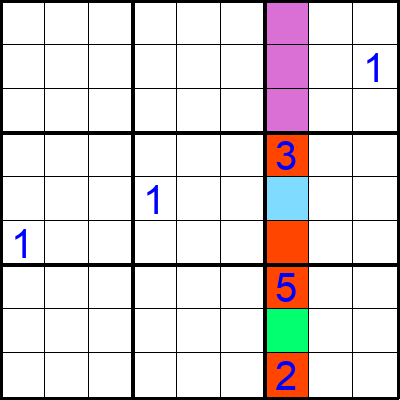}
  \caption{}
  \label{fig:tut_hs3}
\end{subfigure}
\begin{subfigure}{.245\linewidth}
  \centering
  \includegraphics[width=\linewidth]{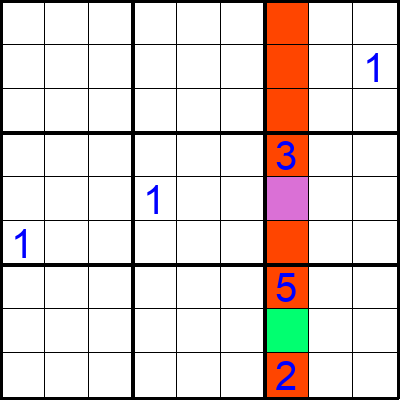}
  \caption{}
  \label{fig:tut_hs4}
\end{subfigure}
\begin{subfigure}{.245\linewidth}
  \centering
  \includegraphics[width=\linewidth]{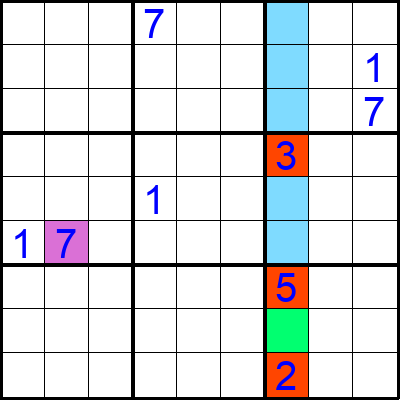}
  \caption{}
  \label{fig:tut_hs6}
\end{subfigure}
\begin{subfigure}{.245\linewidth}
  \centering
  \includegraphics[width=\linewidth]{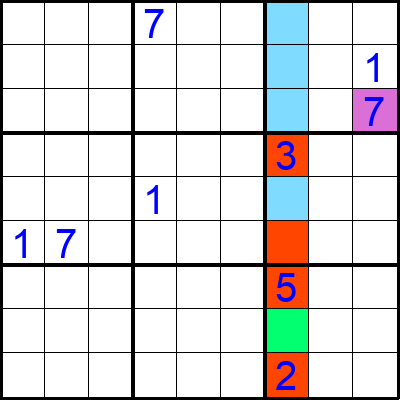}
  \caption{}
  \label{fig:tut_hs7}
\end{subfigure}
\begin{subfigure}{.245\linewidth}
  \centering
  \includegraphics[width=\linewidth]{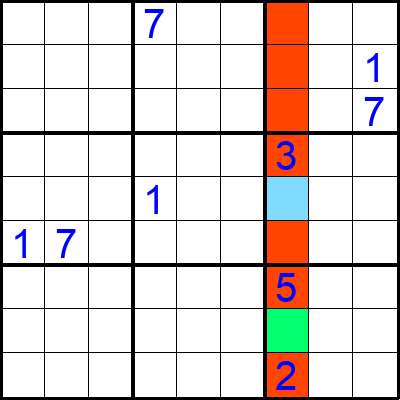}
  \caption{}
  \label{fig:tut_hs8}
\end{subfigure}
\begin{subfigure}{.245\linewidth}
  \centering
  \includegraphics[width=\linewidth]{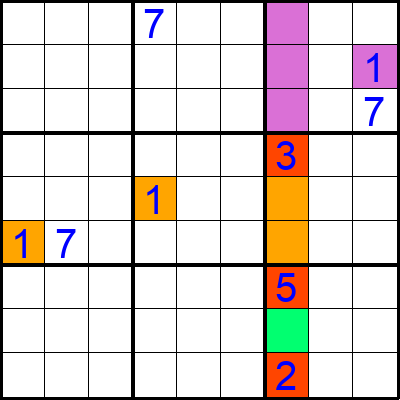}
  \caption{}
  \label{fig:tut_hs9a}
\end{subfigure}
\begin{subfigure}{.245\linewidth}
  \centering
  \includegraphics[width=\linewidth]{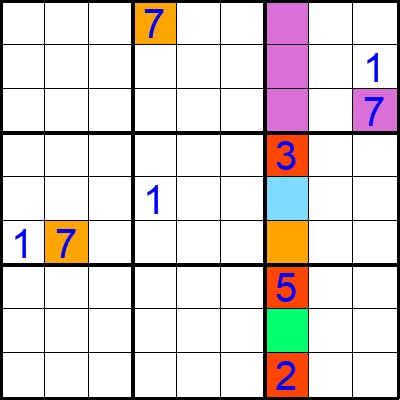}
  \caption{}
  \label{fig:tut_hs9b}
\end{subfigure}
\caption{(a) The tutorial puzzle randomly generated for this participant. (b) The contradiction exercise. The goal is to select the two cells containing 7s. (c) The Full House exercise. The goal is to fill in the green cell with 1. (d) The tutorial puzzle with distractors removed. (e-g) Positive example tasks. (h-j)) Negative example tasks. (k-l) Summary grids for both positive and negative examples.}
\label{fig:tutorial}
\end{figure}

As noted in the main text, each participant was randomly assigned a house type (row or column), a house index (between 1 and 9), a cell index (between 1 and 9), and 2 disjoint sets of 4 digits (all between 1 and 9) which would define the parameters of the tutorial puzzles, practice phase puzzles, and the control conditions in the test phase. 
Using these features, we randomly generated a hidden single puzzle to use during the tutorial (see the example for one participant in Figure \ref{fig:tut_hs1}) following the same generative procedure as puzzles in the practice phase, test phase, and questionnaire.
The tutorial puzzles always had one highlighted row or column based on the assigned house type and house index, and one target cell based on the cell index. Based on this puzzle, we provided two preliminary exercises to help solidify the general concepts of Sudoku before proceeding to the tutorial itself. In the first exercise (Figure \ref{fig:tut_contradiction}), participants were presented a grid with the same house as the tutorial completely filled in. 
However, at where the goal cell would be, rather than the correct digit, the grid would instead have a duplicate of another digit, forming a contradiction.
The goal of the exercise was to select the two cells where the same digit occurred twice to identify the contradiction. 
In the second exercise, the goal cell was cleared and the participant was asked to fill in the cell with the correct digit using the missing digit from the house, a technique known as Full House (Figure \ref{fig:tut_fullhouse}). 
Incorrect responses would invoke feedback noting that the participant's response already exists in the row or column.

\subsubsection{Hidden single tutorial}

The tutorial began with the presentation of the participant's tutorial puzzle with the distractors omitted (Figure \ref{fig:tut_hs1}). 
The tutorial then walked the participants through solving the puzzle by identifying the clues on the grid that constrain each of the blue cells.
As shown in Figure \ref{fig:tut_hs2}, the participant would be asked to find the instance of the target digit (in this case 1) that constrains the purple cell, for which the correct answer would be the 1 on the left-most column. 
They would continue this until all blue cells were eliminated (progressively highlighted red) and it was visually apparent that the only cell the target digit could be in is the goal cell, highlighted in green. 
After solving the simplified puzzle, the distractors were added and the participant was walked through a similar set of tasks. 
However, this time, they were asked to focus on the distractor, thus also providing a negative example. 
Moreover, they were asked to identify the cells that each clue constrains. 
For example, in Figure \ref{fig:tut_hs7}, they would need to select all 3 cells in the 3x3 box. 
This provided an alternative method of approaching the puzzle, starting with the clues to eliminate blue cells instead of checking if a blue cell could be eliminated as a possible location for a digit. 
After inevitably failing to solve for the distractor (one blue cell could not be eliminated), participants were shown color-coded grids for both the target (Figure \ref{fig:tut_hs9a}) and distractor (Figure \ref{fig:tut_hs9b}) side-by-side, indicating the types of constraints on the grid and showing how all blue cells could be eliminated for the target but not for the distractor.

\subsection{Puzzle generation}
\label{sec:supp:puzzle_generation}
All puzzles used in the experiment were generated with the following procedure according to the specified house type, goal cell, and digit set.

First, from the digit set of four numbers, we selected one as the target and one as the distractor.
From the remaining seven, we selected three as the in-house digits.
We chose one of the 2 boxes that intersect with the target house but not containing the goal cell to contain one target digit and one distractor digit.
The three cells in this box intersecting the target house were left empty.
Of the remaining 5 non-goal cells in the target house, we selected three to contain the three in-house digits.
Of the remaining 2 cells, we selected one to contain both the target and distractor cell in its row or column (whichever is orthogonal to the target house), with the other containing only the target digit.
The third distractor was placed anywhere in the grid such that it would intersect with the target house only in cells that already could not contain the distractor.
All selections were made randomly subject to the Sudoku constraints.

One instance is placed to constrain three empty cells in the target house by sharing the same box as them, and another to constrain a single empty cell by sharing a column or row (whichever is orthogonal to the target house) with it.
The third target digit instance forms a second orthogonal constraint, which is what forces the goal cell to be the only remaining cell in the target house that could contain the target, whereas the distractor digit is placed to allow two cells in the target house that could contain it.
3 cells in the highlighted house were filled with random remaining digits.
Because this would require at least 3 hints that share digits with the target, we added a distractor digit with 3 hints that constrain the same box and one of the 2 other unconstrained cells, making both the target and distractor digits salient as potential candidate target digits.

\subsection{Practice phase}
\label{sec:supp:practice_phase}

\begin{figure}[!ht]
\begin{subfigure}{.245\linewidth}
  \centering
  \includegraphics[width=\linewidth]{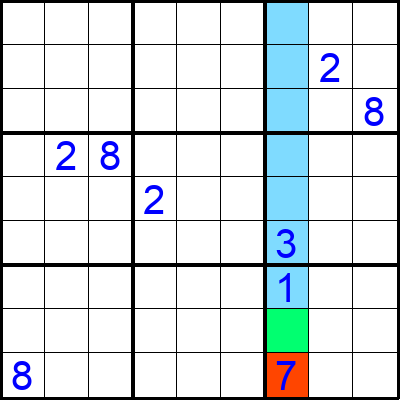}
  \caption{}
  \label{fig:p1_inhouse}
\end{subfigure}
\begin{subfigure}{.245\linewidth}
  \centering
  \includegraphics[width=\linewidth]{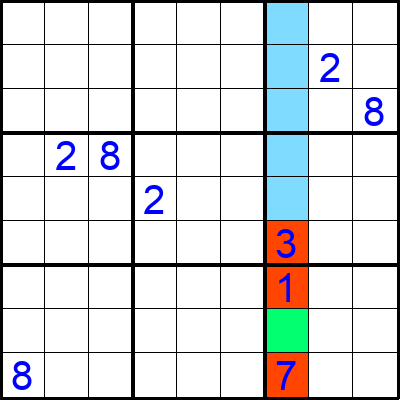}
  \caption{}
  \label{fig:p1_absent}
\end{subfigure}
\begin{subfigure}{.245\linewidth}
  \centering
  \includegraphics[width=\linewidth]{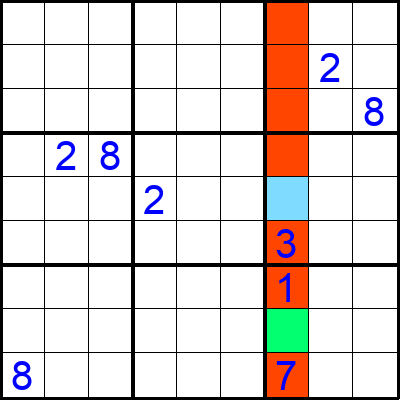}
  \caption{}
  \label{fig:p1_distractor}
\end{subfigure}
\begin{subfigure}{.245\linewidth}
  \centering
  \includegraphics[width=\linewidth]{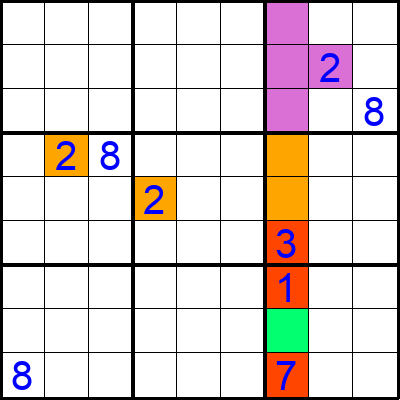}
  \caption{}
  \label{fig:p1_target}
\end{subfigure}
\caption{Sample feedback sequences during practice phase. (a) Feedback for submitting 7, an in-house response. (b) Feedback for submitting 4, an absent response. (c) Feedback for submitting 8, a distractor response. (d) Feedback for a correcting a previously incorrect response.}
\label{fig:p1_feedback}
\end{figure}

During the practice phase, participants were given detailed feedback whenever they gave incorrect responses (Figure \ref{fig:p1_feedback}). The feedback was customized to each puzzle and the type of response. Thus, if the participant gave an in-house response, only the cell containing their input digit would be highlighted red. If the response was an absent digit, all the cells containing digits would be highlighted red. If the response was a distractor, all blue cells would be made red except the cell that is unconstrained by any of the distractor clues. Once the participant corrected an incorrect response, all cells in the house and cells containing target clues were highlighted with different colors to indicate the different constraints. No feedback was given for correct first responses except that they were correct.

The following text would accompany the feedback.

\begin{itemize}
    \item In-house feedback (Figure \ref{fig:p1_inhouse}): 7 cannot be at the\footnote{This typo was made in the actual experiment} because 7 already exists in the same column in the red cell.
    \item Absent feedback (Figure \ref{fig:p1_absent}): It is not certain that 4 must be at the green cell because 4 may potentially be in a blue cell. The red cells cannot be 4 because they already contain digits.
    \item Distractor feedback (Figure \ref{fig:p1_distractor}): It is not certain that 8 must be at the green cell because 8 may potentially be in the blue cell. Note that neither the green cell nor the blue cell share the same row, column, or box with a 8.
    \item Target feedback (Figure \ref{fig:p1_target}): 2 is correct! We can be certain that the green cell must contain 2 because no other cell in its column can be a 2. The red cells cannot be 2 because they already contain digits. The empty purple cells cannot be 2 because they share they share the same box with a 2. The empty orange cells cannot be 2 because they share rows with other 2s.
\end{itemize}

\subsection{Test phase}
In the test phase, participants were presented with 64 puzzles to solve and were provided a single attempt and up to 2 minutes per puzzle.
Unlike the practice phase, participants were only told whether their responses were correct or incorrect.
The puzzles were grouped into 8 sets of 8, each containing the 8 possible combinations of the 3 positional feature dimensions: house type (HT), house index (HI) and cell index (CI).
These 8 combinations were arranged in a balanced Latin square such that each combination occurred once in each set, once in each position within a set, and occurs 4 times before every other combination and 4 times after.
The digit set was varied 4 times in each set of 8 so that all 16 combinations appeared once in every 16 trials.

\subsection{Strategy survey}
\label{sec:questionnaire}

After completing the text phase described in the main text, participants were first instructed on the contents of the following segment and asked 3 attention-check questions. Next, they were given a puzzle sharing the same house type and digit set as the tutorial, but with the goal cell located in the center box and were asked to select and enter the digit that must go in the goal cell. Without providing feedback on correctness, we then displayed the puzzle and their response for the remainder of the questionnaire, allowing the participants to refer to it as necessary. We asked the following questions in order of increasing specificity using free-response questions to elucidate general responses without additional prompting and more specific multiple-choice questions to formalize their strategies.

\begin{enumerate}
    \item "How confident do you feel that your answer is correct, expressed as a percentage?" (Responses were allowed between 0 and 100 in increments of 5.)
    \item "Explain as clearly as possible the steps you went through to choose your answer. Please be as detailed as possible so that someone else could replicate your strategy by following your response."
    \item "There are two numbers in the puzzle that occur three times outside of the row/column containing the target cell. Which of the following best describes how you chose between the two candidate numbers to consider?"
    \begin{enumerate}
        \item I noticed something in the puzzle that initially made one candidate seem more likely to be correct than the other.
        \item I arbitrarily chose between the two candidates because they seemed equally promising to consider.
    \end{enumerate}
    \item "What did you notice in the puzzle that initially made one candidate seem more likely to be correct than the other?"
    \item "Please select the cell(s) that initially made one candidate seem more likely be correct than the other." (Participants could select one or more cells in the grid.)
    \item "Please explain how the cell(s) you selected initially made one seem more likely to be correct than the other." (Previously selected cells were shown.)
    \item "After you selected a candidate to consider, did you check further to determine whether that candidate was actually correct or not?"
    \begin{enumerate}
        \item Yes, I checked to see whether the candidate was actually correct.
        \item No, I just submitted my original guess without checking any further.
    \end{enumerate}
    \item "What did you do to determine if that candidate was actually correct?"
    \item "Which of the following best describes the way you determined whether or not the candidate was actually the correct answer?"
    \begin{enumerate}
        \item I checked whether the candidate I chose could go in any of the empty blue cells in the row/column.
        \item I looked at other numbers in the puzzle until I noticed something that helped me decide whether or not the candidate was correct.
    \end{enumerate}
    \item "Please provide any additional information or clarifications to any of your previous responses so that we can most accurately understand as best we can how you solved this puzzle."
\end{enumerate}

Participants that did not respond to the puzzle with the target or dstractor skipped ahead to Question 10 after Question 2. Participants that responded with option (b) in Question 3 skipped ahead to Question 10. Participants that responded with option (b) in Question 7 skipped ahead to Question 10.

\subsection{Demographics survey}

Following the strategy questionnaire, we also asked the participants about their demographic information including their age, gender, highest level of education, and mathematical topics they have taken courses in. All participants, regardless of diagnostic test and survey results, were asked the following questions about their educational backgrounds. 
Participants that were filtered out from the diagnostics were presented the demographics survey immediately after the diagnostic survey screen.

"What is your highest level of education (including currently pursuing)?" Participants were allowed to select one of the following:

\begin{itemize}
    \item Have not graduated high school
    \item High school graduate, diploma or equivalent
    \item Associate degree
    \item Bachelor's degree
    \item Master's degree
    \item Professional degree (e.g. M.D., J.D.)
    \item Doctoral degree
\end{itemize}

"Degree status" Participants were allowed to select one of the following: 
\begin{itemize}
    \item Currently pursuing
    \item Completed
\end{itemize}

"Which of the following mathematics topics have you taken a course in? Select all that apply." Participants were allowed to select zero or more of the following:

\begin{itemize}
    \item High school algebra
    \item High school geometry
    \item Trigonometric functions
    \item Single-variable calculus
    \item Multi-variable calculus
    \item Linear algebra
    \item Probability \& statistics
    \item Discrete mathematics
    \item Formal logic
\end{itemize}

\clearpage
\newpage
\section{Reported regressions}
\label{sec:supp:lm_reported}

Here, we provide the exact formulas used in the regressions reported in the main text and the full list of fitted coefficient values. 
All regressions were fitted using the BRMS package in R \cite{brms} with default priors and MCMC settings.
All reported coefficients on accuracy models are in logits. 
All reported coefficients on response time models are in log$_2$(seconds).
We report 95\% highest density credible intervals for the parameter estimates.

Only trials with correct responses were used to fit the response time models.  
In each model, we accounted for improvements through practice using a log$_2$(trial) term and for individual variations through random effect intercepts for each participant. 
$t$ refers to the trial number (between 1 and 25 for the practice phase, between 1 and 64 for the test phase) and $s$ refers to the subject. 
The $(1 + \log_2 t | s)$ included in each regression indicates participant-level random effects. 
All other terms are fixed effects.

\subsection{Overall accuracy}
\label{sec:supp:reg_overall_accuracy}

We used Bayesian logistic mixed-effects models for predicting the correctness of each trial for both phases.
We note that the preregistered model was not Bayesian, but we chose to change the method for consistency with the remainder of the paper.
These models were only used for classifying solvers and non-solvers, and all classifications were consistent between both modeling approaches.

$$P(\text{correct}_{t,s}) \sim \log_2 t + (1 + \log_2 t | s)$$

\begin{table*}[!ht]
\begin{center}
\caption{Accuracy regression parameter estimates. Reported numbers in logits.}
\begin{tabular}{ll|lll}
\toprule
Phase & Term & Estimate & CI-L & CI-U \\
\midrule
Practice    & intercept       & -0.493 & -0.702 & -0.280 \\
\rowcolor[HTML]{EFEFEF} 
Practice    & $\log_2$(trial) &  0.344 &  0.287 & 0.404 \\
Test        & intercept       &  0.453 &  0.266 & 0.645 \\
\rowcolor[HTML]{EFEFEF} 
Test        & $\log_2$(trial) &  0.147 &  0.108 & 0.186 \\
\bottomrule
\end{tabular}
\label{table:overall_accuracy}
\end{center}
\end{table*}

\newpage
\subsection{Digit sets (DS)}

\textbf{Accuracy}
$$P(\text{correct}_{t,s}) \sim DS + \log_2 t + (1 + \log_2 t | s)$$
\begin{table*}[!ht]
\begin{center}
\caption{Digit set: accuracy regression parameter estimates. Reported numbers in logits.}
\begin{tabular}{l|lll|lll}
\toprule
& \multicolumn{3}{c}{Trials 1-16} & \multicolumn{3}{c}{Trials 17-64} \\
Term & Estimate & CI-L & CI-U & Estimate & CI-L & CI-U \\
\midrule
intercept & 2.405 & 1.745 & 3.182 & 2.903 & 2.334 & 3.537 \\
\rowcolor[HTML]{EFEFEF} 
$\log_2$(trial) & 0.086 & -0.142 & 0.308 & 0.091 & -0.035 & 0.217  \\
DS & -0.098 & -0.472 & 0.268 & 0.020 & -0.248 & 0.286  \\
\bottomrule
\end{tabular}
\label{table:ds_acc_16_48}
\end{center}
\end{table*}

\textbf{Duration}
$$\log(\text{duration}_{t,s}) \sim DS + \log_2 t + (1 + \log_2 t | s)$$
\begin{table*}[!ht]
\begin{center}
\caption{Digit set: duration regression parameter estimates. Reported numbers in log$_2$ seconds.}
\begin{tabular}{l|lll|lll}
\toprule
& \multicolumn{3}{c}{Trials 1-16} & \multicolumn{3}{c}{Trials 17-64} \\
Term & Estimate & CI-L & CI-U & Estimate & CI-L & CI-U \\
\midrule
intercept & 4.453 & 4.293 & 4.611 & 3.945 & 3.777 & 4.116 \\
\rowcolor[HTML]{EFEFEF} 
$\log_2$(trial)   & -0.126 & -0.168 & -0.084 & -0.076 & -0.100 & -0.053  \\
DS                & -0.022 & -0.086 & 0.043 & 0.000 & -0.032 & 0.032 \\
\bottomrule
\end{tabular}
\label{table:ds_dur_16_48}
\end{center}
\end{table*}

\newpage
\subsection{Goal position (GP)}

\textbf{Accuracy}
$$P(\text{correct}_{t,s}) \sim GP + \log_2 t + (1 + \log_2 t | s)$$
\begin{table*}[!ht]
\begin{center}
\caption{Goal position: accuracy regression parameter estimates. Reported numbers in logits.}
\begin{tabular}{l|lll|lll}
\toprule
& \multicolumn{3}{c}{Trials 1-16} & \multicolumn{3}{c}{Trials 17-64} \\
Term & Estimate & CI-L & CI-U & Estimate & CI-L & CI-U \\
\midrule
intercept & 2.223 & 1.525 & 3.042 & 3.151 & 2.539 & 3.829 \\
\rowcolor[HTML]{EFEFEF} 
$\log_2$(trial)   & 0.089 & -0.143 & 0.305 & 0.092 & -0.030 & 0.209  \\
GP                & 0.165 & -0.240 & 0.570 & -0.314 & -0.668 & 0.019  \\
\bottomrule
\end{tabular}
\label{table:gp_acc_16_48}
\end{center}
\end{table*}

\textbf{Duration}
$$\log(\text{duration}_{t,s}) \sim GP + \log_2 t + (1 + \log_2 t | s)$$
\begin{table*}[!ht]
\begin{center}
\caption{Goal position: duration regression parameter estimates. Reported numbers in log$_2$ seconds.}
\begin{tabular}{l|lll|lll}
\toprule
& \multicolumn{3}{c}{Trials 1-16} & \multicolumn{3}{c}{Trials 17-64} \\
Term & Estimate & CI-L & CI-U & Estimate & CI-L & CI-U \\
\midrule
intercept & 4.354 & 4.198 & 4.515 & 3.877 & 3.708 & 4.055 \\
\rowcolor[HTML]{EFEFEF} 
$\log_2$(trial)   & -0.126 & -0.168 & -0.086 & -0.076 & -0.099 & -0.053  \\
GP                & 0.115 & 0.040 & 0.189 & 0.089 & 0.050 & 0.128 \\
\bottomrule
\end{tabular}
\label{table:gp_dur_16_48}
\end{center}
\end{table*}

\newpage
\subsection{House type (HT)}

\textbf{Accuracy}
$$P(\text{correct}_{t,s}) \sim HT + \log_2 t + (1 + \log_2 t | s)$$
\begin{table*}[!ht]
\begin{center}
\caption{House type: accuracy regression parameter estimates. Reported numbers in logits.}
\begin{tabular}{l|lll|lll}
\toprule
& \multicolumn{3}{c}{Trials 1-16} & \multicolumn{3}{c}{Trials 17-64} \\
Term & Estimate & CI-L & CI-U & Estimate & CI-L & CI-U \\
\midrule
intercept         & 2.508  &  1.842 & 3.308 & 2.867 & 2.286  & 3.497 \\
\rowcolor[HTML]{EFEFEF} 
$\log_2$(trial)   & 0.089  & -0.138 & 0.312 & 0.089 & -0.033 & 0.215  \\
HT                & -0.302 & -0.666 & 0.059 & 0.095 & -0.170 & 0.364  \\
\bottomrule
\end{tabular}
\label{table:ht_acc_16_48}
\end{center}
\end{table*}

\textbf{Duration}
$$\log(\text{duration}_{t,s}) \sim HT + \log_2 t + (1 + \log_2 t | s)$$
\begin{table*}[!ht]
\begin{center}
\caption{House type: duration regression parameter estimates. Reported numbers in log$_2$ seconds.}
\begin{tabular}{l|lll|lll}
\toprule
& \multicolumn{3}{c}{Trials 1-16} & \multicolumn{3}{c}{Trials 17-64} \\
Term & Estimate & CI-L & CI-U & Estimate & CI-L & CI-U \\
\midrule
intercept         & 4.292 & 4.136 & 4.447 & 3.911 & 3.746 & 4.082 \\
\rowcolor[HTML]{EFEFEF} 
$\log_2$(trial)   & -0.128 & -0.168 & -0.087 & -0.076 & -0.098 & -0.053  \\
GP                & 0.319 & 0.258 & 0.379 & 0.055 & 0.022 & 0.088 \\
\bottomrule
\end{tabular}
\label{table:ht_dur_16_48}
\end{center}
\end{table*}

\newpage
\subsubsection{Accounting for tutorial house type}
\label{sec:supp:reg_ht0}

We checked to see if the effects were moderated by whether the participants were taught the hidden single technique using rows or column. 
Specifically, we added a new term $C$ indicating whether the participant's tutorial used columns.

\textbf{Accuracy}
$$P(\text{correct}_{t,s}) \sim HT + C + HT * C + \log_2 t + (1 + \log_2 t | s)$$
\begin{table*}[!ht]
\begin{center}
\caption{House type accounting for tutorial house type: accuracy regression parameter estimates. Reported numbers in logits.}
\begin{tabular}{l|lll|lll}
\toprule
& \multicolumn{3}{c}{Trials 1-16} & \multicolumn{3}{c}{Trials 17-64} \\
Term & Estimate & CI-L & CI-U & Estimate & CI-L & CI-U \\
\midrule
intercept         & 2.688  &  1.937 & 3.490 & 2.969 & 2.334  & 3.652 \\
\rowcolor[HTML]{EFEFEF} 
$\log_2$(trial)   & 0.104  & -0.111 & 0.321 & 0.092 & -0.027 & 0.215  \\
HT                & -0.405 & -0.918 & 0.114 & 0.018 & -0.367 & 0.407  \\
\rowcolor[HTML]{EFEFEF} 
C           & -0.419 & -1.169 & 0.353 & -0.207 & -0.857 & 0.453  \\
HT*C        & -0.183 & -0.528 & 0.876 & 0.151 & -0.391 & 0.680  \\
\bottomrule
\end{tabular}
\label{table:ht0_acc_16_48}
\end{center}
\end{table*}

\textbf{Duration}
$$\log(\text{duration}_{t,s}) \sim HT + C + HT * C + \log_2 t + (1 + \log_2 t | s)$$
\begin{table*}[!ht]
\begin{center}
\caption{House type accounting for tutorial house type: duration regression parameter estimates. Reported numbers in log$_2$ seconds.}
\begin{tabular}{l|lll|lll}
\toprule
& \multicolumn{3}{c}{Trials 1-16} & \multicolumn{3}{c}{Trials 17-64} \\
Term & Estimate & CI-L & CI-U & Estimate & CI-L & CI-U \\
\midrule
intercept         & 4.318  &  4.119 & 4.523 & 4.014 & 3.798  & 4.225 \\
\rowcolor[HTML]{EFEFEF} 
$\log_2$(trial)   & -0.128  & -0.169 & 0.-0.089 & -0.076 & -0.100 & -0.053  \\
HT                & 0.331 & 0.246 & 0.421 & 0.041 & -0.003 & 0.083  \\
\rowcolor[HTML]{EFEFEF} 
C           & -0.056 & -0.316 & 0.206 & -0.205 & -0.463 & 0.062  \\
HT*C        & -0.023 & -0.150 & 0.101 & 0.028 & -0.036 & 0.090  \\
\bottomrule
\end{tabular}
\label{table:ht0_dur_16_48}
\end{center}
\end{table*}

\clearpage
\subsection{Education}
\label{sec:supp:reg_education}

In the questionnaire, we asked for highest education pursued, whether completed or in-progress. However, because we found that some education levels were extremely rare in our dataset (e.g. PhD), we converted them into years of education with the following mapping: \textit{Incomplete High School} $\rightarrow$ 10, \textit{High School} $\rightarrow$ 12, \textit{Associate's Degree}  $\rightarrow$ 14, \textit{Bachelor's Degree}  $\rightarrow$ 16, \textit{Master's Degree}  $\rightarrow$ 18, \textit{Professional Degree}  $\rightarrow$ 20, \textit{PhD}  $\rightarrow$ 21.

\subsubsection{Education model}

$$\text{num\_solved} \sim \text{education}$$

\begin{table*}[!ht]
\begin{center}
\caption{Years of education regression parameter and $R^2$ estimates}
\begin{tabular}{l|lll}
\toprule
Term & Estimate & CI-L & CI-U \\
\midrule
intercept & 36.129 & 20.514 & 51.166 \\
\rowcolor[HTML]{EFEFEF} 
education & 1.455 & 0.448 & 2.453 \\
$R^2$     & 0.032 & 0.003 & 0.077 \\
\bottomrule
\end{tabular}
\label{table:edu}
\end{center}
\end{table*}

\subsubsection{All math model}
$$\text{num\_solved} \sim \text{alg} + \text{geom} + \text{trig} + \text{sv\_calc} + \text{mv\_calc} + \text{linalg} + \text{pr\_stat} + \text{disc} + \text{logic}$$

\begin{table*}[!ht]
\begin{center}
\caption{Math courses regression parameter and $R^2$ estimates}
\begin{tabular}{l|lll}
\toprule
Term & Estimate & CI-L & CI-U \\
\midrule
intercept & 38.757 & 32.505 & 44.885 \\
\rowcolor[HTML]{EFEFEF} 
alg       &  9.684 &  2.874 & 16.781 \\
geom      &  9.813 &  3.785 & 16.034 \\
\rowcolor[HTML]{EFEFEF} 
trig      &  3.139 & -2.368 &  8.724 \\
sv\_calc   &  4.617 & -2.156 & 11.540 \\
\rowcolor[HTML]{EFEFEF} 
mv\_calc   & -1.006 & -8.985 &  6.611 \\
linalg    &  0.392 & -5.576 &  6.540 \\
\rowcolor[HTML]{EFEFEF} 
pr\_stat   &  2.619 & -2.267 &  7.237 \\
disc      &  6.361 & -3.138 & 15.396 \\
\rowcolor[HTML]{EFEFEF} 
logic     & -1.280 & -8.287 &  5.729 \\
$R^2$     & 0.207 & 0.132 & 0.279 \\
\bottomrule
\end{tabular}
\label{table:math}
\end{center}
\end{table*}

\clearpage
\subsubsection{Algebra and geometry model}

$$\text{num\_solved} \sim \text{alg} + \text{geom}$$

\begin{table*}[!ht]
\begin{center}
\caption{Algebra and geometry courses regression parameter and $R^2$ estimates}
\begin{tabular}{l|lll}
\toprule
Term & Estimate & CI-L & CI-U \\
\midrule
intercept   & 40.722 & 34.966 & 46.395 \\
\rowcolor[HTML]{EFEFEF} 
alg         &  9.131 & 2.146 & 16.065 \\
geom        & 12.704 & 7.070 & 18.691 \\
\rowcolor[HTML]{EFEFEF} 
$R^2$       & 0.152 & 0.086 & 0.226 \\
\bottomrule
\end{tabular}
\label{table:alg_geom}
\end{center}
\end{table*}

\subsubsection{Education, algebra and geometry model}

$$\text{num\_solved} \sim \text{education} + \text{alg} + \text{geom}$$

\begin{table*}[!ht]
\begin{center}
\caption{Years of education, algebra course, and geometry course regression parameter and $R^2$ estimates}
\begin{tabular}{l|lll}
\toprule
Term & Estimate & CI-L & CI-U \\
\midrule
intercept  & 25.476 & 10.401 & 39.890 \\
\rowcolor[HTML]{EFEFEF} 
education  &  1.046 &  0.099 &  1.995 \\
alg        &  9.591 &  2.573 & 16.401 \\
\rowcolor[HTML]{EFEFEF} 
geom       & 11.479 &  5.605 & 17.328 \\
$R^2$       & 0.168 &  0.098 &  0.240 \\
\bottomrule
\end{tabular}
\label{table:edu_alg_geom}
\end{center}
\end{table*}

\clearpage
\subsubsection{All math and education model}

$$\text{num\_solved} \sim \text{education} + \text{alg} + \text{geom} + \text{trig} + \text{sv\_calc} + \text{mv\_calc} + \text{linalg} + \text{pr\_stat} + \text{disc} + \text{logic}$$

\begin{table*}[!ht]
\begin{center}
\caption{Years of education and math courses regression parameter and $R^2$ estimates}
\begin{tabular}{l|lll}
\toprule
Term & Estimate & CI-L & CI-U \\
\midrule
intercept & 30.14 & 15.36 & 45.15 \\
\rowcolor[HTML]{EFEFEF} 
education &  0.61 & -0.39 &  1.56 \\
alg       &  9.93 &  2.98 & 16.81 \\
\rowcolor[HTML]{EFEFEF} 
geom      &  9.40 &  3.39 & 15.53 \\
trig      &  2.70 & -3.22 &  8.55 \\
\rowcolor[HTML]{EFEFEF} 
sv\_calc   &  4.64 & -2.18 & 11.55 \\
mv\_calc   & -1.28 & -8.87 &  6.83 \\
\rowcolor[HTML]{EFEFEF} 
linalg    &  0.40 & -6.02 &  6.69 \\
pr\_stat   &  2.10 & -2.73 &  6.96 \\
\rowcolor[HTML]{EFEFEF} 
disc      &  6.20 & -3.09 & 15.17 \\
logic     & -1.62 & -8.52 &  5.51 \\
\rowcolor[HTML]{EFEFEF} 
$R^2$     & 0.213 & 0.139 & 0.285 \\
\bottomrule
\end{tabular}
\label{table:all}
\end{center}
\end{table*}

\clearpage
\newpage
\section{Unreported regressions}
\label{sec:supp:lm_unreported}

Here, we describe the regressions that were not included in the main article but were committed in the preregistration. Due to the large number of parameters, we do not include their estimates here. For the coefficients of these models, see the spreadsheet in the project repository.

As in the previous section, $t$ refers to the trial number (between 1 and 64) and $s$ refers to the subject. The $(1 + \log_2 t | s)$ included in each regression indicates participant-level random effects. All other terms are fixed effects.

\subsection{Digit sets (DS)}

One additional regression was conducted for the digit sets analysis which included an interaction term between the treatment and practice effects. This model was fitted using data from all 64 test phase trials.

\textbf{Accuracy}
$$P(\text{correct}_{t,s}) \sim DS + log_2 t + DS * log_2 t + (1 + log_2 t | s)$$

\textbf{Duration}
$$\log(\text{duration}_{t,s}) \sim DS + log_2 t + DS * log_2 t + (1 + log_2 t | s)$$

\subsection{Goal position (GP)}

One additional regression was conducted for the goal position analysis which included an interaction term between the treatment and practice effects. While this model was not committed in the preregistration, we include it for completeness. This model was fitted using data from all 64 test phase trials.

\textbf{Accuracy model}
$$P(\text{correct}_{t,s}) \sim GP + log_2 t + GP * log_2 t + (1 + log_2 t | s)$$

\textbf{Duration model}
$$\log(\text{duration}_{t,s}) \sim GP + log_2 t + GP * log_2 t + (1 + log_2 t | s)$$

\subsection{House index (HI) and cell index (CI)}
\label{sec:supp:unreg_hi_ci}

These regressions were the original formulations for analyzing the effect of moving the goal cell. Due to the transposition of the grid when applying the house type (HT) condition and the high perceptual dissimilarity, we intended to perform separate regressions for puzzles with the HT condition applied and for puzzles without the HT condition applied. However, post preregistration, we had decided that the effect of interest was less about exactly which axis the goal cell had translated across and more about the presence of a change at all. Moreover, we had originally planned to account for an interaction between HI and CI, but decided that the interaction complicated the interpretation. Thus, we collapsed house index and cell index conditions as the goal position (GP) condition in favor of improved power and interpretability, considering the condition to be applied if one or both of the house index or cell index had been applied.

Note that each of the four regressions below have been fitted separately using trials with and without house type conditions.

\textbf{Accuracy: 16-trial and 48-trial models}

$$P(\text{correct}_{t,s}) \sim HI + CI + HI * CI + (1 + log_2 t | s)$$

\textbf{Duration: 16-trial and 48-trial models}

$$P(\text{duration}_{t,s}) \sim HI + CI + HI * CI + (1 + log_2 t | s)$$

\textbf{Accuracy: 64-trial model}

$$P(\text{correct}_{t,s}) \sim HI + CI + HI * CI + HI * log_2 t + CI * log_2 t + HI * CI * log_2 t + (1 + log_2 t | s)$$

\textbf{Duration: 64-trial model}

$$P(\text{duration}_{t,s}) \sim HI + CI + HI * CI + HI * log_2 t + CI * log_2 t + HI * CI * log_2 t + (1 + log_2 t | s)$$

\subsection{House type (HT)}

One additional regression was conducted for the house type analysis which included an interaction term between the treatment and practice effects. This model was fitted using data from all 64 test phase trials.

\textbf{Accuracy model}

$$P(\text{correct}_{t,s}) \sim HT + log_2 t + HT * log_2 t + (1 + log_2 t | s)$$

\textbf{Duration model}

$$\log(\text{duration}_{t,s}) \sim HT + log_2 t + HT * log_2 t + (1 + log_2 t | s)$$

\clearpage
\clearpage
\section{Practice phase models}
\label{sec:supp:hmm}

\subsection{Aggregate model parameters}
\label{sec:supp:agg_model_params}

Parameters were fitted using 5,000 gradient updates.
A priori (unfitted) parameters are shown as integers.
$\mathbf{WR}$ indicates response likelihoods for each strategy.
Response types are listed in the order of in-house, absent, distractor, and target.
Strategies are listed in the order of uniform guess, avoid direct constraints, prevalent digits, and successful.

\subsubsection{Solvers}

\begin{align*}
    \mathbf{a} = 
        \begin{bmatrix}
          0.085  & 0.241  & 0.310  & 0.365 \\
        \end{bmatrix}
  &&
  \mathbf{X} = 
    \begin{bmatrix}
      0.000  & 0.999  & 0.000  & 0.000 \\
      0      & 0.106  & 0.892  & 0.001 \\
      0      & 0      & 0.783  & 0.217 \\
      0      & 0      & 0      & 1 
    \end{bmatrix}
  \\\\
  \mathbf{W} = 
    \begin{bmatrix}
      1.000 & 0 & 0 & 0 \\
      0.000 & 1.000 & 0 & 0 \\
      0.000 & 0.054 & 0.946 & 0 \\
      0.000 & 0.000 & 0.090 & 0.910
    \end{bmatrix}
  &&
  \mathbf{WR} = 
    \begin{bmatrix}
      0.333 & 0.444 & 0.111 & 0.111 \\
      0.000 & 0.667 & 0.167 & 0.167 \\
      0.000 & 0.036 & 0.482 & 0.482 \\
      0.000 & 0.000 & 0.045 & 0.955
    \end{bmatrix}
\end{align*}

\subsubsection{Non-solvers}

\begin{align*}
    \mathbf{a} = 
        \begin{bmatrix}
          0.109  & 0.397  & 0.463  & 0.032 \\
        \end{bmatrix}
    &&
  \mathbf{X} = 
    \begin{bmatrix}
      0.560  & 0.001  & 0.437  & 0.002 \\
      0      & 0.770  & 0.229  & 0.001 \\
      0      & 0      & 0.993  & 0.007 \\
      0      & 0      & 0      & 1 
    \end{bmatrix}
  \\\\
  \mathbf{W} = 
    \begin{bmatrix}
      1.000 & 0 & 0 & 0 \\
      0.109 & 0.891 & 0 & 0 \\
      0.024 & 0.038 & 0.939 & 0 \\
      0.016 & 0.016 & 0.185 & 0.783
    \end{bmatrix}
  &&
  \mathbf{WR} = 
    \begin{bmatrix}
      0.333 & 0.444 & 0.111 & 0.111 \\
      0.036 & 0.642 & 0.161 & 0.161 \\
      0.008 & 0.036 & 0.478 & 0.478 \\
      0.005 & 0.018 & 0.097 & 0.880 \\
    \end{bmatrix}
\end{align*}

\clearpage
\section{Questionnaire}
\label{sec:supp:questionnaire}

\subsection{Multiple choice questions}
\label{sec:supp:questionnaire_mcq}

\begin{figure}
    \centering
    \includegraphics[width=.8\linewidth]{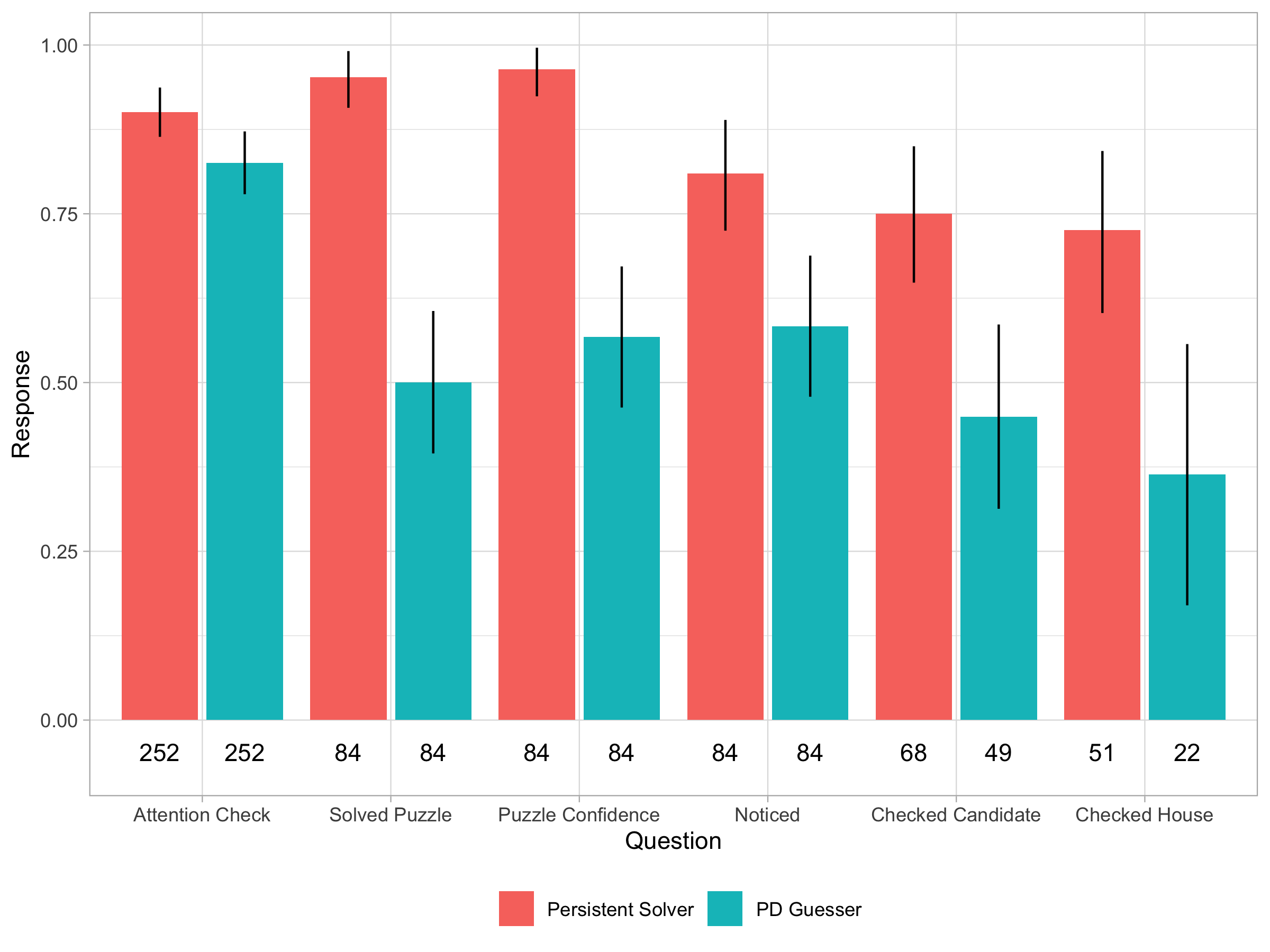}
    \caption{Quantitative responses by group. Total number of responses at the bottom of each bar for each question and group. Each participant was asked 3 attention check questions.}
    \label{fig:q_mc}
\end{figure}

Comparing the responses between the persistent-solvers and PD-guessers, we found significant differences across all 3 questions.
Specifically, when asked about how they chose which of the two prevalent digits they considered, 80.95\% of persistent-solvers and 58.33\% of PD-guessers responded that they had noticed something in the puzzle that made one candidate seem more likely versus chose arbitrarily.
Among those that noticed something, 75.00\% of persistent-solvers and 45.90\% of PD-guessers responded that they had further checked to see if their chosen candidate was correct versus submitted without checking.
Finally, among those that checked, 72.55\% of persistent-solvers and 36.36\% of PD-guessers responded that they checked to see if the candidate could go in another cell in the house versus looking for information in other numbers.
Figure~\ref{fig:q_mc} and Table~\ref{table:q_mc} show group means and 95\% CI for each question.

\begin{table*}[!ht]
\begin{center}
\caption{Questionnaire multiple choice responses rates and differences between persistent solvers and PD guessers}
\begin{tabular}{l|lllll}
\toprule
Measure & Mean PS & Mean PDG & Mean Diff & Diff CI-L & Diff CI-U \\
\midrule
Attention Check   & 90.08\% & 82.54\% & 7.54\% & 1.68\% & 13.55\% \\
\rowcolor[HTML]{EFEFEF} 
Solved Puzzle     & 95.24\% & 50.00\% & 45.24\% & 32.75\% & 55.98\% \\
Puzzle Confidence & 96.43\% & 55.95\% & 40.48\% & 28.60\% & 51.18\% \\
\rowcolor[HTML]{EFEFEF} 
Noticed           & 80.95\% & 58.33\% & 22.62\% & 8.83\% & 35.78\% \\
Checked Candidate & 75.00\% & 45.90\% & 30.10\% & 12.24\% & 45.54\% \\
\rowcolor[HTML]{EFEFEF} 
Checked House     & 72.55\% & 36.36\% & 36.19\% & 12.33\% & 56.94\% \\
\bottomrule
\end{tabular}
\label{table:q_mc}
\end{center}
\end{table*}

\clearpage
\subsection{Free response ratings}

\subsubsection{Participant selection}
\label{sec:supp:qrating_subselection}

For our ratings of questionnaire results, we focused our analysis on subsets of the solver and non-solver groups based on test phase performance.  
Among the solvers, we selected a subset we call \textit{persistent solvers} that demonstrated high accuracy at the end of the test phase according to a logistic regression model fitted to the test phase data, similar to the logistic regression fitted to the practice phase data in the original solver classification.
84 solvers that had a predicted accuracy of at least 80\% on the 64th trial of the test phase were classified as persistent solvers and were included in the strategy ratings.

Among the non-solvers, we focused on a subset we call \textit{PD-guessers} that exhibited evidence of consistently guessing between the target and distractor. 
PD-guessers were defined as non-solvers that had a predicted accuracy of at most 60\% at the end of the test phase according to the logistic regression model, had solved 3 to 5 of the last 8 puzzles in the test phase, and had selected the target or the distractor in at least 58 of the 64 test phase puzzles.
84 of 183 non-solvers were identified as PD-guessers and were included in the strategy ratings.

\subsubsection{Rating options}
\label{sec:supp:qrating_options}

The following options were available for identifying whether or not the responses indicated awareness of error. This was only done for puzzles that were not solved correctly.

\begin{enumerate}
    \item \textbf{Yes with explanation:} The participant indicates a clear realization that the answer they entered was not correct, and explains why they decided that the other answer was correct.
    \item \textbf{No or yes w/o explanation:} The participant's response was incorrect, but they did not explicitly report realizing that they had answered incorrectly, or the response suggests they think they were wrong without certainty or clarity about the reason.
\end{enumerate}

The following options were available for identifying prevalent digits mentioned in the responses.

\begin{enumerate}
    \item \textbf{Both:} The participant mentioned both prevalent digits by name, or otherwise led you to believe they realize that they have to choose among these two digits.
    \item \textbf{Target:} The participant mentioned the target, by name or in some other way, but did not mention the distractor, and did not indicate that there were two prevalent digits.
    \item \textbf{Distractor:} The participant mentioned the distractor, by name or in some other way, but did not mention the target.
    \item \textbf{Neither:} The participant did not mention either the target or the distractor, either by name or some other way, and there is no indication of awareness that there are two prevalent digits.
    \item \textbf{Vague or uncertain:} The participants' answer could be signaling that they were selecting between the two prevalent digits or that they were focusing on a digit that cannot go elsewhere or the distractor, but they do so vaguely or in a way that it is hard to be certain.
\end{enumerate}

\begin{figure}[ht!]
\centering
\includegraphics[width=.7\linewidth]{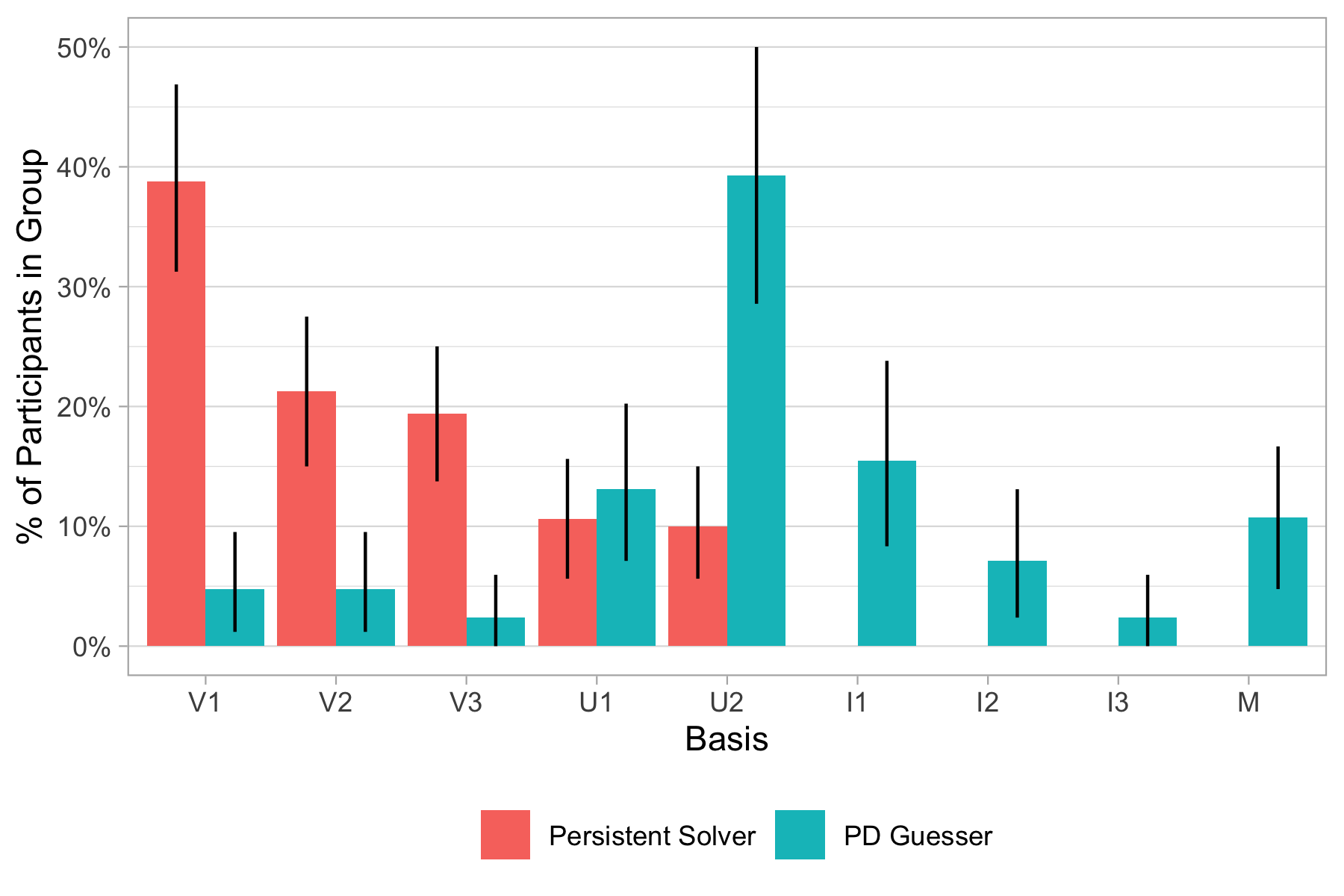}
\caption{Ratings of self-reported strategies between persistent solvers and PD guessers. `Other' category was never used by either rater.}
\label{fig:supp:q_ratings}
\end{figure}

The following options were available for identifying the basis for choosing between the two prevalent digits.
Figure~\ref{fig:supp:q_ratings} shows the distributions of ratings for solvers and non-solvers.

\begin{itemize}
    \item [\textbf{V1}] \textbf{Chose digit that cannot go elsewhere:} The participant states that a digit cannot go in any of the empty blue cells and gives that digit as the answer. As a shortcut variant of category A, participant states that they found a cell in the target house that was only constrained by one of the two prevalent digits and chose that digit as the answer.  To be valid, this must be a cell that is not in fact constrained by the distractor.
    \item [\textbf{V2}] \textbf{One PD can go elsewhere, chose the other:} The participant states that one of the prevalent digits can go in one of the empty blue cells and concludes that the other prevalent digit is the answer.
    \item [\textbf{V3}] \textbf{Found cell where one PD could go and the other could not:} The participant mentions finding a blue cell that can contain one of the two prevalent digits and not the other, and chooses the digit that cannot go in that cell as the answer.
    \item [\textbf{U1}] \textbf{Potentially valid but general or not fully specified:} The participant’s response provides incomplete information about how they came to choose their answer, but could be a general or under-specified description of a valid procedure.
    \item [\textbf{U2}] \textbf{Unclear, confused or missing basis:} The answer attempts to provide information about how the choice was made, but is unclear, incorrect, confusing, or fails to specify the procedure used to select between the PDs.
    \item [\textbf{I1}] \textbf{Explicit guess:} The participant indicates that they know they are guessing (either completely at random or between the two prevalent digits)
    \item [\textbf{I2}] \textbf{Irrelevant basis for choice:} The participant indicated a basis for choosing the answer that was not related to the logic of the puzzle.
    \item [\textbf{I3}]  \textbf{Chose the most frequent digit:} The participant indicates they chose the digit that occurred the most frequently in the puzzle, apparently not realizing that there were always two digits that both occurred exactly 3 times.
    \item [\textbf{M}] \textbf{Did not answer the question:} Response does not address the question.
    \item [\textbf{O}] \textbf{Other: } None of the above.
\end{itemize}

\begin{figure}[t]
  \hspace*{\fill}%
    {\includegraphics[width=2in]{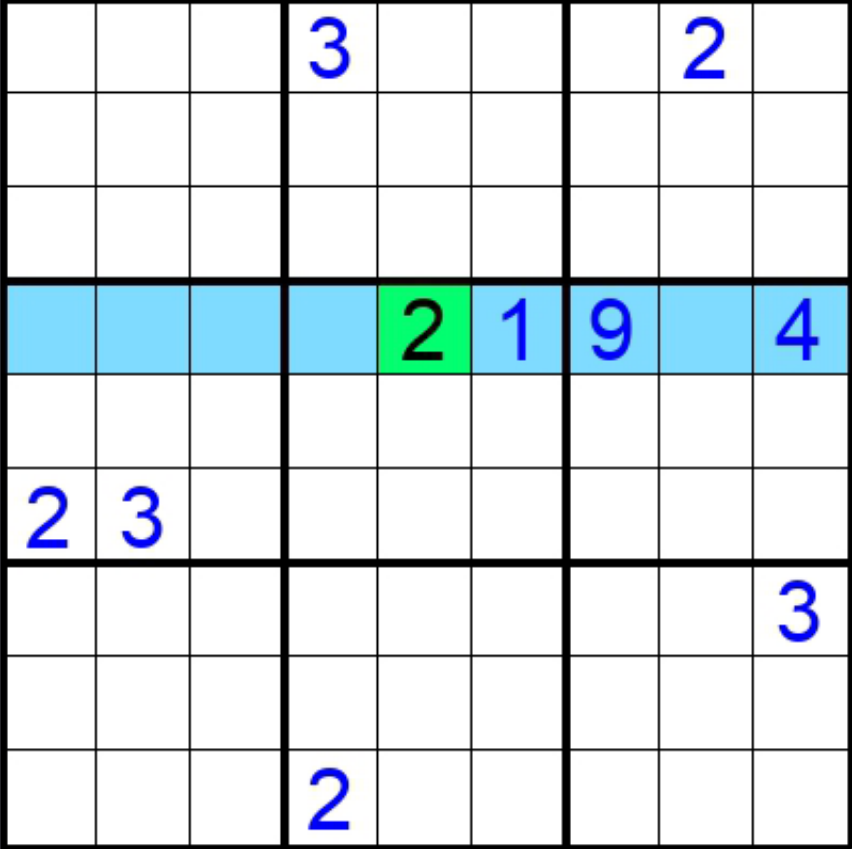}}
  \hfill%
    {\includegraphics[width=2in]{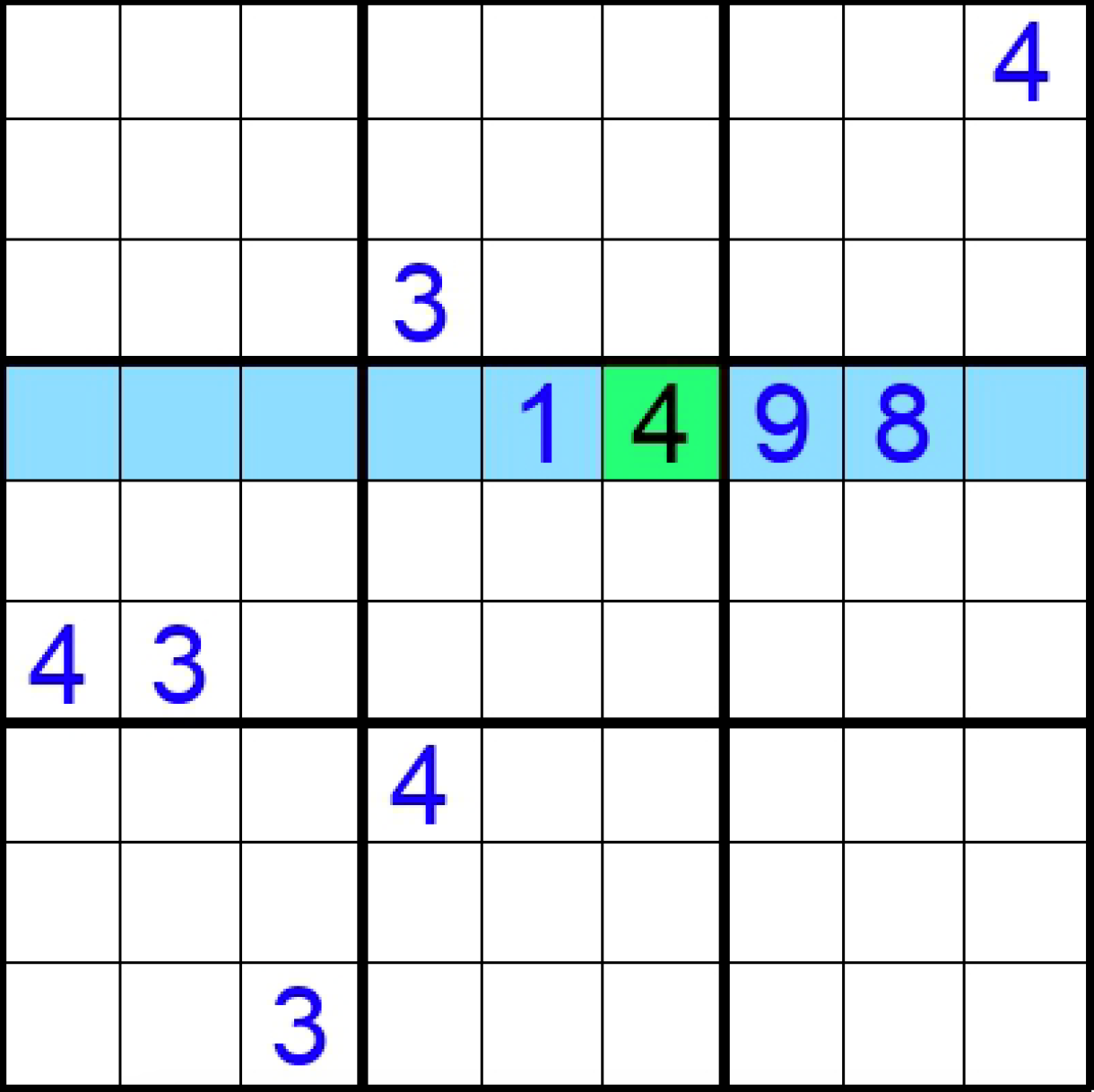}}%
  \hspace*{\fill}%
  \caption{Left: A hidden single puzzle in which the shortcut variant of option V1 would produce the correct answer. Right: A hidden single puzzle in which the shortcut variant of option V1 would \textit{not} produce the correct answer.}
  \label{fig:supp:qpuzzles}
\end{figure}

The shortcut variant of option V1 was one heuristic that we observed that did not clearly fit into one of the four strategies outlined in Section~\ref{sec:main:results_p1}.
As an illustration, take one solver's response after having correctly solved the left puzzle in Figure~\ref{fig:supp:qpuzzles}:
``The first thing I did was locate the square section that contains both of the possible numbers and the three empty blue squares. After locating this section, I can completely ignore this area since it is irrelevant to finding the answer now. Next I look for either of the two possible numbers (in this case 2 or 3) which is isolated, in a row or column all by itself. Upon finding that number, I'm finished. Just take that number, and since you know that its counterpart will be filled in in the blue row, you can just take the isolated number and make it your answer.''
    
This description is a valid method of arriving at the target digit for the given puzzle, but is not a valid strategy as it is not guaranteed to work in all the hidden single puzzles, such as the right puzzle in Figure~\ref{fig:supp:qpuzzles}.
In this case, the 3 on the bottom row and the 4 on the top row are both sole occupants of their respective columns so that both the distractor and the target digits meet the ``isolated number'' criterion.
This strategy could be made valid by checking that the cell in the blue highlighted house that the 3 intersects with is also constrained by a 3 in its 3x3 box and selecting the other prevalent digit as the target.
The method used to generate the hidden single puzzles has a 40\% chance to produce puzzles with these redundantly constraining distractors.
If one exclusively uses this strategy and randomly guesses between the 2 prevalent digits when this situation occurs, the expected accuracy would be 80\%, which coincidentally is also the value of the decision boundary we used to classify solvers and non-solvers at the end of the practice phase.
Of the 84 persistent-solvers' responses examined, only 6 were noted to be describing a strategy that resembled this heuristic.
    
\subsubsection{Rating design}
\label{sec:supp:questionnaire_rating_design}

One author (JLM) went through all 168 persistent solvers' and PD-guessers' responses to the first free-response question in an effort to develop a set of categories into which individual participants' responses could be sorted, with access to the specific puzzle each participant had just attempted to solve, the participant's chosen response, and summary scores characterizing the participant's performance both on the practice and test phases of the experiment.  In so doing, he noted that persistent solvers typically (a) referenced the target digit or both the target and the distractor; (b) provided one of three types of responses that seemed to describe a valid solution strategy given the rules of Sudoku and further constraints on the constructions of the puzzles used, and, (c) in the small number of cases in which a solver incorrectly chose the distractor, explicitly reported realizing that they had selected the wrong digit and explained why their choice was wrong.  The author also noted that non-solvers were (a) less likely to mention either or both prevalent digits even if they almost always chose one of the two prevalent digits, (b) rarely described a valid solution strategy, and (c) in the cases where they chose the distractor, never expressed a clear understanding that the response was incorrect.  The author began to develop a set of scoring criteria that could be used to corroborate these impressions.  He developed preliminary versions of the awareness of error, mention of prevalent digits, and basis for choice rating categories, including preliminary versions of the set of options for raters to choose among with made up illustrative examples. The other author (AJN) checked the categories and descriptions for clarity and distinctness with minimum reference to the actual data. The two authors then met to refine the categories, their short names, their longer descriptions, and the illustrative examples in an attempt to span the range of variation of responses that fell into each of the categories. The authors jointly agreed on the criteria used to identify the persistent solvers as a subset of the solvers and the PD-guessers as a subset of the non-solvers. Note that this excluded several individuals who might have been late or partial solvers as well as others who chose responses other than one of the two PDs on more than 6 (corresponding to more than 10\%) of the test trials.

Next, both authors rated a pilot set of 20 participants' responses.  As with the final ratings discussed below, these ratings were carried out with access to the specific puzzle each of the participants solved and the answer the participant gave, but without reference to information about the participant's classification as a persistent solver or a PD-guesser or any other information about the participant including their performance in the training or test phases of the experiment. To make the pilot set representative without inspecting the actual responses, we sampled 9 \textit{persistent solvers} that solved the puzzle, 1 \textit{persistent solver} that did not solve the puzzle, 6 \textit{PD-guessers} that solved the puzzle, and 4 \textit{PD-guessers} that did not solve the puzzle. After independently rating all 20 participants, the authors further discussed and refined the categories, descriptions, and examples. In this process we noted that some responses were difficult to categorize with certainty, but that in these cases the uncertainty was restricted to two alternative possibilities. Accordingly we adjusted the scoring system to allow raters to provide a second choice category to capture these cases.

For the final ratings, we wished to measure the extent to which persistent solvers (a) referenced the target digit or both the target and the distractor in their response to the first free response question; (b) indicated that they had employed one of the three valid solution strategies; and (c), reported a realization of their error and explanation of why it was wrong on cases where they entered the distractor as their response instead of the target.  We wanted to demonstrate that this determination could be made reliably and in the absence of any other information beyond the specific puzzle the participant received during the questionnaire phase, the digit choice the participant entered as their solution, and their response to the first free response question about how they solved the problem.  We further wished to measure the extent to which PD-guessers exhibited these same tendencies, and if not, to document the distribution of responses they provided related to all three of these questions.  Because author JLM had previously considered all of the participants responses with additional information, this author did not contribute to the final ratings.  Instead we sought an individual who would be motivated to work conscientiously to identify the strategy used by each participant without having had any opportunity to consider any other information about each participant beyond the specific puzzle the participant saw, their digit choice, and their answer to the first free response question.  A Stanford computer science Master's student (LKS) pursuing research on human problem solving in collaboration with author JLM met this criterion.  LKS had participated in lab meetings in which the behavioral choice results from the study had been discussed, including the classification of participants as solvers or non-solvers, but had not been exposed to details of the performance of any of the participants.  LKS and author AJN then served as the two raters for the full set of participants.  Although AJN had previously considered some of the participants free responses with other information about their performance, we reasoned that if agreement between the two raters was high, this would be a sufficient indication that the ratings could be made reliably without access to additional information about the participants.  LKS was paid \$30 per hour for his participation. He first completed the experiment to familiarize himself with the study, and was classified as a persistent solver, achieving 100\% accuracy in the practice phase and 100\% accuracy in the test phase of the experiment.  He then reviewed the rating instructions and guide thoroughly before rating the 20 pilot participants.  Subsequently JLM, AJN, and LKS met to discuss the ratings criteria and discrepancies between AJN's and LKS's ratings of these participants.  As a result of this discussion we made some final refinements in the category descriptions and examples until all three were satisfied that they were as clear as possible and that all three had a common understanding of the categories. AJN and LKS then adjusted their ratings in accordance with the refined categories.  We did not require the two raters to necessarily agree perfectly on their first choice rating category, allowing disagreements to reflect the ambiguities that were present in some of the responses.

Finally AJN and LKS proceeded to rate the responses of the remaining 148 persistent solvers and PD-guessers.  To rate each participant, the rater first looked at the specific puzzle the participant had received and judged whether or not the participant's response was correct (all of these participants choices were one of the PDs).  The rater entered this judgment in a spreadsheet, which then checked whether the rater's judgment of the participant's response was correct and alerted the rater in the rare case that the rater was incorrect in judging the correctness of the participant's response (this occurred 0 times for rater LKS and 1 time for AJN).  The rater proceeded to rate the participants' free responses, proceeding through the ratings in the order listed above.  Raters were asked to complete their ratings in several sessions spread out over a one-week period, and to proceed slowly and to work for no more than an hour at a single sitting, though they could resume work after a short break.  Raters were also asked to check their ratings and to reconsider cases where they were initially uncertain, and were allowed to adjust any of their ratings after completing a first pass rating the full set.  We expected there would be some cases of inconsistency in the ratings, and did not seek to resolve these inconsistencies, so that for all but 20 of the rated participants, the final ratings of the two raters were reached without influence from the other rater.

\subsubsection{Rating consistency}
\label{sec:supp:questionnaire_rating_consistency}

\begin{figure}
    \centering
    \includegraphics[width=.75\linewidth]{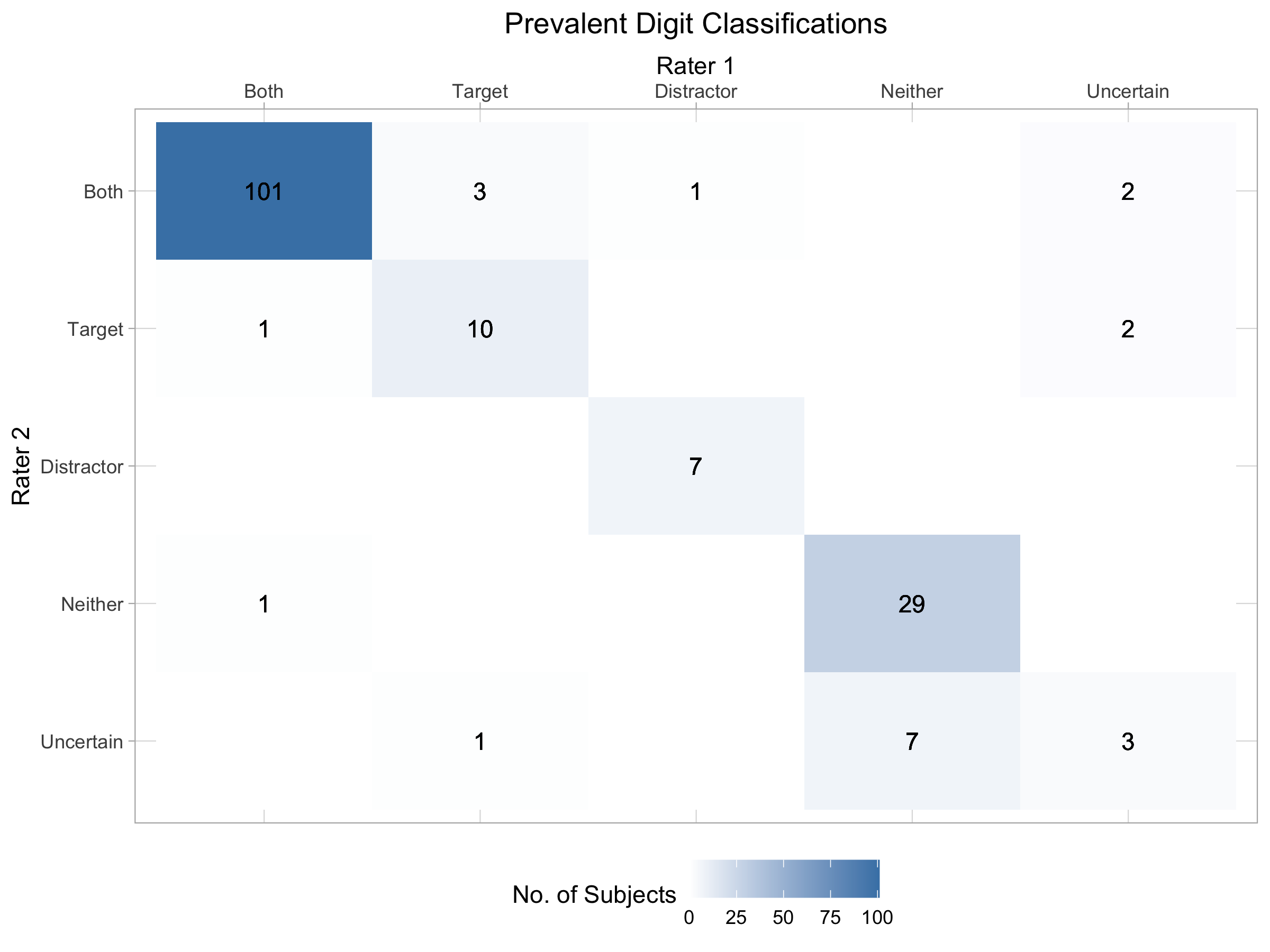}
    \caption{Prevalent Digit classifications.}
    \label{fig:q_pd}
\end{figure}

For each participant, as an attention check, raters first determined whether or not the participant's answer to the puzzle was correct. Both raters scored very highly on this with accuracies of 97\% and 100\%. Raters were also asked to determine whether or not the incorrect participants expressed awareness of their errors. Of the 4 persistent solvers and 42 PD-guessers that did not correctly solve the puzzles, both raters agreed on all 46 identifications.
The two raters had 89.3\% agreement on which PDs were mentioned by the participants (Figure \ref{fig:q_pd}).

Identifying bases for choice was overall a much tougher task, often due to incomplete ideas or ambiguous phrasing from the participants. Even when the responses were clear and relatively easy to understand, the different valid bases could be viewed as different ways of approaching the same logic, making classification difficult for some responses. Moreover, some participants mentioned multiple strategies, sometimes accounting for counterfactual scenarios where the first PD they considered ended up being the distractor but acknowledging it could just as likely have been the target.
Therefore, we allowed a second option in case a response mentioned multiple legitimate strategies or was borderline unclear.

At the superclass level, the raters had 91.7\% agreement for the solvers and 82.1\% for the non-solvers, with an overall agreement of 86.9\%.
At the subclass level, the raters had 64.3\% agreement for the solvers and 75.0\% for the non-solvers, with an overall agreement of 69.6\%.

\begin{figure}
    \centering
    \includegraphics[width=.75\linewidth]{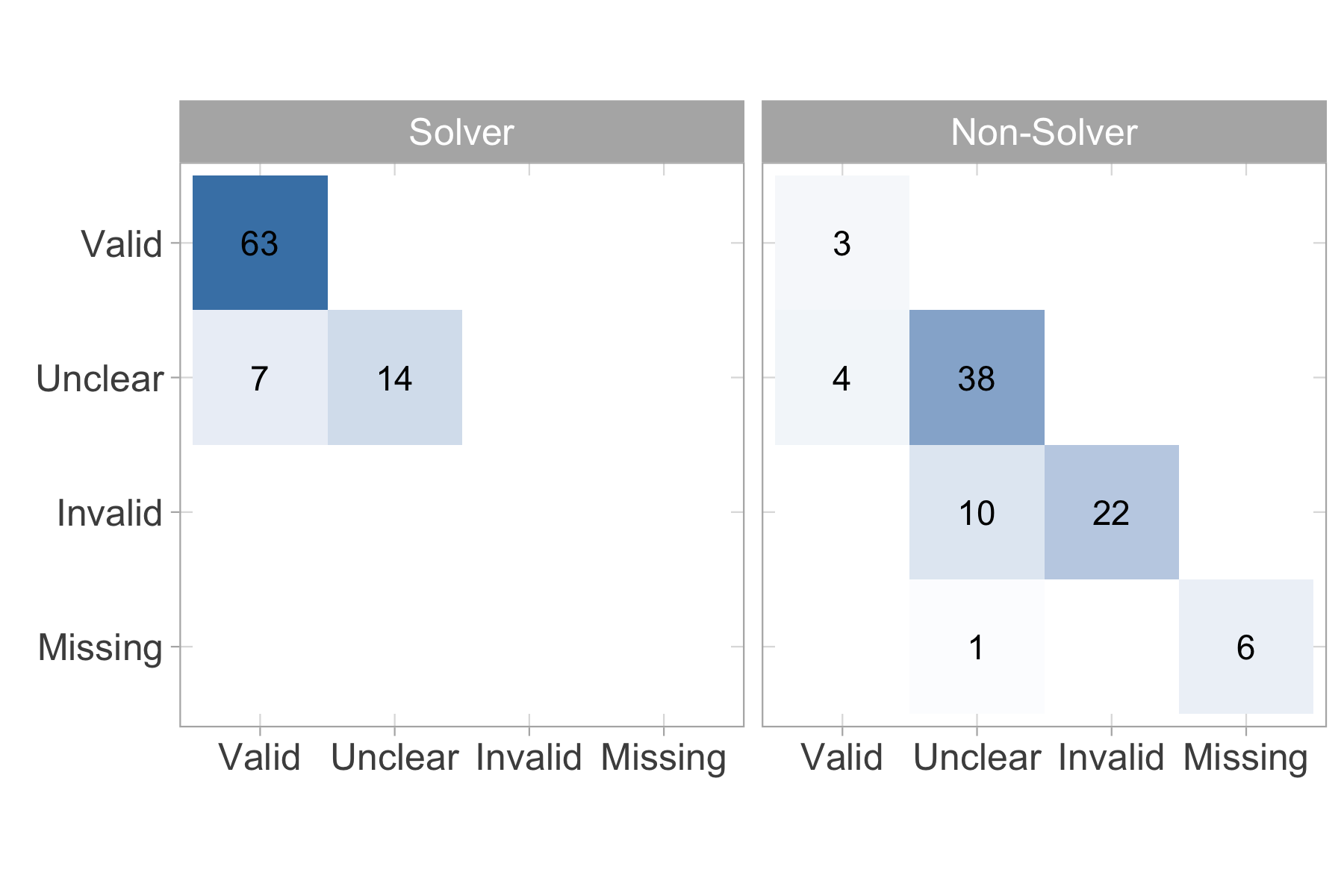}
    \caption{Superclass ratings of basis for choice by each rater. Only first choices shown.}
    \label{fig:supp:agreement_basis}
\end{figure}

\begin{figure}
    \centering
    \includegraphics[width=.75\linewidth]{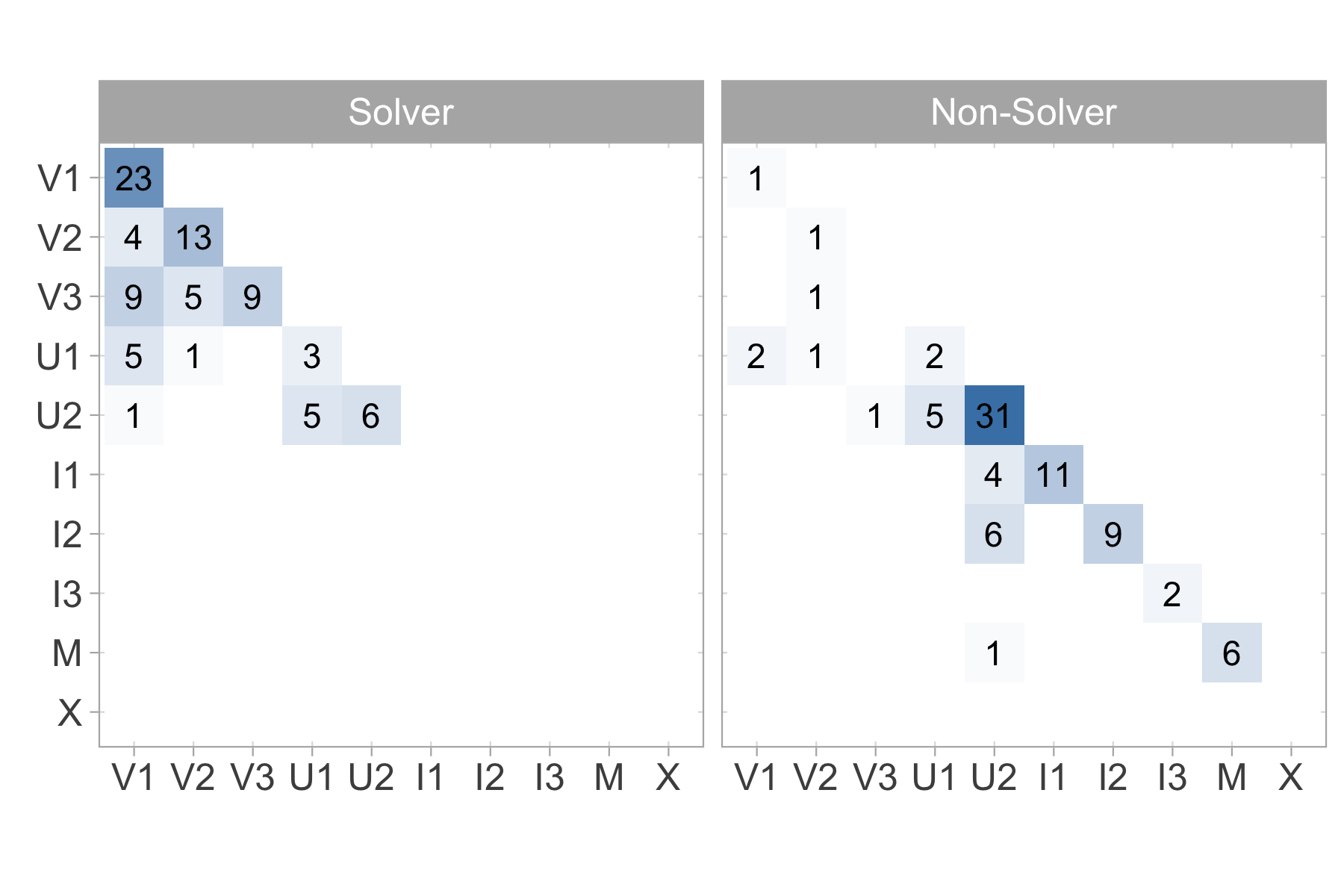}
    \caption{Subclass ratings of basis for choice by each rater. Only first choices shown.}
    \label{fig:supp:agreement_subbasis}
\end{figure}

\section{Recurrent relational network}
\label{sec:supp:rrn}

To compare how contemporary neural network models learn and generalize solving the hidden single puzzles, we replicated and adapted a model that is state-of-the-art in solving Sudoku puzzles \cite{palm2018recurrent}.
The recurrent relational network (RRN) uses a relational message passing scheme where in each time step, it computes for each cell in the grid an update instruction for each other cell that the cell shares a house with. 
The model achieves systematicity with respect to the positional variables by sharing the same connection weights to each cell from all of the relevant constraining cells, thereby remaining invariant to the particular cell it solves for and the particular relevant cell constraining it. 

\subsection{Model architecture}

The recurrent relational network uses a local message passing scheme in a graph where each cell in the Sudoku grid is a separate node. 
Cells that share houses (row, column, or box) are considered neighbors and their nodes share edges between them. Each node $i$ at time step $t$ is represented by a hidden state vector $h_i^t$, where $h_i^1 = x_i$ is an embedding of the cell's initial state (e.g. blank or a clue). In the original paper, the initial cell state is given by
\begin{equation}
    x_i = \text{MLP(concat(embed(}d_i), \text{embed(row}_i), \text{embed(column}_i)))
\end{equation}
where $d_i$ is the initial content of the cell, if any. 
In our implementation, we found the cell coordinate information detrimental to performance and simplified the embedding to $x_i =$ embed($d_i$).

At each step, for each cell $i$ and its neighbor $j$, a vector representing the message from $j$ to $i$ is computed using 
\begin{equation}
    m_{ij}^t = \text{MLP}(\text{concat}(h_i^{t-1}, h_j^{t-1}))
\end{equation}
Then the messages from all of $i$'s neighbors are summed as
\begin{equation}
    m_i^t = \sum\limits_j m_{ij}^t
\end{equation}

The subsequent hidden state vector of each cell is updated using
\begin{equation}
    h_i^t, c_i^t = \text{LSTM(MLP(concat(}x_i, m_i^t)), c_i^{t-1})
\end{equation}

In our implementation, however, we found a simple linear layer to be sufficient replacements for the two MLPs.

An output vector is also calculated for each step and cell using a linear decoding layer
\begin{equation}
    o_i^t = \text{linear}(h_i^t)
\end{equation}
which is then used to calculate the cross-entropy loss with the solution for the cell.

\subsection{Replication of results}
We attempted to replicate the findings of Palm et al. using the same dataset and parameters as the original network. However, due to limited computational resources, we had to use our simplified architecture and fewer parameters. Specifically, we used a digit embedding size of 10 (9 digits + blank; original model used 16) and reduced the MLPs to single linear layers. Moreover, we reduced the number of training epochs to 100 and the batch size to 20. Our replication attempt resulted in 97.0\% of all cells solved and 74.2\% of test puzzles fully solved for all 81 cells. While this is much lower than the 96.6\% reported in the original paper, it demonstrates that our simplified architecture can achieve a significant level of success while retaining key features of the original architecture.

\begin{figure}[!ht]
\centering
\includegraphics[width=\linewidth]{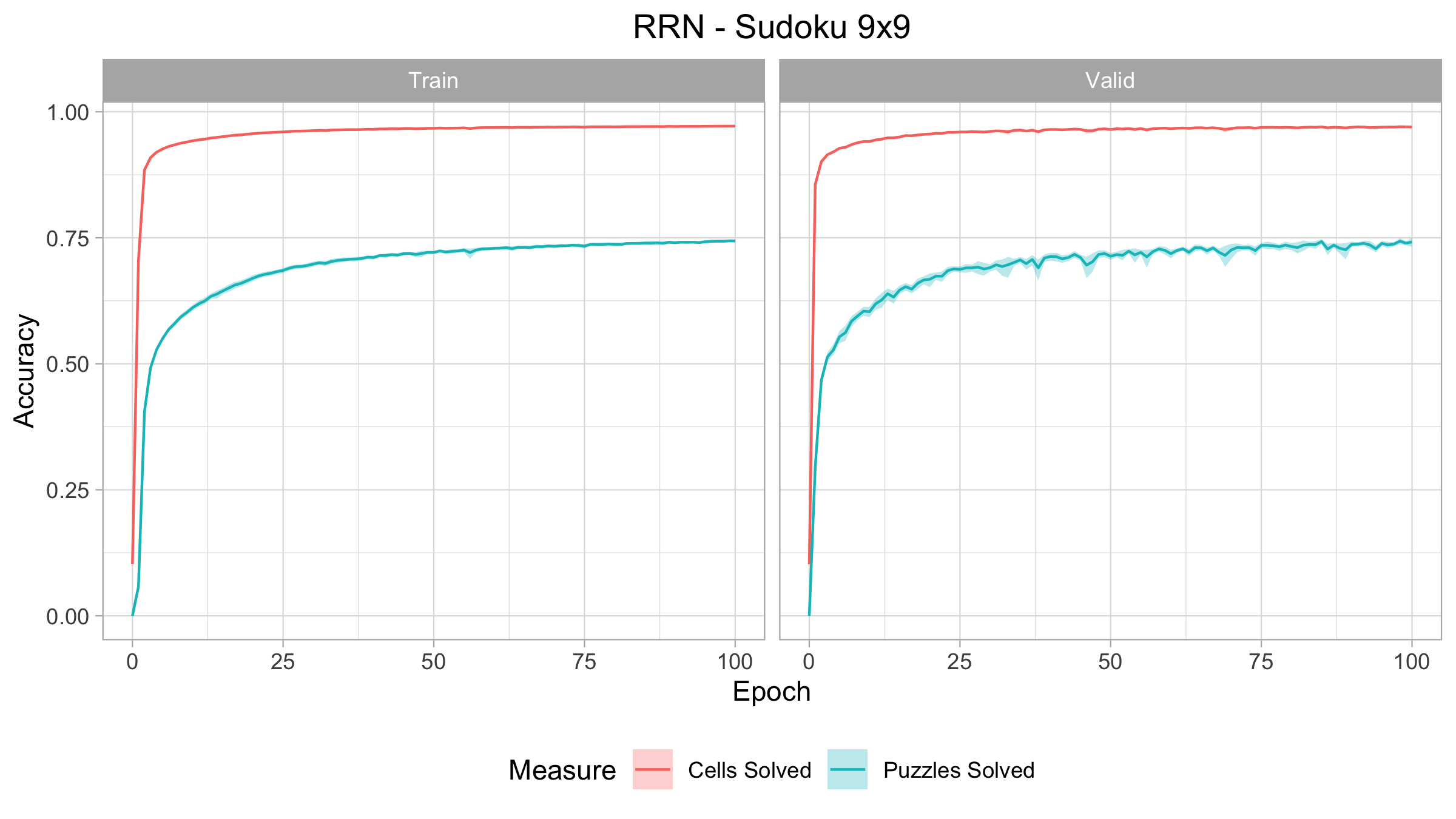}
\caption{Replication results of Palm et al. on full 9x9 Sudoku puzzles. Results show averages across 10 different model instances.}
\label{fig:rrn_sudoku}
\end{figure}

\subsection{Solving hidden single puzzles}

To compare the model to human performance, we trained and tested the RRN model using the hidden single puzzles similar to what we provided for our participants. 
Specifically, we created a set of puzzles with the same digit set, house index, cell index, and house type to train the model. 
We then generated 64 puzzles with varied features to test for the model's ability to generalize. Because the majority of the grid was empty, we only calculated the cross-entropy loss for the goal cell and the initial hint cells. We found the latter to be a critical auxiliary signal for the model to successfully train.

The model was written using PyTorch and trained using Adam with batch size = 100, learning rate = 0.001 (original model used 0.0002), and L2 regularization of 0.0001. We used a digit embedding size of 10 and a hidden layer size of 48 (original model used 96). We found these parameters sufficient given the significantly lower complexity of the hidden single puzzles compared to full Sudoku puzzles.

Using the same set of restrictions on the practice puzzles as described in our human experiments, we trained the model with varying numbers of training samples, stopping if and when it reached 99\% accuracy for puzzles with the same features.
We then tested the trained model with the same systematic variations we considered for human participants.
Comparing the model's training and generalization results to human solvers, we observed two key properties of the network.

\begin{figure}[!ht]
\begin{subfigure}{.49\linewidth}
  \centering
  \caption{RRN - Validation Set Accuracy}
  \vspace{-6pt}
  \includegraphics[width=\linewidth]{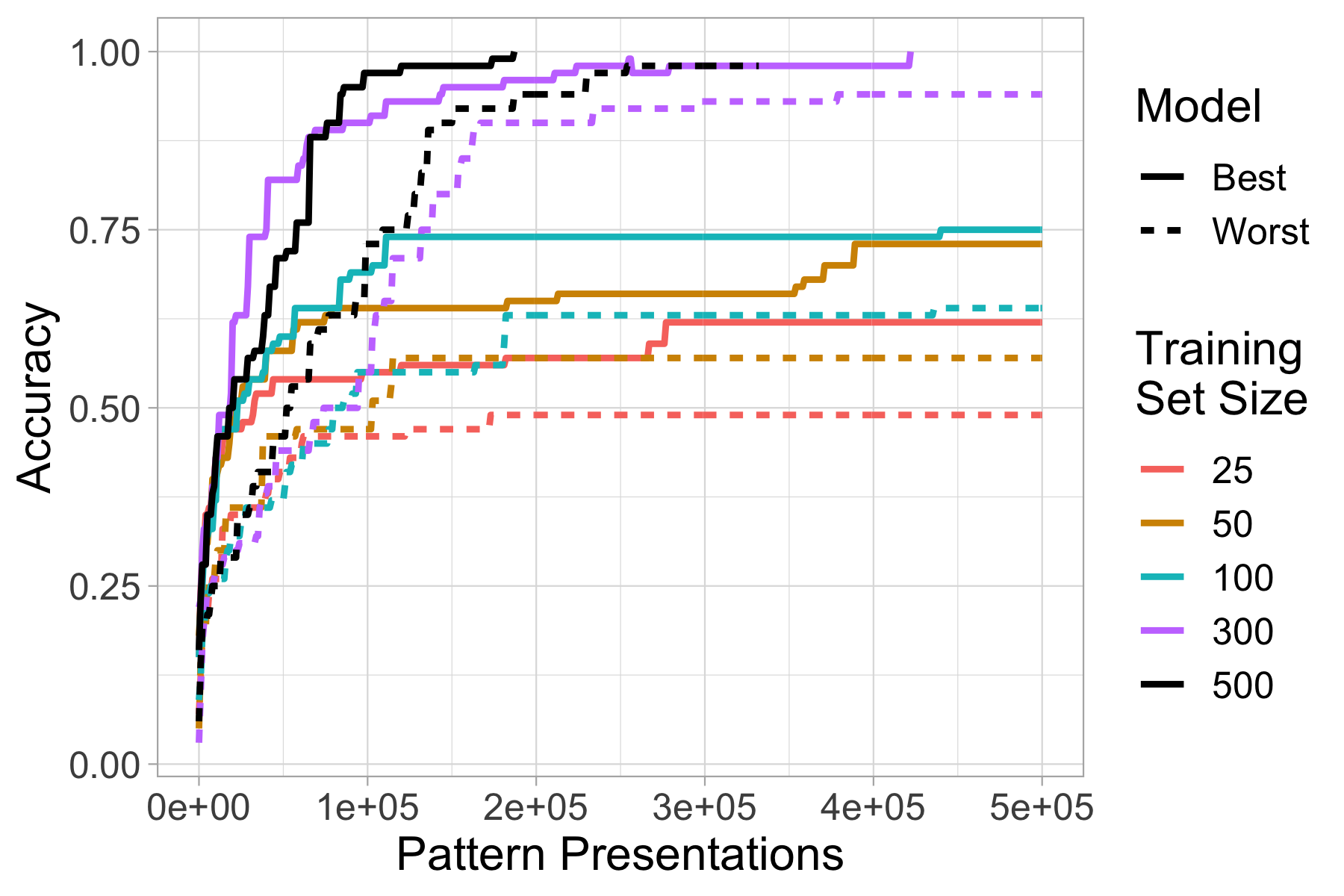}
  \label{fig:supp:rrn_a}
  \vspace{14pt}
\end{subfigure}
\begin{subfigure}{.49\linewidth}
  \centering
  \caption{DRRN - Validation Set Accuracy}
  \vspace{-6pt}
  \includegraphics[width=\linewidth]{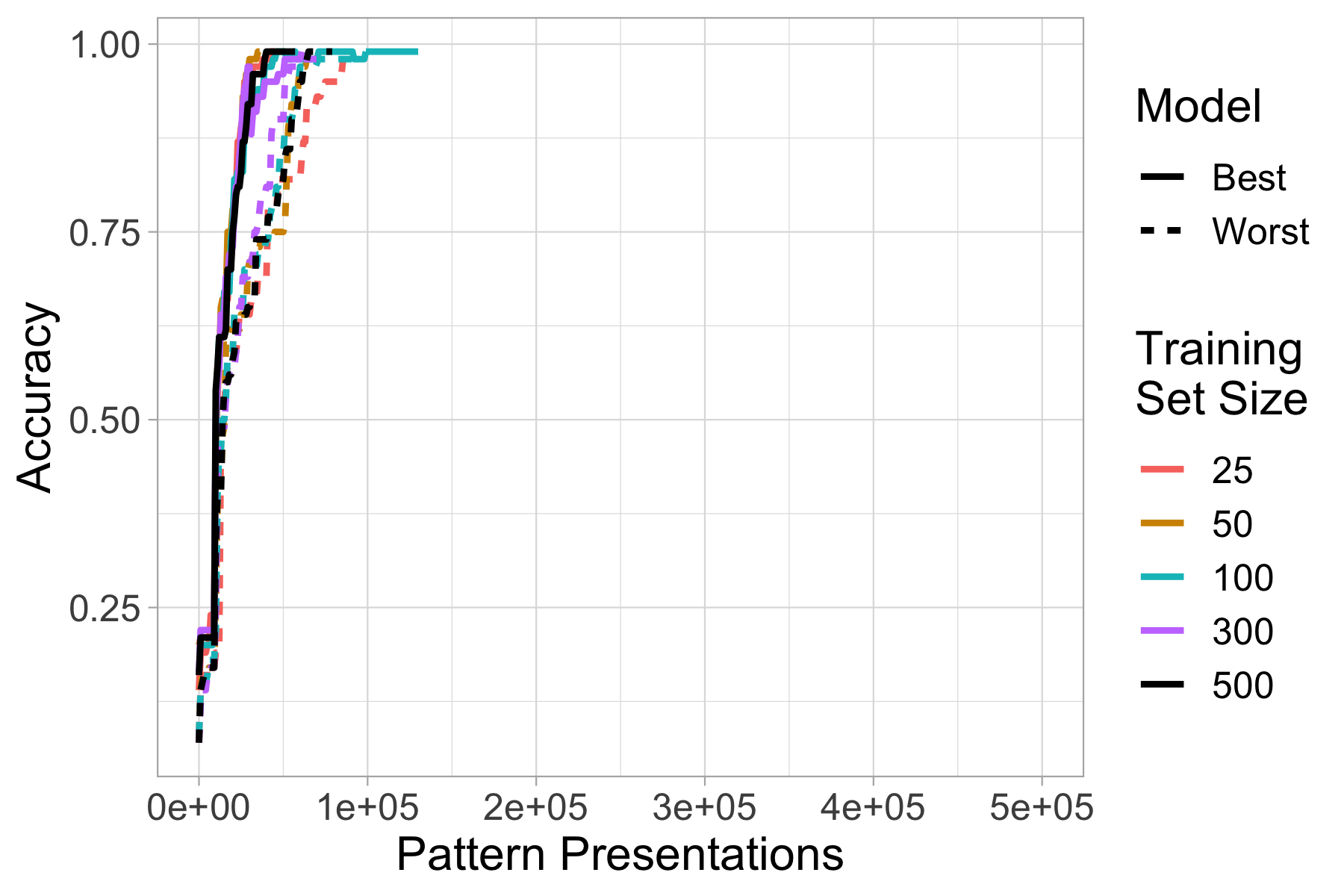}
  \label{fig:supp:rrn_b}
  \vspace{14pt}
\end{subfigure}
\begin{subfigure}{.49\linewidth}
  \centering
  \caption{Test Set Accuracy}
  \vspace{-6pt}
  \includegraphics[width=\linewidth]{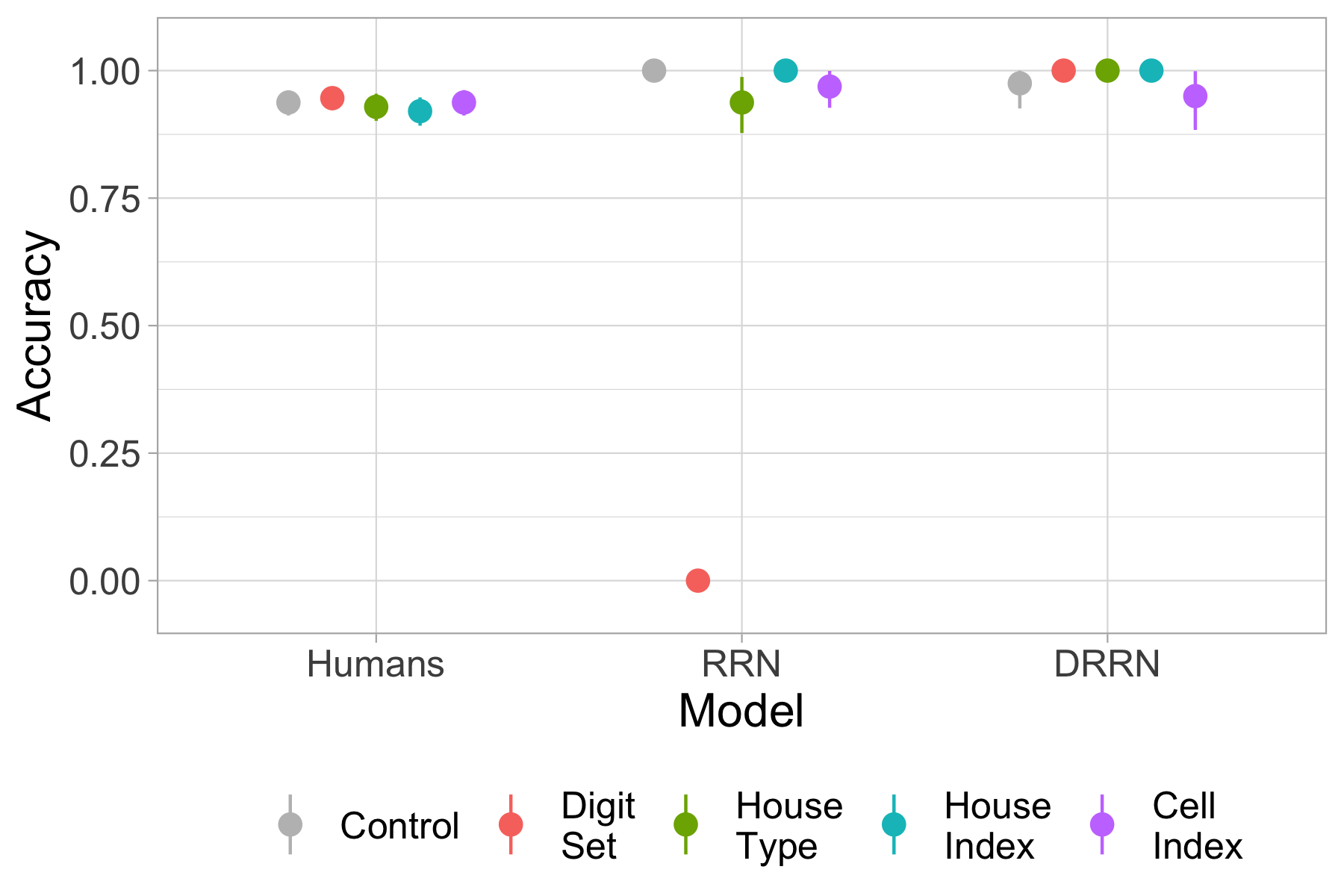}
  \label{fig:supp:rrn_c} 
\end{subfigure}
\caption{
(a-b) Recurrent relational network's (RRN) and Digit-invariant RRN's (DRRN) validation accuracies on held-out puzzles with the same features as training samples. The x-axis represents total number of training puzzles presented to the model and the y-axis represents cumulative maximum accuracy during training. 10 models were trained for each training set size. \textit{Best} and \textit{Worst} lines indicate the highest and lowest cumulative maximum accuracy among the 10 model instances respectively.
(c) Test set accuracy by feature for human solvers, the RRN, and the DRRN. Error bars indicate 95\% highest density intervals. 
}
\label{fig:supp:rrn}
\end{figure}

First, as shown in Figure~\ref{fig:supp:rrn_a}, the model trained inefficiently compared to the human participants, requiring 300 unique puzzles with tens of thousands of total pattern presentations to gradually reach accuracies comparable to the solvers. 
Second, given the same restrictions on the particulars of the training puzzles as described in our human experiments, the model generalized immediately, performing nearly perfectly on problems with changes to the positional features (house type, house index, and cell index) but could not solve a single puzzle with changed digit sets (Figure~\ref{fig:supp:rrn_c}).

\subsection{Inducing digit invariance}
\label{sec:supp:drrn}

If weight sharing across spatial features produces spatial invariance, it stands to reason that weight sharing across the digit feature can produce digit invariance.
Following this logic, we expanded the Palm RRN by defining a node for each cell and digit (x, y, digit), thus producing $9^3 = 729$ nodes. We defined edges between any two nodes that represent different digits of the same cell or two nodes that represent the same digit of two cells sharing a house. 
For example, nodes for (3, 3, 1) and (3, 3, 6) would share an edge because they both represent the cell at (3, 3). Nodes for (3, 3, 9) and (3, 6, 9) would share an edge as they share the same house at Row 3 and the same digit, but (3, 3, 7) and (3, 6, 9) would not share an edge since they do not share the same digit.

Since individual nodes represent digits, we no longer needed a layer to embed numbers. 
The output layer was also modified such that each node maps to a scalar logit, and the 9 digit nodes of a cell would together be softmaxed to produce the output probability vector. 
Due to memory resource constraints, the hidden state and message vector sizes were reduced to 16 and the batch size reduced to 10.
We refer to this variant of the RRN as the digit-invariant recurrent relational network (DRRN)

Adding this inductive bias strongly constrained the solution space of the model such that it could reach near perfect accuracy using only 25 training examples, a level comparable to human solvers.
In general, the DRRN was able to solve the hidden single puzzles but was often unstable throughout its training process, possibly due to its low-dimensional hidden vectors. Therefore, to offer a fair comparison, we evaluated the test puzzles on epochs where the models had the highest validation accuracy.
As hypothesized, we found that the DRRN successfully generalized to the digit set conditioned puzzles as it did in the other three feature variations.

\begin{figure}[!ht]
    \centering
    \includegraphics[width=\linewidth]{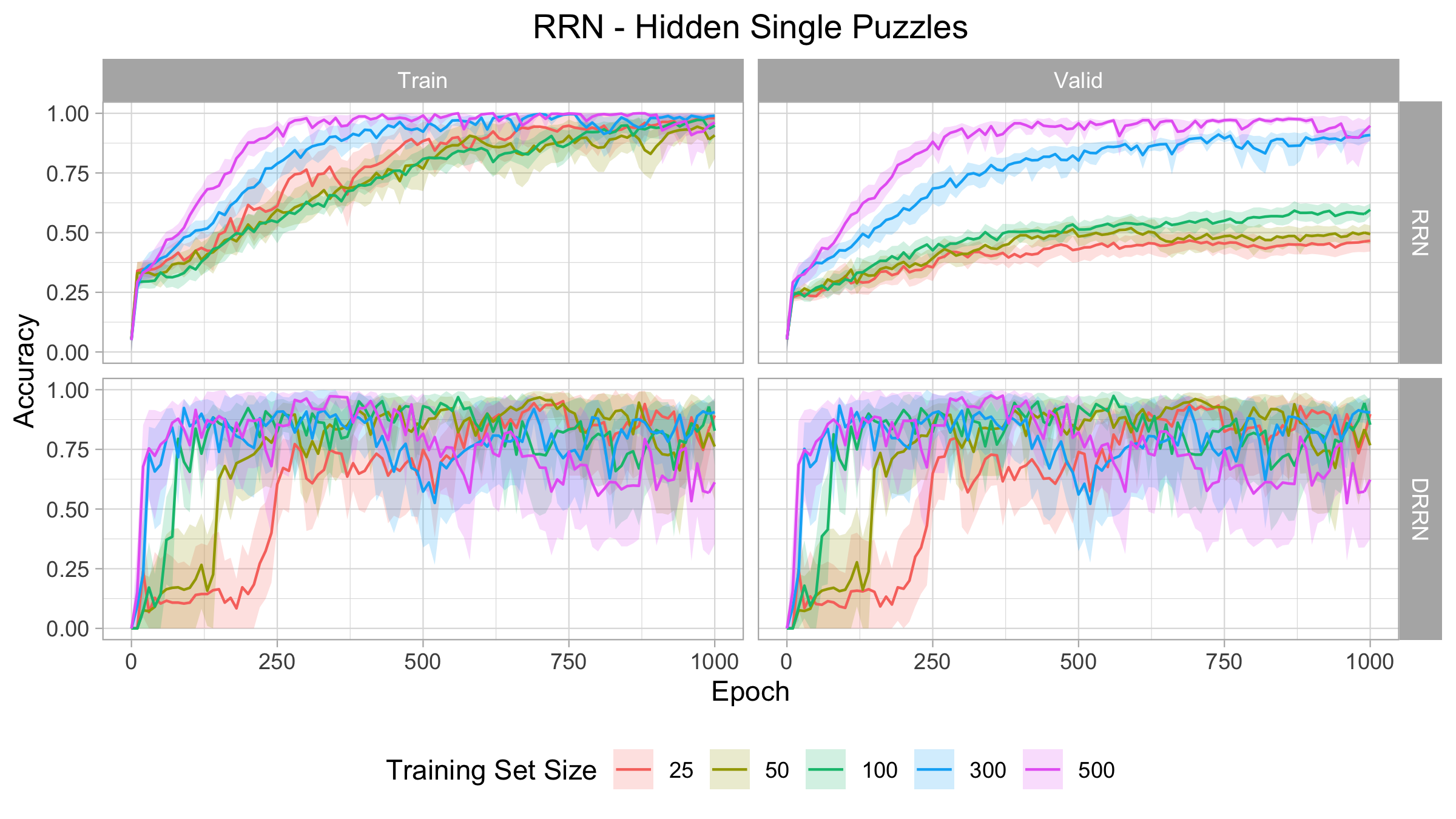}
    \caption{Accuracy of relational networks on training and held-out validation (puzzles sharing features as training set) sets. Results show average performance across 10 different model instances trained on 10 different datasets. The same 10 datasets were used to train both RRNs and DRRNs.}
    \label{fig:rrn_train}
\end{figure}

\begin{figure}[!ht]
    \centering
    \includegraphics[width=\linewidth]{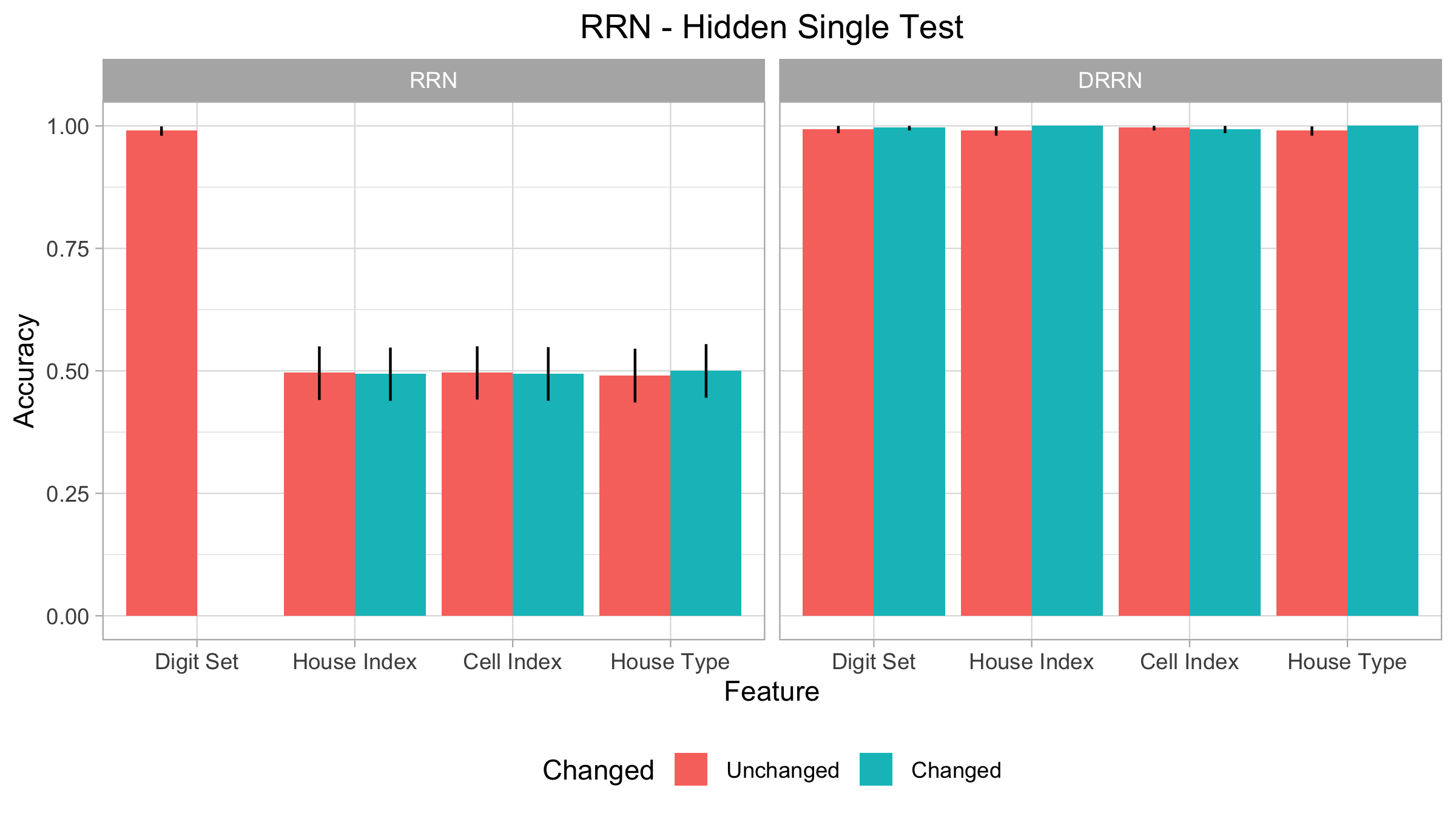}
    \caption{Accuracy of relational networks on held-out test set with systematically varied features (marginalizing over others), evaluated on models trained with 500 puzzles selected on epochs with highest validation accuracies. Results show average performance across 10 different model instances. Error bars indicate 95\% highest density intervals.}
    \label{fig:rrn_test}
\end{figure}

\begin{figure}[!ht]
    \centering
    \includegraphics[width=\linewidth]{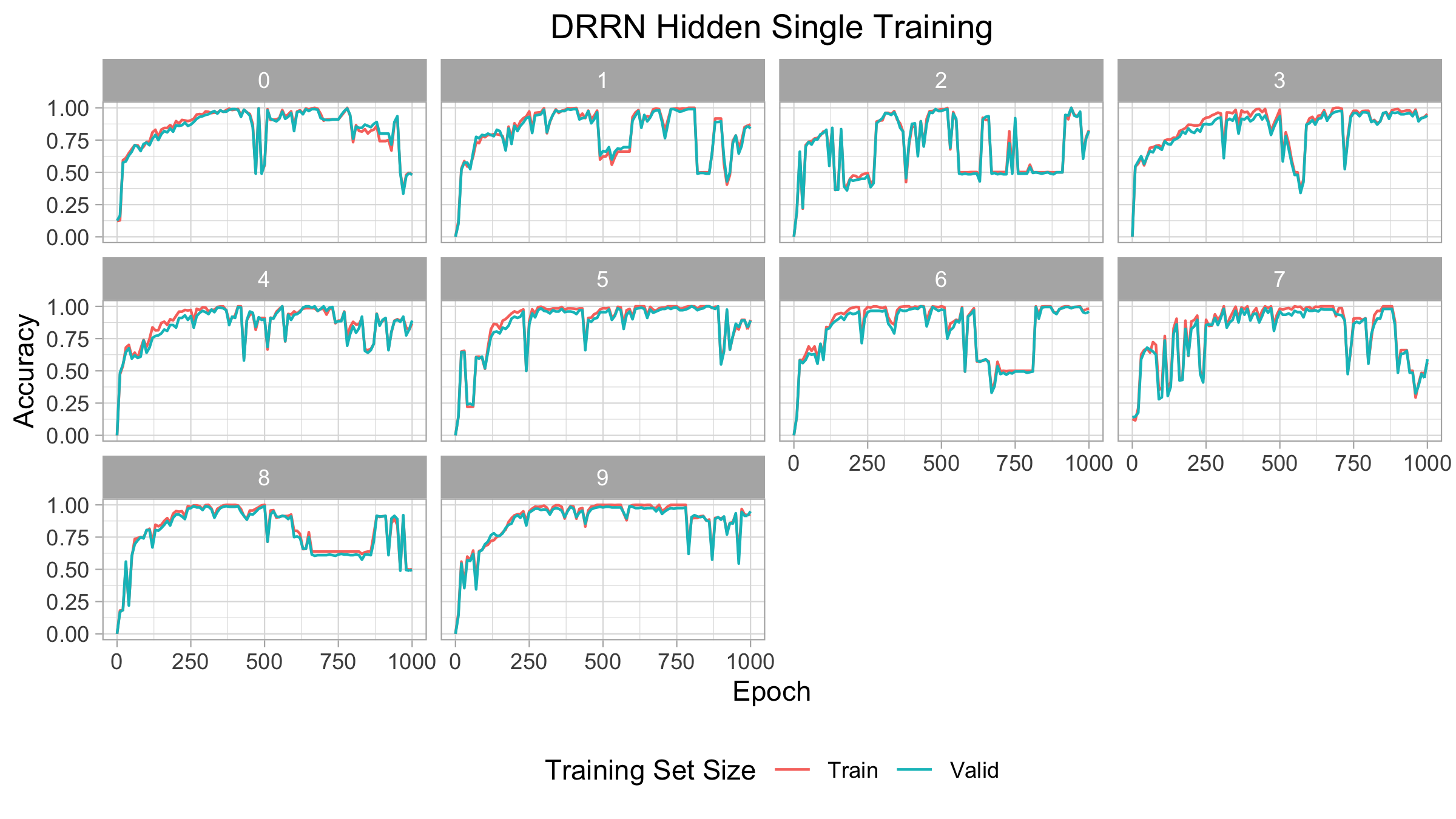}
    \caption{Accuracy of DRRN on training and held-out validation sets. Each plot shows a different instance of the model trained on different sets of 500 training puzzles.}
    \label{fig:drrn}
\end{figure}

\section{GPT-3}
\label{sec:supp:gpt3}

We evaluated GPT-3's ability to solve the same hidden single puzzles that were presented to the participants.
Because GPT-3 does not have a visual interface, we encoded the puzzles as a series of (row, column, digit) tuples.
While this is admitted very different from the usual grid format, we found that GPT-3 could use the information to solve simple queries using this format and generally outperformed an array representation which we also tried.
We used GPT-3 davinci-003 for all experiments, and although this is only one iteration of large language models, we hope that it offers some insight into how state-of-the-art foundation models may reason.

\subsection{Preliminaries}

Before attempting to query GPT-3 with the hidden single puzzles, we first sought to gauge its familiarity with Sudoku-related puzzles in general using problems that were used during the tutorial of the experiment.
Each prompt was presented in a separate session and to our knowledge should not affect GPT-3's responses to other prompts.

\begin{enumerate}
    \item \textbf{Prompt:} What is Sudoku? \\
        \textbf{GPT-3:}
        Sudoku is a logic-based number-placement puzzle. The objective is to fill a 9×9 grid with digits so that each column, each row, and each of the nine 3×3 sub-grids that compose the grid contain all of the digits from 1 to 9.\\
        \textbf{Evaluation:} Correct
    \item \textbf{Prompt:} What is the hidden single technique in Sudoku?\\
        \textbf{GPT-3:}
        The hidden single technique in Sudoku is a strategy used to solve puzzles by looking for a number that can only appear in one cell within a row, column, or box. This number can then be entered into the cell, eliminating the need to try other possibilities.\\
        \textbf{Evaluation:} Correct
    \item \textbf{Prompt:} Here are the contents of a 9 by 9 grid.\\
        (row 1, column 7) contains 6.\\
        (row 2, column 7) contains 8.\\
        (row 3, column 7) contains 7.\\
        (row 4, column 7) contains 4.\\
        (row 5, column 7) contains 1.\\
        (row 6, column 7) contains 3.\\
        (row 7, column 7) contains 2.\\
        (row 8, column 7) contains 5.\\
        (row 9, column 7) contains 9.\\
        Where is the 1 located?\\
        \textbf{GPT-3:}
        The 1 is located in row 5, column 7.\\
        \textbf{Evaluation:} Correct
    \item \textbf{Prompt:} Sudoku is a puzzle with a 9x9 grid of numbers where each row, column, and 3x3 box must contain exactly one of each number from 1 to 9.\\
        Assume you have a grid with the following numbers.\\
        (row 1, column 7) contains 6.\\
        (row 2, column 7) contains 7.\\
        (row 3, column 7) contains 8.\\
        (row 4, column 7) contains 2.\\
        (row 5, column 7) contains 1.\\
        (row 6, column 7) contains 4.\\
        (row 7, column 7) contains 1.\\
        (row 8, column 7) contains 5.\\
        (row 9, column 7) contains 9.\\
        Column 7 does not contain every number between 1 and 9, but rather contains two copies of the digit 1, forming a contradiction. Which two cells create this contradiction?\\
        \textbf{GPT-3:}
        The cells that create the contradiction are (row 5, column 7) and (row 7, column 7), which both contain the digit 1.\\
        \textbf{Evaluation:} Correct
    \item \textbf{Prompt:} Assume you have a Sudoku grid with the following numbers.\\
        (row 1, column 7) contains 6.\\
        (row 2, column 7) contains 7.\\
        (row 3, column 7) contains 8.\\
        (row 4, column 7) contains 2.\\
        (row 6, column 7) contains 4.\\
        (row 7, column 7) contains 1.\\
        (row 8, column 7) contains 5.\\
        (row 9, column 7) contains 9.\\
        What number should now go in (row 5, column 7) so that the column contains every number between 1 and 9?\\
        \textbf{GPT-3:} 3\\
        \textbf{Evaluation:} Correct
    \item \textbf{Prompt:} Assume you have a Sudoku grid with the following numbers.\\
        (row 1, column 8) contains 3.\\
        (row 4, column 7) contains 2.\\
        (row 6, column 7) contains 4.\\
        (row 7, column 2) contains 3.\\
        (row 8, column 7) contains 5.\\
        (row 9, column 6) contains 3.\\
        Where is the 3 that is preventing (row 9, column 7) from being a 3?\\
        \textbf{GPT-3:}
        The 3 that is preventing (row 9, column 7) from being a 3 is located in (row 9, column 6).\\
        \textbf{Evaluation:} Correct
    \item \textbf{Prompt:} Given a Sudoku grid with the following numbers, where is the 3 that is preventing (row 1, column 7), (row 2, column 7), and (row 3, column 7) from containing a 3?\\
        (row 1, column 8) contains 3.\\
        (row 4, column 7) contains 2.\\
        (row 6, column 7) contains 4.\\
        (row 7, column 2) contains 3.\\
        (row 8, column 7) contains 5.\\
        (row 9, column 6) contains 3.\\
        \textbf{GPT-3:} The 3 that is preventing (row 1, column 7), (row 2, column 7), and (row 3, column 7) from containing a 3 is located at (row 5, column 7).\\
        \textbf{Evaluation:} Incorrect
    \item \textbf{Prompt:} Given a Sudoku grid with the following numbers, can (row 1, column 7) contain a 3?\\
        (row 1, column 8) contains 3.\\
        (row 4, column 7) contains 2.\\
        (row 6, column 7) contains 4.\\
        (row 7, column 2) contains 3.\\
        (row 8, column 7) contains 5.\\
        (row 9, column 6) contains 3.\\
        \textbf{GPT-3:} No, (row 1, column 7) cannot contain a 3 since it would create a conflict with the 3 already in (row 1, column 8).\\
        \textbf{Evaluation:} Correct
    \item \textbf{Prompt:} In a Sudoku grid containing the following numbers, can (row 2, column 7) contain a 3?\\
        (row 1, column 8) contains 3.\\
        (row 4, column 7) contains 2.\\
        (row 6, column 7) contains 4.\\
        (row 7, column 2) contains 3.\\
        (row 8, column 7) contains 5.\\
        (row 9, column 6) contains 3.\\
        \textbf{GPT-3:} No, (row 2, column 7) cannot contain a 3, as that would violate the rule of Sudoku that every row, column, and 3x3 subgrid must contain each number from 1 to 9 exactly once.\\
        \textbf{Evaluation:} Correct
    \item \textbf{Prompt:} In a Sudoku grid containing the following numbers, can (row 7, column 7) contain a 3?\\
        (row 1, column 8) contains 3.\\
        (row 4, column 7) contains 2.\\
        (row 6, column 7) contains 4.\\
        (row 7, column 2) contains 3.\\
        (row 8, column 7) contains 5.\\
        (row 9, column 6) contains 3.\\
        \textbf{GPT-3:} Yes, (row 7, column 7) can contain a 3.\\
        \textbf{Evaluation:} Incorrect
\end{enumerate}

\subsection{Hidden single puzzles}

We gave GPT-3 several different forms of prompts in attempt to maximize its ability to solve the puzzles.
For each type of prompt, we tried both list and array representations of the grid, but found that the list representation elicited responses as good as, if not better than, the responses elicited using array representations.

At the beginning of each prompt, we prepended the phrase ``I am a highly intelligent puzzle solving bot.'' styled after the Q\&A prompt example on OpenAI's API website: 
``I am a highly intelligent question answering bot. If you ask me a question that is rooted in truth, I will give you the answer. If you ask me a question that is nonsense, trickery, or has no clear answer, I will respond with `Unknown'.''

In all cases without chain-of-thought prompts, we found that GPT-3' generally responds with numbers that do not appear in the puzzle grid, suggesting that it uses a strong form of the ADC strategy, guaranteeing a 0\% accuracy.
We explicitly introduced the strategy of choosing between the two prevalent digits in the chain-of-thought examples, which allowed the model to reach 50\% by guessing between the target and the distractor.

Below, an example of each format using the list representation and one example using the array representation.
The full set of prompts and responses can be found in the online repository.

\clearpage
\subsubsection{0-shot using array representation}
Below is one example of the hidden single puzzle presented using the array representation.
We used this representation in other formats, including CoT prompting, but we show only one here to illustrate what the array representation looked like.

GPT-3 solved 0 out of 50 puzzles using this representation.

\textbf{Prompt:}
Q: Consider a Sudoku grid with the following numbers:
\begin{verbatim}
[X 0 0 0 0 0 0 0 0]
[7 0 0 0 0 0 0 0 0]
[0 0 0 9 0 0 0 0 6]
[5 0 0 0 0 0 0 0 0]
[8 0 0 0 0 0 0 0 0]
[0 0 0 0 0 0 9 0 0]
[0 0 0 6 0 0 0 0 0]
[0 6 0 0 0 0 0 0 0]
[0 9 0 0 0 0 0 0 0]
\end{verbatim}
What number must X be?\\
Hint: use the hidden single technique along the column that contains X.\\
A: X must be\\
\textbf{GPT-3:} 4.\\
\textbf{Expected:} 9

\clearpage
\subsubsection{0-shot using list representation}
Below is one example of the hidden single puzzle presented using the list representation.

GPT-3 solved 0 out of 50 puzzles using this representation.

\textbf{Prompt:}
Q: Consider a Sudoku grid with the following numbers:\\
(row 6, column 7) contains 9\\
(row 7, column 4) contains 6\\
(row 2, column 1) contains 7\\
(row 8, column 2) contains 6\\
(row 9, column 2) contains 9\\
(row 3, column 4) contains 9\\
(row 4, column 1) contains 5\\
(row 3, column 9) contains 6\\
(row 5, column 1) contains 8\\
What number must (row 1, column 1) contain?\\
Hint: use the hidden single technique along column 1.\\
A: (row 1, column 1) must contain\\
\textbf{GPT-3:} 4.\\
\textbf{Expected:} 9

\clearpage
\subsubsection{4-shot prompts}
We gave GPT-3 four examples of puzzles and their solutions before presenting it with a 5th puzzle to solve.
GPT-3 solved 3 out of 50 puzzles using this format.

Below is one solved example:

Q: Consider a Sudoku grid with the following numbers:\\
(row 5, column 3) contains 6\\
(row 9, column 1) contains 1\\
(row 2, column 5) contains 5\\
(row 6, column 2) contains 5\\
(row 8, column 1) contains 9\\
(row 7, column 1) contains 7\\
(row 3, column 8) contains 5\\
(row 2, column 8) contains 6\\
(row 4, column 9) contains 6\\
What number must (row 1, column 1) contain?\\
Hint: use the hidden single technique along column 1.\\
A: (row 1, column 1) must contain 5

\clearpage
\subsubsection{4-shot chain-of-thought prompts}
We tried giving GPT-3 four detailed chain-of-thought examples before presenting it with a 5th puzzle to solve.
The solved examples identify the two prevalent digits to choose between and follows the logic of solving using one of them.
Since human solvers generally selected between the target and distractor arbitrarily as the candidate digit to explore first, we wrote CoT prompts to reason similarly.
Depending on whether the initial candidate digit happened to be the target or the distractor, the steps would either lead to finding that the target digit cannot go in the target house other than in the goal cell, or finding that the distractor could potentially go in another cell.
Thus, we wrote two types of solution steps: target-first and distractor-first.

We tested GPT-3 using four target-first examples with which it solved 27 out of 50 puzzles.
It solved 35 out of 50 puzzles when given two target-first examples and distractor-first puzzles.

Below is an example for each type of solution.

\clearpage
\textbf{Target-first}\\
Q: Consider a Sudoku grid with the following numbers:\\
(row 8, column 1) contains 9\\
(row 2, column 5) contains 5\\
(row 9, column 1) contains 1\\
(row 7, column 1) contains 7\\
(row 5, column 3) contains 6\\
(row 3, column 8) contains 5\\
(row 6, column 2) contains 5\\
(row 2, column 8) contains 6\\
(row 4, column 9) contains 6\\
What number must (row 1, column 1) contain?\\
Hint: use the hidden single technique along column 1.\\
A: To solve for (row 1, column 1), we need to find the number that cannot go in any other cell in its column. Between the two numbers that appear 3 times on the grid, 5 and 6, we arbitrarily choose 5 as a candidate to check whether it can go in the other cells. 5 cannot go in (row 8, column 1), (row 9, column 1), or (row 7, column 1) because these cells already have numbers in them. (row 4, column 1), (row 5, column 1), and (row 6, column 1) cannot contain 5 because they share a 3x3 box with 5 in (row 6, column 2). (row 3, column 1) cannot contain 5 because it shares a row with 5 in (row 3, column 8), and (row 2, column 1) cannot contain 5 because it shares a row with 5 in (row 2, column 5). Since we have eliminated all other eight cells in column 1 as potential locations to place to put the 5, we can conclude that (row 1, column 1) must contain the 5.

\clearpage
\textbf{Distractor-first}\\
Q: Consider a Sudoku grid with the following numbers:\\
(row 8, column 7) contains 6\\
(row 6, column 1) contains 3\\
(row 2, column 5) contains 9\\
(row 7, column 8) contains 9\\
(row 1, column 2) contains 9\\
(row 7, column 4) contains 6\\
(row 5, column 1) contains 1\\
(row 3, column 3) contains 6\\
(row 9, column 1) contains 5\\
What number must (row 4, column 1) contain?\\
Hint: use the hidden single technique along column 1.\\
A: To solve for (row 4, column 1), we need to find the number that cannot go in any other cell in its column. Between the two numbers that appear 3 times on the grid, 9 and 6, we arbitrarily choose 9 as a candidate to check whether it can go in the other cells. 9 cannot go in (row 9, column 1), (row 6, column 1), or (row 5, column 1) because these cells already have numbers in them. (row 1, column 1), (row 2, column 1), and (row 3, column 1) cannot contain 9 because they share a 3x3 box with 9 in (row 1, column 2). (row 7, column 1) cannot contain 9 because it shares a row with 9 in (row 7, column 8). However, there are no numbers that prevent either (row 4, column 1) or (row 8, column 1) from containing 9, so we cannot be certain about either. Choosing 6 as the new candidate and applying a similar logic, we can see that the only cell in column 1 that can contain 6 is (row 4, column 1), so we can conclude that the answer is 6.

\end{document}